\documentclass{article}
\pdfoutput=1
\PassOptionsToPackage{round, compress}{natbib}

\usepackage[preprint]{neurips_2024}

\usepackage[utf8]{inputenc} 
\usepackage[T1]{fontenc}    
\usepackage{hyperref}       
\usepackage{url}            
\usepackage{booktabs}       
\usepackage{amsfonts}       
\usepackage{nicefrac}       
\usepackage{microtype}      
\usepackage{xcolor}         

\usepackage{graphicx}
\usepackage{mathptmx}
\usepackage{multirow}
\usepackage{bookmark}
\usepackage{amsmath}

\usepackage{algorithm}
\usepackage{algpseudocode}
\usepackage{float}
\usepackage{amsthm}
\newtheorem{theorem}{Theorem}
\usepackage{cases}

\usepackage{wrapfig}
\usepackage{subcaption}
\usepackage{bbm}
\usepackage{bm}
\usepackage{threeparttable}
\usepackage{colortbl}
\usepackage{amssymb}
\usepackage{mathtools}
\usepackage{array} 

\title{Minusformer: Improving Time Series Forecasting by Progressively Learning Residuals}

\author{%
  Daojun Liang \\
  School of Information Science \\ and Engineering, 
  Shandong University \\
  \texttt{liangdaojun@mail.sdu.edu.cn} 
  \And
  Haixia Zhang \\
  School of Control Science \\ and Engineering, 
  Shandong University \\
  \texttt{haixia.zhang@sdu.edu.cn} \\
  \AND
  Dongfeng Yuan \\
  School of Qilu Transportation,\\ Shandong University \\
  \texttt{dfyuan@sdu.edu.cn} \\
  \And
  Bingzheng Zhang \\
  School of Control Science \\ and Engineering, 
  Shandong University \\
  \texttt{bzzhang@mail.sdu.edu.cn} \\
}

\begin{document}

\maketitle

\begin{abstract}
  In this paper, we find that ubiquitous time series (TS) forecasting models are prone to severe overfitting.
  To cope with this problem, we embrace a de-redundancy approach to progressively reinstate the intrinsic values of TS for future intervals.
  Specifically, we introduce a dual-stream and subtraction mechanism, which is a deep Boosting ensemble learning method. And the vanilla Transformer is renovated by reorienting the information aggregation mechanism from addition to subtraction. 
  Then, we incorporate an auxiliary output branch into each block of the original model to construct a highway leading to the ultimate prediction. The output of subsequent modules in this branch will subtract the previously learned results, enabling the model to learn the residuals of the supervision signal, layer by layer.
  This designing facilitates the learning-driven implicit progressive decomposition of the input and output streams, empowering the model with heightened versatility, interpretability, and resilience against overfitting.
  Since all aggregations in the model are minus signs, which is called Minusformer.
  Extensive experiments demonstrate the proposed method outperform existing state-of-the-art methods, yielding an average performance improvement of 11.9\% across various datasets. 
  Experimental results can be reproduced through \href{https://github.com/Anoise/Minusformer}{Github}.
\end{abstract}

\section{Introduction}
\label{sec_intro}

\textit{``The sculpture is already complete within the marble block, before I start my work. It is already there. I just have to chisel away the superfluous material.''}
\quad --- Michelangelo

In this paper, we leverage the concept of de-redundancy to propose a progressive learning approach aimed at systematically acquiring the components of the supervision signal, thereby enhancing the performance of time series (TS) forecasting.
Before officially launching, let us scrutinize the conventional methodologies in TS forecasting.

TS recorded from the real world tend to exhibit myriad forms of non-stationarity due to their evolution under complex transient conditions \citep{1976_timeseries}.
The characteristics of non-stationary TS are reflected in continuously changing statistical properties and joint distributions \citep{cheng2015time}, which makes accurate prediction extremely difficult \citep{hyndman2018forecasting}.
Classical methods such as ARIMA \citep{piccolo1990distance}, exponential smoothing \citep{gardner1985exponential} and Kalman filter \citep{li2010parsimonious} are based on the stationarity assumption or statistical properties of time series to predict the future missing values, which are no longer suitable for non-stationary situation \citep{de200625}.

Recently, deep learning are introduced for TS forecasting due to its powerful nonlinear fitting capabilities \citep{hornik1991approximation}, including Attention-based long-term forecasting \citep{Zhou2021Informer, nie2022time, liu2023itransformer, shabani2022scaleformer} or Graph Neural Networks (GNNs) based forecasting methods \citep{li2018diffusion, wu2019graph}.
However, the latest studies suggest that the improvements in predictive performance using Attention-based methods, compared to Multi-Layer  Perceptrons (MLP), have not been significant \citep{zeng2023transformers, liang2023does}. And their inference speed has slowed down relative to the vanilla Transformer \citep{liang2023does}.
In addition, GNN-based methods have not shown substantial improvement in predictive performance compared to MLP \citep{shao2022spatial}. 

\begin{wrapfigure}{r}{0.5\columnwidth}
  \vspace{-25pt}
  \includegraphics[width=0.5\textwidth]{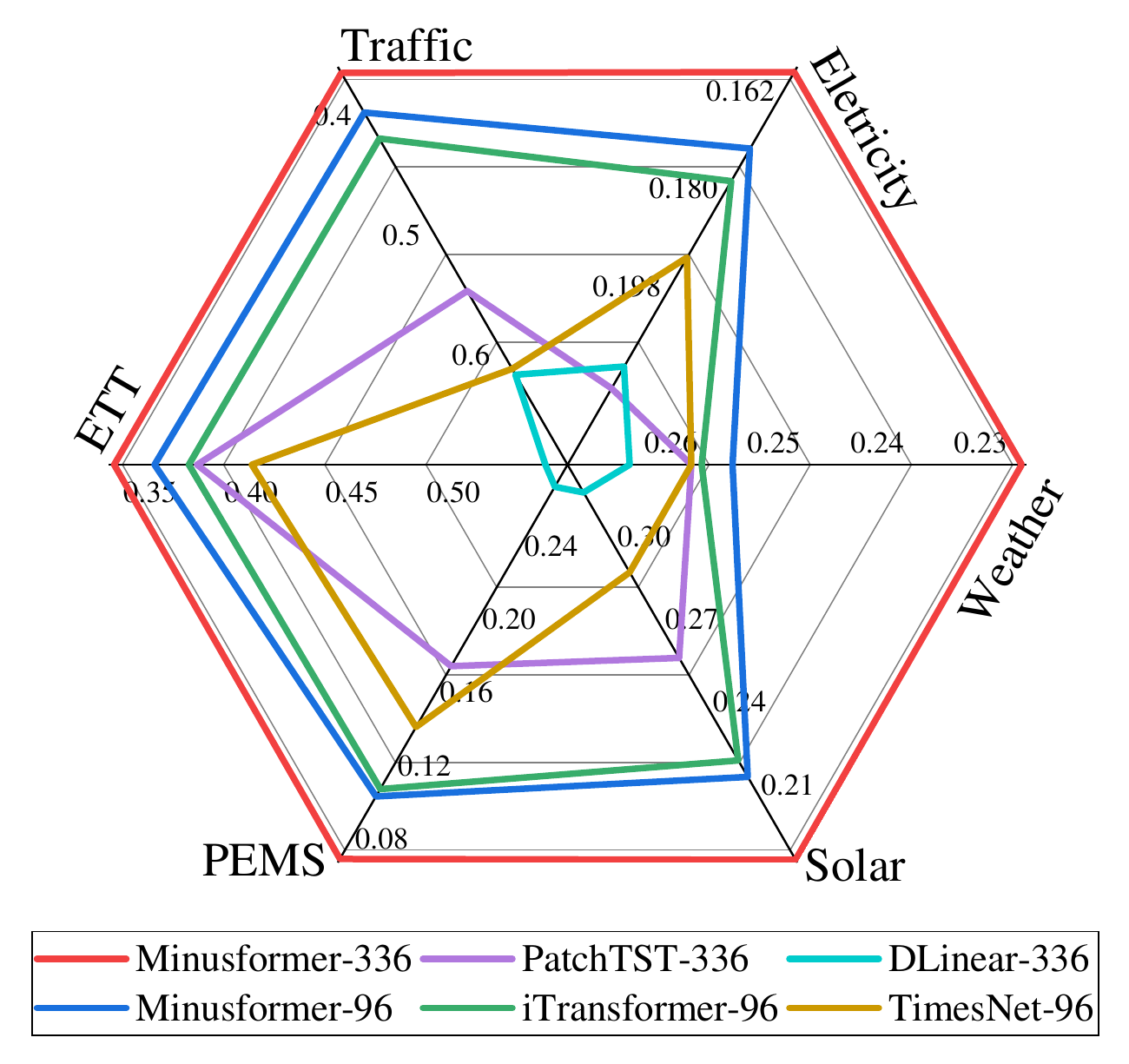}
  \caption{Comparison of the proposed Minusformer and other latest advanced models. The results (MSE) are averaged across all prediction lengths. The numerical suffix after the model indicates the input length of the model.
  Minusformer is configured with two versions of input length in order to align with other models.}
  \label{fig1} 
  \vspace{-15pt} 
\end{wrapfigure}
\begin{figure}[t]
  \centerline{\includegraphics[width=0.95\columnwidth]{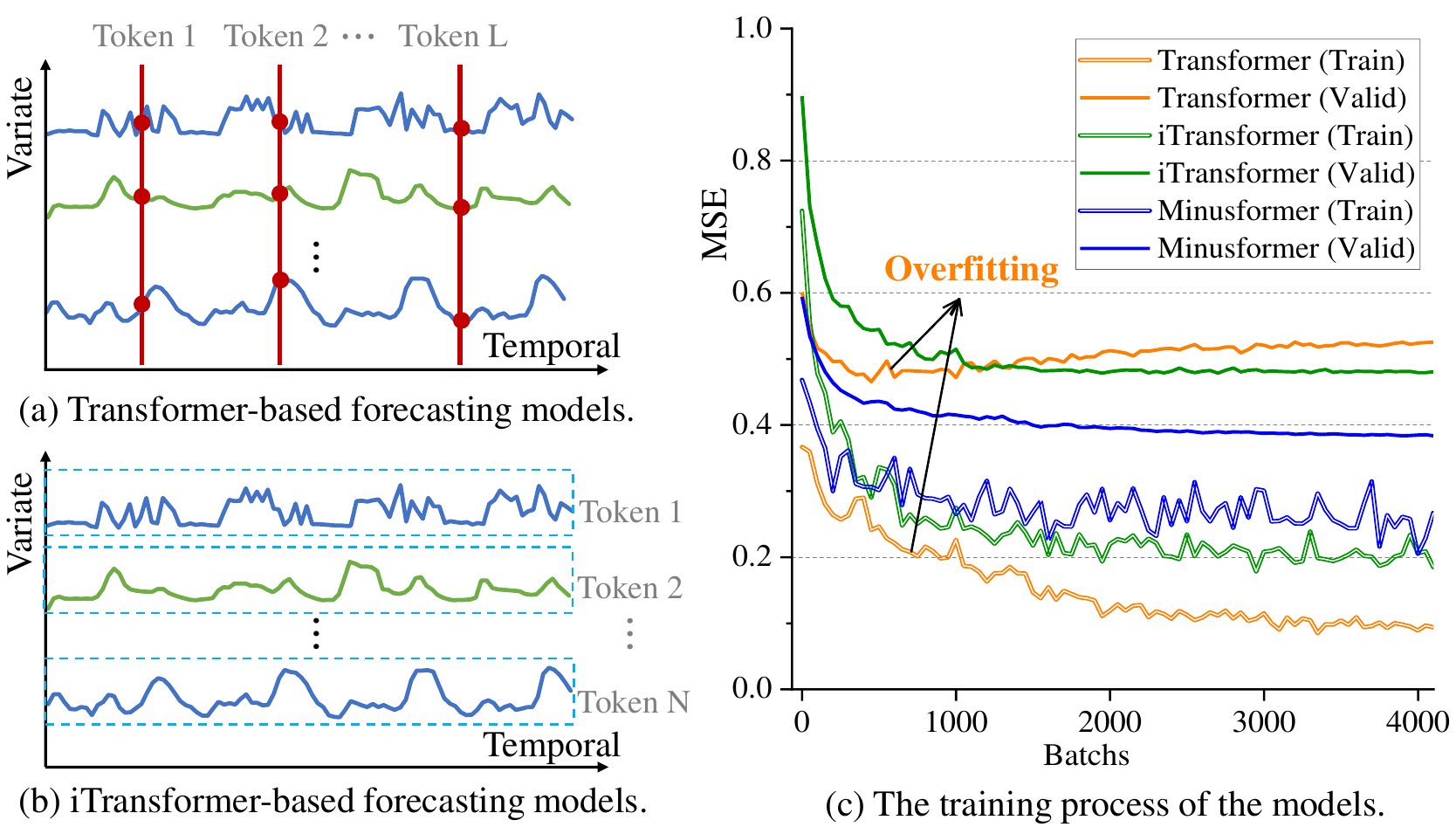}}
  \caption{Generalization of the model when time series are aggregated in different directions. The experiment was conducted utilizing Transformer with 4 blocks (baseline) on the Traffic dataset.}
  \label{fig_temp}
\end{figure}

Inspired by previous works, we found that prevalent deep models, e.g., Transformer-based models, are prone to severe overfitting on TS data. 
As shown in Fig. \ref{fig_temp}, overfitting occurs early during training (validation loss increases significantly), even though the training loss is still declining sharply (orange line).
Despite numerous methods to embed multivariate attributes into tokens (Fig. \ref{fig_temp}.a) or embed individual series into temporal tokens (Fig. \ref{fig_temp}.b), overfitting persists.
Reorienting the aggregation direction to the temporal dimension, e.g., iTransformer-based models, offer a minor alleviation of overfitting, yet its impact is highly constrained (green line).
Hence, it is imperative to develop a targeted network structure specifically tailored to mitigate the overfitting issue inherent in TS forecasting.


In this paper, we delve into a de-redundancy approach that implicitly decomposes the supervision signals to progressively steer the learning process to cope with the overfitting problem.
Concretely, we renovate the vanilla Transformer architecture by modifying the information aggregation mechanism, replacing addition with subtraction.
Then, we incorporate an auxiliary output stream into each block, thus constructing a highway that guides towards the final prediction.
The output of subsequent modules in this stream will subtract the previously learned results, facilitating the model to progressively learn the residuals of the supervision signal, layer by layer.
The incorporation of a dual stems design promotes the learning-driven implicit progressive decomposition of both the inputs and labels, which is equivalent to the Boosting ensemble learning \citep{kearns1994cryptographic}, thereby empowering the model with enhanced versatility, interpretability, and resilience against overfitting.
Given that all aggregations consist of minus signs, this architecture is referred to as Minusformer.

Further, we provide the theoretical rationale behind the effectiveness of the subtraction-based model. 
We demonstrate that the subtraction in Minusformer can effectively reduce the variance of the model by progressively learning the residuals of the supervision signal, thereby mitigating the overfitting problem.
Finally, we validate the proposed method across a diverse range of real-world TS datasets spanning various domains.
Extensive experiments show tht the proposed method outperform existing SOTA methods, yielding an average performance improvement of 11.9\% across various datasets. 

\section{Method}
\label{sec_arch}

\subsection{Preliminaries}

The purpose of TS forecasting is to use the observed value of $I$ historical moments to predict the missing value of $O$ future moments, which can be denoted as Input-$I$-Predict-$O$. If the feature dimension of the series is denoted as $D$, its input data can be denoted as $X^t = \{s_1^t, \cdots, s_I^t | s_i^t \in \mathbb{R}^D  \}$, and its target label can be denoted as $Y^t = \{s_{I+1}^t, \cdots, s_{I+O}^t | s_{I+o}^t \in \mathbb{R}^D \}$, where $s_i^t$ is a subseries with dimension $d$ at the $t$-th moment. Then, we can predict $\hat{Y}^t$ by designing a model $\mathcal{F}$ given an input $X^t$, which can be expressed as: $\hat{Y}^t = \mathcal{F}(X^t)$. Therefore, it is crucial to choose an appropriate $\mathcal{F}$ to improve the performance of the model. For denotation simplicity, the superscript $t$ will be omitted if it does not cause ambiguity in the context.

\subsection{Subtraction alleviates overfitting}
\label{app_overfit}

The main reason for overfitting is that the model has low bias and high variance on the testset \citep{hastie2009elements}.
Currently, TS forecasting models, especially very deep ones, can contain millions of parameters. 
While skip connections help in training deeper networks by mitigating the vanishing gradient problem, the sheer number of parameters increases the model's complexity, which easily leads to overfitting when training on highly volatile TS datasets.

\begin{figure}[!h]
  \centering 
  \includegraphics[width=\textwidth]{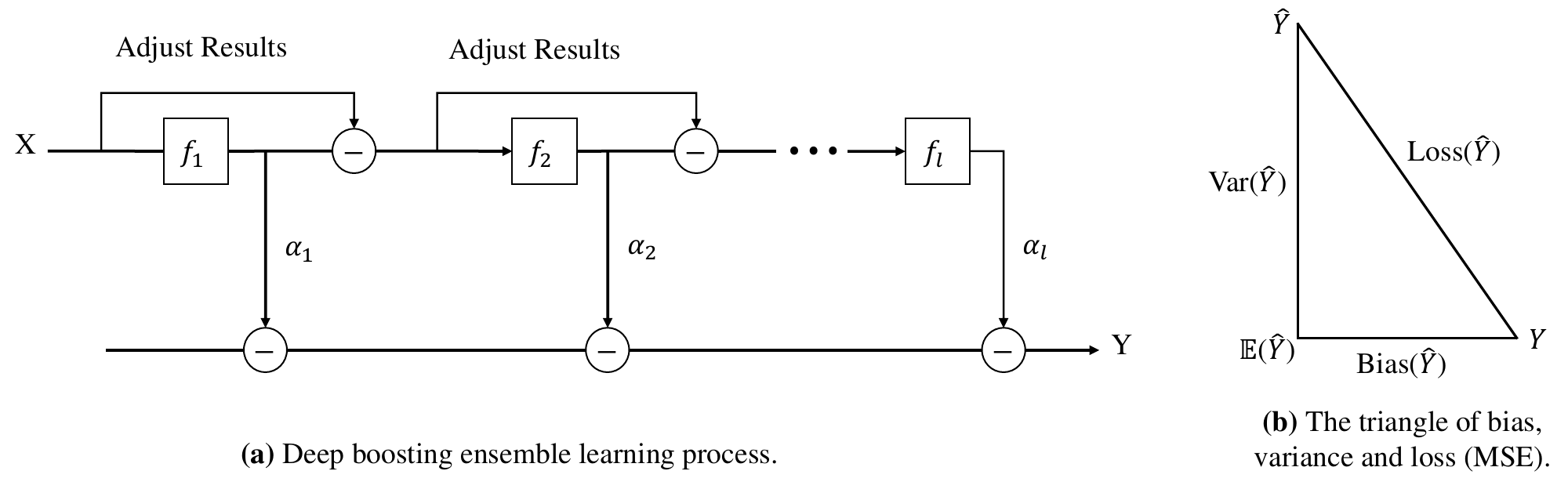}
  \caption{(a) Deep ensemble learning is equivalent to meta-algorithmic Boosting. (b) The relationship between model bias, variance, and loss.}
  \label{fig_boost} 
\end{figure}

We show that the subtraction operation is an implicit decomposition of the input and output streams, which is equivalent to meta-algorithmic Boosting, reducing the complexity of the model and thus mitigates the risk of overfitting. 
As shown in Fig. \ref{fig_boost}, the input stream is obviously a decomposition of $X$ because:
\begin{align}
  X & = \sum_{l=0}^{L-1} f_{l}(X) + R_L, \label{eq_a1} 
\end{align}
where $R_L$ is the residual term. Decomposition helps in identifying and understanding the underlying patterns within a time series.
By isolating and modeling the implicit components separately, one can improve the accuracy of future forecasts.
It is easier to model these components individually and then recombine them for prediction.

Further, the idea of output stems is to learn $L$ simple blcoks in a hierarchy where every block gives more attention (larger weight) to the hard samples by the previous block. This is equivalent to the Boosting ensemble learning process, where the final prediction is a weighted sum of $L$ simple blocks, and the weights are determined by the previous block.
Let $f_l(x)$ denote the $l$-th block in deep model, and $\alpha_l$ denote the weight of the $l$-th block. 
The overall estimation $\mathcal{F}(X)$ in deep model is a weighted subtraction of the $L$ estimations. For an sample $X$, we have:
  \begin{align}
    \hat{Y} & = \mathcal{F}(X) = i \sum_{l=0}^{\hbar} \alpha_{2l+1} f_{2l+1}(X) - i \sum_{l=0}^{\hbar} \alpha_{2l} f_{2l}(X), \label{eq_a2} 
  \end{align}
where $ i = 1$ if $L\bmod 2 = 1$, else $i = -1$, and $ \hbar = \lfloor \frac{L}{2} \rfloor$.

\subsection{Deep ensemble learning helps alleviate overfitting}
\label{sec3_overfit_theory}

Now, we provide a theoretical analysis of how deep ensemble learning alleviates overfitting.
In practice, observations $Y$ often contain additive noise $\varepsilon$: 
$Y  = \mathcal{Y} + \varepsilon $, 
where $\varepsilon \sim \mathcal{N}(0, \xi )$. 
Then, the estimation error (MSE) of the final model is:
\begin{equation}
  \underbrace{\text{Var}(\hat{Y}) + (\text{Bias}(\hat{Y}))^2 + \xi^2}_{\text{Test Error}}  =  \underbrace{\mathbb{E}[ (\hat{Y} - Y )^2] + 2\mathbb{E}(\varepsilon(\hat{Y} - \mathcal{Y} ))}_{\text{Training Error}}.
  \label{eq9}
\end{equation}
The proof is given in Appendix \ref{app_overfit_theory}.
Equation \ref{eq9} demonstrates that when a model exhibits low bias and high variance, it tends to indicate overfitting.
Otherwise, it behaves as underfitting. 
For modern complex deep learning models, their biases are typically very low \citep{goodfellow2016deep, zhang2021understanding}.
Therefore, we focus on how subtraction in deep model mitigates overfitting.
Specifically, for Equation \ref{eq_a2}, we have the following theorem.
\begin{theorem}
  Without loss of generality, assume that the estimation error of block $f_l(X)$ is $e_l$, $e_l \overset{i.i.d}{\sim}  \mathcal{N}(0, \nu)$. Let $\alpha_l = \alpha$ be the weight of $f_l$, $l\in [0,L] $, and the covariance of estimations of two different blocks by $\mu$, we have
  \begin{align}
    \text{Var}(\hat{Y}) < \frac{4}{L} \alpha^2 (\nu + \mu). \label{eq10}
  \end{align}
\label{th1}
\vspace{-1em}
\end{theorem}
The proof is given in Appendix \ref{seca_proof_th1}.
Clearly, the variance of the deep ensemble models is bound by the estimation error (noise error) of each block, the covariance between blocks.
It is evident that the subtraction adopted in deep ensemble models can reduce the variance, thereby mitigating overfitting. On the contrary, switching the aggregation operation of the output stream in the Minsformer to addition results in an approximate variance of $\frac{4}{L} \alpha^2 \nu + 3\alpha^2 \mu $, which is much larger than the that of subtraction in Equation \ref{eq_a2}.
Furthermore, Theorem \ref{th1} also demonstrates that increasing the number of layers $L$ does not escalate the risk of overfitting, which prove that the deep ensemble models can go deeper.

\subsection{Minusformer}
\label{sec_pf_arch}

\begin{wrapfigure}{r}{0.5\columnwidth}
  \vspace{-40pt}
  \includegraphics[width=0.5\textwidth]{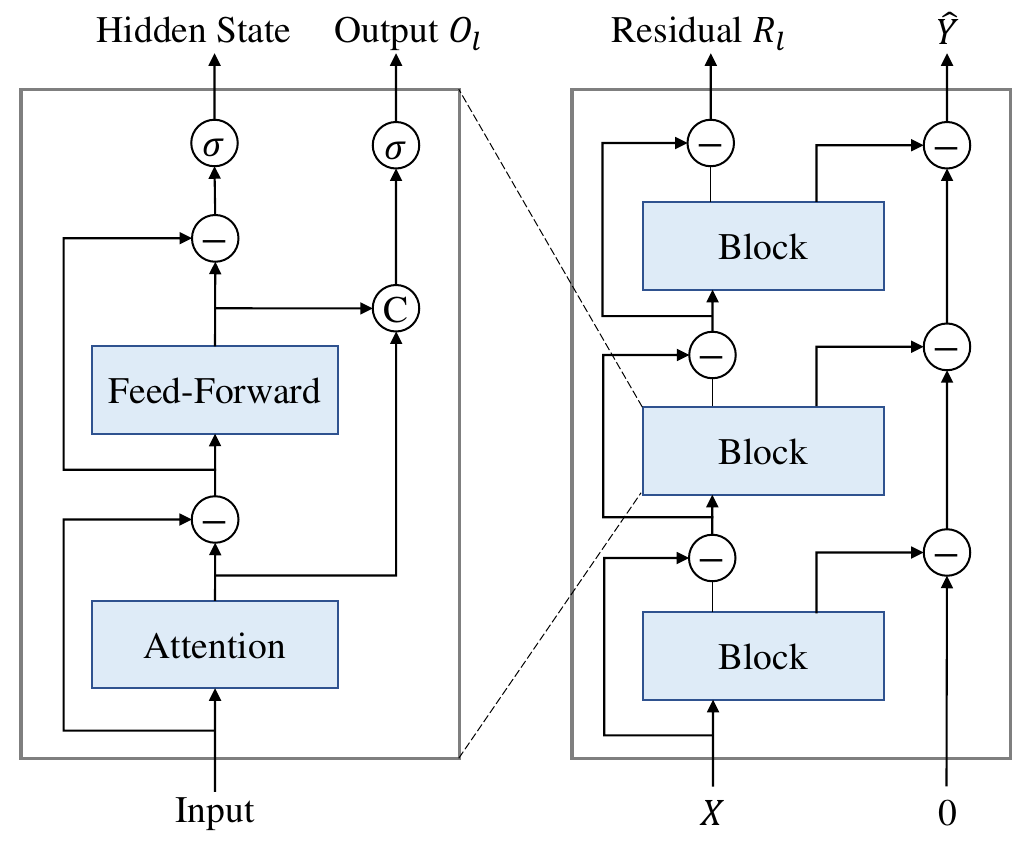}
  \caption{The architecture of Minusformer.}
  \label{fig_arch}
  \vspace{-25pt}
\end{wrapfigure}
As shown in Fig. \ref{fig_arch}, Minusformer is designed inspired by deep ensemble learning, which is comprises two primary data streams. One is the input stream decomposed through multiple residual blocks with subtraction operations, while the other is the output stream that progressively learning the residuals of the supervised signals. 
Along the way, they pass through multiple neural blocks capable of extracting and converting signals.
The base architecture is simple and versatile, yet powerful and interpretable. 
We now delve into how these properties are incorporated into the proposed architecture.
The pseudocode is given in Appendix \ref{sec_algo}.

\textbf{Backbone}: The fundamental building backbone features a fork architecture, which accepts one input $X_l$ and produces two distinct streams, $R_l$ and $O_l$. 
Concretely, $R_l$ is the remaining portion of $X_l$ after it has undergone processing within a neural module, which can be expressed as
\begin{subequations}{\label{eq2}}
  \begin{align}
    \hat{X}_l & = \text{Block} (X_l), \label{eq2a}\\
    R_l & = X_l - \hat{X}_l, \label{eq2b}
  \end{align}
  \end{subequations}
Equation \ref{eq2} represents an implicit decomposition of $X$, which differs from the moving average adopted by \citep{wu2021autoformer, zhou2022fedformer} and \citep{liang2023does} but is similar to \citep{oreshkin2019n}.
The residual information $R_l$ captures what remains unaltered or unprocessed, providing a basis for comparison with the transformed portion. 

In the subsequent steps, our intention is to maximize the utilization of the subtracted portion $\hat{X}_l$. First, $\hat{X}_l$ is projected into the same dimension as the label that is anticipated, $Y$. 
This process can be expressed as
\begin{align}
  O_l & = \text{Linear} (\hat{X}_l), \label{eq3}
\end{align}
where $O_l$ is the prediction results of the $l$-th predictor.
Then, $O_l$ will be subtracted from the outputs of the next predictor sequentially until the final predictions $\hat{Y}$ is achieved.
This iterative subtraction process is crucial because it facilitates the model in refining its understanding incrementally, with the objective of converging towards more accurate predictions as the layers go deeper.

\textbf{Block}: 
Exclusively focusing on learning the variate aspects of TS can result in substantial overfitting, as shown in Fig. \ref{fig_temp}. 
Conversely, exclusively prioritizing the learning of the temporal aspect of TS will hinder the capacity of the model to capture nuanced patterns and relationships within the attributes.
A ready-made solution is to utilize Attention mechanisms to learn subtle relationships between attributes. 
Due to the intrinsic sparsity of softmax activation, the model can incorporate potential attribute patterns without succumbing to overfitting.

The drawback of using Attention is that it becomes ineffective or even worsens  predictive performance when various attributes are independent of each other.
To mitigate this limitation, we implement a corrective measure by subtracting the Attention output from the input. This ensures that Attention can effectively capitalize on its inherent advantages, enhancing overall performance.
This process can be expressed as
\begin{subequations}{\label{eq4}}
  \begin{align}
    \hat{X}_{l,1} & =  \text{Attention} (X_{l,1}), \label{eq4a}\\
    R_{l,1} & = X_{l,1} - \delta \hat{X}_{l,1}. \label{eq4b}
  \end{align}
\end{subequations}
where $\delta$ is the Dirac function. $\delta$ serves to eliminate the Attention layer when it exerts adverse effects, thereby enabling the unimpeded flow of input towards the feedforward layers. 


Presently, within the entire block, there are two streams: one consists of the outputs $\hat{X}_l$ transformed by the neural modules, and the other encompasses the residuals $R_{l}$ obtained by subtracting $\hat{X}_l$ from the input.
Upon passing through a gate mechanism, they are directed toward the next block or projected into the output space.

\textbf{Gate}: Drawing inspiration from RNNs, our aspiration is for each neural module to autonomously regulate the pace of information transmission, akin to the inherent control exhibited by cells in RNNs.
Therefore, we introduced a gate mechanism at the conclusion of each block for both streams.
For residual stream, its gate mechanism can be expressed as
\begin{align}
  X_{l+1} & = \sigma (\theta_1(R_{l,2})) \cdot \theta_2(R_{l,2}), \label{eq6}
\end{align}
where $\sigma$ is the sigmoid function, $\theta_1$ and $\theta_2$ are learnable neurons with different parameters.
Likewise, for the intermediate $\hat{X}_l$, the gate mechanism can be expressed as
\begin{align}
  {O}_{l+1} & = \sigma (\theta_3([\hat{X}_{l,1},\hat{X}_{l,2}])) \cdot \theta_4([\hat{X}_{l,1},\hat{X}_{l,2}]), \label{eq7}
\end{align}
where the square brackets `[ ]' indicate concatenation operation.
Equation \ref{eq7} facilitates the comprehensive utilization of outputs from both the Attention and feedforward layers.

\begin{table*}[!ht]
    \centering
    \caption{Multivariate TS forecasting results on six benchmark datasets.}
    \label{tb2}
    \resizebox{\textwidth}{!}
    {
      \begin{threeparttable}
        \begin{tabular}{ccc >{\columncolor{gray!15}} c >{\columncolor{green!10}} cc >{\columncolor{gray!15}} c >{\columncolor{green!10}} cc >{\columncolor{gray!15}} cc >{\columncolor{gray!15}} cc >{\columncolor{gray!15}} cc >{\columncolor{gray!15}} cc >{\columncolor{gray!15}} cc >{\columncolor{gray!15}} cc >{\columncolor{gray!15}} cc >{\columncolor{gray!15}} cc >{\columncolor{gray!15}} c}
        \toprule
        \multicolumn{2}{c}{Model}          & \multicolumn{3}{c}{Minusformer-336} & \multicolumn{3}{c}{Minusformer-96} & \multicolumn{2}{c}{iTransformer-96} & \multicolumn{2}{c}{PatchTST-336} & \multicolumn{2}{c}{Crossformer-720$^{\diamondsuit}$} & \multicolumn{2}{c}{SCINet-168} & \multicolumn{2}{c}{TimesNet-96} & \multicolumn{2}{c}{DLinear-336} & \multicolumn{2}{c}{FEDformer-96} & \multicolumn{2}{c}{Autoformer-96} & \multicolumn{2}{c}{Informer-96} \\ \toprule
                                  & Length  & MSE       & MAE       & IMP         & MSE       & MAE       & IMP        & MSE              & MAE              & MSE             & MAE            & MSE               & MAE              & MSE            & MAE           & MSE            & MAE            & MSE            & MAE            & MSE               & MAE          & MSE             & MAE             & MSE           & MAE             \\ \toprule
        \multirow{5}{*}{\rotatebox{90}{ETT}}         & 96  & {\bf\color{red} 0.280}      & {\bf\color{red}0.336}     & 4.72\%      & \underline{\color{blue} 0.289}     & \underline{\color{blue}0.340}     & 2.64\%     & 0.297            & 0.349            & 0.302           & 0.348          & 0.745             & 0.584            & 0.707          & 0.621         & 0.340          & 0.374          & 0.333          & 0.387          & 0.358             & 0.397        & 0.346           & 0.388           & 3.755         & 1.525           \\
                                      & 192 & {\bf\color{red} 0.332}     & {\bf\color{red}0.371}     & 9.94\%      & \underline{\color{blue}0.349}     & \underline{\color{blue}0.374}     & 7.33\%     & 0.380            & 0.400            & 0.388           & 0.400          & 0.877             & 0.656            & 0.860          & 0.689         & 0.402          & 0.414          & 0.477          & 0.476          & 0.429             & 0.439        & 0.456           & 0.452           & 5.602         & 1.931           \\
                                      & 336 & {\bf\color{red} 0.362}     & {\bf\color{red}0.392}     & 12.34\%     & \underline{\color{blue}0.393}     & \underline{\color{blue}0.405}     & 7.21\%     & 0.428            & 0.432            & 0.426           & 0.433          & 1.043             & 0.731            & 1.000          & 0.744         & 0.452          & 0.452          & 0.594          & 0.541          & 0.496             & 0.487        & 0.482           & 0.486           & 4.721         & 1.835        \\
                                      & 720 & {\bf\color{red} 0.410}     & {\bf\color{red}0.429}    & 3.79\%      & 0.433     & \underline{\color{blue}0.435}     & 0.42\%     & \underline{\color{blue}0.427}            & 0.445            & 0.431           & 0.446          & 1.104             & 0.763            & 1.249          & 0.838         & 0.462          & 0.468          & 0.831          & 0.657          & 0.463             & 0.474        & 0.515           & 0.511           & 3.647         & 1.625        \\ \cline{2-26}
       \rowcolor{cyan!15}  \cellcolor{white}  & Avg & {\bf\color{red}0.346}     & {\bf\color{red}0.382}     & \cellcolor{green!30} 7.90\%      & \underline{\color{blue}0.366}     & \underline{\color{blue}0.388}     & \cellcolor{green!30} 4.55\%     & 0.383            & 0.407            & 0.387           & 0.407          & 0.942             & 0.684            & 0.954          & 0.723         & 0.414          & 0.427          & 0.559          & 0.515          & 0.437             & 0.449        & 0.450           & 0.459           & 4.431         & 1.729           \\ \toprule
        \multirow{5}{*}{\rotatebox{90}{Traffic}}     & 96  & {\bf\color{red}0.350}     & {\bf\color{red}0.250}     & 9.05\%      & \underline{\color{blue}0.386}     & \underline{\color{blue}0.258}     & 3.00\%     & 0.395            & 0.268            & 0.544           & 0.359          & 0.522             & 0.290            & 0.788          & 0.499         & 0.593          & 0.321          & 0.650          & 0.396          & 0.587             & 0.366        & 0.613           & 0.388           & 0.719         & 0.391           \\
                                      & 192 & {\bf\color{red}0.375}     & {\bf\color{red}0.261}     & 7.75\%      & \underline{\color{blue}0.398}     & \underline{\color{blue}0.263}     & 4.63\%     & 0.417            & 0.276            & 0.540           & 0.354          & 0.530             & 0.293            & 0.789          & 0.505         & 0.617          & 0.336          & 0.598          & 0.370          & 0.604             & 0.373        & 0.616           & 0.382           & 0.696         & 0.379           \\
                                      & 336 & {\bf\color{red}0.381}     & {\bf\color{red}0.264}     & 9.36\%      & \underline{\color{blue}0.409}     & \underline{\color{blue}0.270}      & 5.07\%     & 0.433            & 0.283            & 0.551           & 0.358          & 0.558             & 0.305            & 0.797          & 0.508         & 0.629          & 0.336          & 0.605          & 0.373          & 0.621             & 0.383        & 0.622           & 0.337           & 0.777         & 0.420           \\
                                      & 720 & {\bf\color{red}0.386}     & {\bf\color{red}0.268}     & 14.30\%     & \underline{\color{blue}0.431}     & \underline{\color{blue}0.287}     & 6.34\%     & 0.467            & 0.302            & 0.586           & 0.375          & 0.589             & 0.328            & 0.841          & 0.523         & 0.640          & 0.350          & 0.645          & 0.394          & 0.626             & 0.382        & 0.660           & 0.408           & 0.864         & 0.472           \\ \cline{2-26}
        \rowcolor{cyan!15}  \cellcolor{white} & Avg & {\bf\color{red}0.373}     & {\bf\color{red}0.261}     & \cellcolor{green!30} 10.15\%     & \underline{\color{blue}0.406}     & \underline{\color{blue}0.270}    & \cellcolor{green!30} 4.79\%     & 0.428            & 0.282            & 0.555           & 0.362          & 0.550             & 0.304            & 0.804          & 0.509         & 0.620          & 0.336          & 0.625          & 0.383          & 0.610             & 0.376        & 0.628           & 0.379           & 0.764         & 0.416           \\ \toprule
        \multirow{5}{*}{\rotatebox{90}{Electricity}} & 96  & {\bf\color{red}0.128}     & {\bf\color{red}0.223}     & 10.30\%     & \underline{\color{blue}0.143}     & \underline{\color{blue}0.235}     & 2.73\%     & 0.148            & 0.240            & 0.195           & 0.285          & 0.219             & 0.314            & 0.247          & 0.345         & 0.168          & 0.272          & 0.197          & 0.282          & 0.193             & 0.308        & 0.201           & 0.317           & 0.274         & 0.368           \\
                                      & 192 & {\bf\color{red}0.148}     & {\bf\color{red}0.24}      & 6.89\%      & \underline{\color{blue}0.162}     & \underline{\color{blue}0.253}     & 0.00\%     & 0.162            & 0.253            & 0.199           & 0.289          & 0.231             & 0.322            & 0.257          & 0.355         & 0.184          & 0.289          & 0.196          & 0.285          & 0.201             & 0.315        & 0.222           & 0.334           & 0.296         & 0.386           \\
                                      & 336 & {\bf\color{red}0.164}     & {\bf\color{red}0.26}      & 5.61\%      & 0.179     & 0.271     & -0.65\%    & \underline{\color{blue}0.178}            & \underline{\color{blue}0.269}            & 0.215           & 0.305          & 0.246             & 0.337            & 0.269          & 0.369         & 0.198          & 0.300          & 0.209          & 0.301          & 0.214             & 0.329        & 0.231           & 0.338           & 0.300         & 0.394           \\
                                      & 720 & {\bf\color{red}0.192}     & {\bf\color{red}0.284}     & 12.54\%     & \underline{\color{blue}0.204}     & \underline{\color{blue}0.294}     & 8.29\%     & 0.225            & 0.317            & 0.256           & 0.337          & 0.280             & 0.363            & 0.299          & 0.390         & 0.220          & 0.320          & 0.245          & 0.333          & 0.246             & 0.355        & 0.254           & 0.361           & 0.373         & 0.439           \\ \cline{2-26}
        \rowcolor{cyan!15}  \cellcolor{white}   & Avg & {\bf\color{red}0.158}     & {\bf\color{red}0.252}     & \cellcolor{green!30} 8.95\%      & \underline{\color{blue}0.172}     & \underline{\color{blue}0.263}     & \cellcolor{green!30} 2.94\%     & 0.178            & 0.270            & 0.216           & 0.304          & 0.244             & 0.334            & 0.268          & 0.365         & 0.192          & 0.295          & 0.212          & 0.300          & 0.214             & 0.327        & 0.227           & 0.338           & 0.311         & 0.397           \\ \toprule
        \multirow{5}{*}{\rotatebox{90}{Weather}}     & 96  & {\bf\color{red}0.150}     & {\bf\color{red}0.201}     & 9.93\%      & 0.169     & \underline{\color{blue}0.209}     & 2.61\%     & 0.174            & 0.214            & 0.177           & 0.218          & \underline{\color{blue}0.158}             & 0.230            & 0.221          & 0.306         & 0.172          & 0.220          & 0.196          & 0.255          & 0.217             & 0.296        & 0.266           & 0.336           & 0.300         & 0.384           \\
                                      & 192 & {\bf\color{red}0.194}     & {\bf\color{red}0.244}     & 8.08\%      & 0.220     & \underline{\color{blue}0.254}     & 0.23\%     & 0.221            & 0.254            & 0.225           & 0.259          & \underline{\color{blue}0.206}             & 0.277            & 0.261          & 0.340         & 0.219          & 0.261          & 0.237          & 0.296          & 0.276             & 0.336        & 0.307           & 0.367           & 0.598         & 0.544           \\
                                      & 336 & {\bf\color{red}0.245}     & {\bf\color{red}0.282}     & 8.30\%      & 0.276     & \underline{\color{blue}0.296}     & 0.36\%     & 0.278            & 0.296            & 0.278           & 0.297          & \underline{\color{blue}0.272}             & 0.335            & 0.309          & 0.378         & 0.280          & 0.306          & 0.283          & 0.335          & 0.339             & 0.380        & 0.359           & 0.395           & 0.578         & 0.523           \\
                                      & 720 & {\bf\color{red}0.320}     & {\bf\color{red}0.336}     & 7.17\%      & 0.354     & \underline{\color{blue}0.346}     & 0.99\%     & 0.358            & 0.349            & 0.354           & 0.348          & 0.398             & 0.418            & 0.377          & 0.427         & 0.365          & 0.359          & \underline{\color{blue}0.345}          & 0.381          & 0.403             & 0.428        & 0.419           & 0.428           & 1.059         & 0.741           \\ \cline{2-26}
        \rowcolor{cyan!15}  \cellcolor{white}  & Avg & {\bf\color{red}0.227}     & {\bf\color{red}0.266}     & \cellcolor{green!30} 8.85\%      & \underline{\color{blue}0.255}     & \underline{\color{blue}0.276}     & \cellcolor{green!30} 1.66\%     & 0.258            & 0.279            & 0.259           & 0.281          & 0.259             & 0.315            & 0.292          & 0.363         & 0.259          & 0.287          & 0.265          & 0.317          & 0.309             & 0.360        & 0.338           & 0.382           & 0.634         & 0.548           \\ \toprule
        \multirow{5}{*}{\rotatebox{90}{Solar}}       & 96  & {\bf\color{red}0.181}     & {\bf\color{red}0.222}     & 8.58\%      & \underline{\color{blue}0.192}     & \underline{\color{blue}0.222}     & 5.87\%     & 0.203            & 0.237            & 0.234           & 0.286          & 0.310             & 0.331            & 0.237          & 0.344         & 0.250          & 0.292          & 0.290          & 0.378          & 0.242             & 0.342        & 0.884           & 0.711           & 0.236         & 0.259           \\
                                      & 192 & {\bf\color{red}0.198}     & {\bf\color{red}0.239}     & 11.73\%     & 0.230      & \underline{\color{blue}0.251}     & 2.56\%     & 0.233            & 0.261            & 0.267           & 0.310          & 0.734             & 0.725            & 0.280          & 0.380         & 0.296          & 0.318          & 0.320          & 0.398          & 0.285             & 0.380        & 0.834           & 0.692           & \underline{\color{blue}0.217}         & 0.269           \\
                                      & 336 & {\bf\color{red}0.202}     & {\bf\color{red}0.245}     & 14.40\%     & \underline{\color{blue}0.243}     & \underline{\color{blue}0.263}     & 2.84\%     & 0.248            & 0.273            & 0.290           & 0.315          & 0.750             & 0.735            & 0.304          & 0.389         & 0.319          & 0.330          & 0.353          & 0.415          & 0.282             & 0.376        & 0.941           & 0.723           & 0.249         & 0.283           \\
                                      & 720 & {\bf\color{red}0.206}    & {\bf\color{red}0.252}     & 12.82\%     & 0.243     & \underline{\color{blue}0.265}     & 3.02\%     & 0.249            & 0.275            & 0.289           & 0.317          & 0.769             & 0.765            & 0.308          & 0.388         & 0.338          & 0.337          & 0.356          & 0.413          & 0.357             & 0.427        & 0.882           & 0.717           & \underline{\color{blue}0.241}         & 0.317           \\ \cline{2-26}
        \rowcolor{cyan!15}  \cellcolor{white} & Avg & {\bf\color{red}0.197}     & {\bf\color{red}0.240}     & \cellcolor{green!30} 11.92\%     & \underline{\color{blue}0.227}     & \underline{\color{blue}0.250}     & \cellcolor{green!30} 3.58\%     & 0.233            & 0.262            & 0.270           & 0.307          & 0.641             & 0.639            & 0.282          & 0.375         & 0.301          & 0.319          & 0.330          & 0.401          & 0.291             & 0.381        & 0.885           & 0.711           & 0.235         & 0.280           \\ \toprule
        \multirow{5}{*}{\rotatebox{90}{PEMS}}        & 12  & {\bf\color{red}0.057}     & {\bf\color{red}0.157}    & 14.74\%     & 0.067     & \underline{\color{blue}0.171}     & 3.68\%     & 0.071            & 0.174            & 0.099           & 0.216          & 0.090             & 0.203            & \underline{\color{blue}0.066}          & 0.172         & 0.085          & 0.192          & 0.122          & 0.243          & 0.126             & 0.251        & 0.272           & 0.385           & 0.126         & 0.233           \\
                                      & 24  & {\bf\color{red}0.070}     & {\bf\color{red}0.173}     & 19.33\%     & 0.093     & 0.203     & -0.50\%    & 0.093            & 0.201            & 0.142           & 0.259          & 0.121             & 0.240            & \underline{\color{blue}0.085}          & \underline{\color{blue}0.198}         & 0.118          & 0.223          & 0.201          & 0.317          & 0.149             & 0.275        & 0.334           & 0.440           & 0.139         & 0.250           \\
                                      & 36  & {\bf\color{red}0.083}     & {\bf\color{red}0.186}     & 27.39\%     & 0.125     & 0.237     & -0.21\%    & \underline{\color{blue}0.125}            & \underline{\color{blue}0.236}            & 0.211           & 0.319          & 0.202             & 0.317            & 0.127          & 0.238         & 0.155          & 0.260          & 0.333          & 0.425          & 0.227             & 0.348        & 1.032           & 0.782           & 0.186         & 0.289           \\
                                      & 48  & {\bf\color{red}0.091}     & {\bf\color{red}0.195}     & 35.45\%     & \underline{\color{blue}0.151}     & \underline{\color{blue}0.262}     & 4.29\%     & 0.160            & 0.270            & 0.269           & 0.370          & 0.262             & 0.367            & 0.178          & 0.287         & 0.228          & 0.317          & 0.457          & 0.515          & 0.348             & 0.434        & 1.031           & 0.796           & 0.233         & 0.323           \\  \cline{2-26}
        \rowcolor{cyan!15}  \cellcolor{white}  & Avg & {\bf\color{red}0.075}     & {\bf\color{red}0.178}     & \cellcolor{green!30} 26.54\%     & \underline{\color{blue}0.109}     & \underline{\color{blue}0.218}     & \cellcolor{green!30} 2.39\%     & 0.113            & 0.221            & 0.180           & 0.291          & 0.169             & 0.281            & 0.114          & 0.224         & 0.147          & 0.248          & 0.278          & 0.375          & 0.213             & 0.327        & 0.667           & 0.601           & 0.171         & 0.274           \\ \toprule
        \multicolumn{2}{c}{{\bf \color{red}$1^{st}$} or \underline{\color{blue}$2^{st}$} Count} & {\bf\color{red} 30}  & {\bf\color{red}30}   & {\bf\color{red}30}  & \underline{\color{blue}19}   & \underline{\color{blue}27}   & \underline{\color{blue}27}     & \underline{\color{blue}3}   & \underline{\color{blue}2}      & 0    & 0     & \underline{\color{blue}3}    & 0    & \underline{\color{blue}2}  & \underline{\color{blue}1}   & 0               & 0         & \underline{\color{blue}1}  & 0    & 0       & 0   & 0     & 0     & \underline{\color{blue}2}      & 0          \\  \bottomrule
        \end{tabular}
        \begin{tablenotes}
          \large
          \item[*] {\bf IMP} refers to the average performance improvement compared to the latest iTransformer with the best average performance. ${\diamondsuit}$ denotes the maximum search range of the input length.
        \end{tablenotes}
      \end{threeparttable}
    }
  \end{table*}

\section{Experiments}
\label{sec_exp}

Minusformer undergoes a comprehensive evaluation across the widely employed real-world datasets, including multiple mainstream TS forecasting applications such as energy, traffic, electricity, weather, transportation and exchange.

\textbf{Implementation details}: 
The model undergoes training utilizing the ADAM optimizer \citep{kingma2014adam} and minimizing the Mean Squared Error (MSE) loss function.
The training process is halted prematurely, typically within 10 epochs.
The Minusformer architecture solely comprises the embedding layer and backbone architecture, devoid of any additional introduced hyperparameters.
Refer to Appendix \ref{app_hpyer} for the hyperparameter sensitivity analysis.
During model validation, two evaluation metrics are employed: MSE and Mean Absolute Error (MAE).
Given the potential competitive relationship between the two indicators, MSE and MAE, we use the average of the two ($\frac{MSE+MAE}{2}$) to evaluate the overall performance of the model.

\textbf{Baselines}: 
We employ recent 14 SOTA methods for comparisons, including iTransformer \citep{liu2023itransformer}, PatchTST \citep{nie2022time}, Crossformer \citep{zhang2022crossformer}, SCINet \citep{liu2022scinet}, TimesNet \citep{wu2022timesnet}, DLinear \citep{zeng2023transformers}, Periodformer \citep{liang2023does}, FEDformer \citep{zhou2022fedformer}, Autoformer \citep{wu2021autoformer}, Informer \citep{Zhou2021Informer}, LogTrans \citep{li2019enhancing} and Reformer \citep{Kitaev2020Reformer}.
In particular, competitive models such as N-BEATS \citep{oreshkin2019n} and N-Hits \citep{challu2023nhits} are also employed for comparing univariate forecasting.
The numerical suffix attached to each model represents the input length utilized by the respective model.

\subsection{Main experimental results}

All datasets are adopted for both multivariate (multivariate predict multivariate) and univariate (univariate predicts univariate) tasks. 
The detailed information pertaining to the datasets can be located in Appendix \ref{app_dataset}.
The models used in the experiments are evaluated over a wide range of prediction lengths to compare performance on different future horizons: 96, 192, 336 and 720. 
The experimental settings are the same for both multivariate and univariate tasks.
Please refer to Appendix \ref{app_ett} for more experiments on the full ETT dataset.

\textbf{Multivariate results}:
The results for multivariate TS forecasting are outlined in Table \ref{tb2}, with the optimal results highlighted in {\bf \color{red} red} and the second-best results emphasized with \underline{\color{blue} underlined}. 
Due to variations in input lengths among different methods, for instance, PatchTST and DLinear employing an input length of 336, while Crossformer and Periodformer search for the input length without surpassing the maximum setting (720 in Crossformer, 144 in Periodformer), we have configured two versions of the proposed Minusformer, each with different input lengths (96 and 336), for performance evaluation.

As shown in Table \ref{tb2}, the proposed Minusformer achieves the consistent SOTA performance across all datasets and prediction length configurations. 
iTransformer and PatchTST stand out as the latest models acknowledged for their exceptional average performance.
Compared with them, the proposed Minusformer-336 demonstrates an average performance increase of {\bf 11.9\%} and {\bf 20.6\%}, respectively, achieving a substantial performance improvement.
Next, our primary focus shifts to analyzing the performance gains of Minusformer-96.
Minusformer-96 demonstrates an average performance increase of {\bf 3.0\%} and {\bf 13.4\%}, respectively.
It achieves advanced performance, averaging {\bf 23} items across six datasets, with an improvement (IMP) in average performance on each dataset. 
These experimental results confirm that the proposed Minusformer demonstrates superior prediction performance across different datasets with varying horizons. 

\begin{table*}[htpb]
  \centering
  \caption{Univariate TS forecasting results on five benchmark datasets. }
  \label{tb3}
  \resizebox{0.95\textwidth}{!}
  {
    \tiny
    \begin{threeparttable}
    \begin{tabular}{cc  cc  cc  cc  cc  cc  cc  c}
      \toprule
      Model       & \multicolumn{2}{l}{Minusformer-96} & \multicolumn{2}{l}{Periodformer-144$^{\diamondsuit}$} & \multicolumn{2}{l}{FEDformer-96} & \multicolumn{2}{l}{Autoformer-96} & \multicolumn{2}{l}{Informer-96} & \multicolumn{2}{l}{LogTrans-96} & \multicolumn{2}{l}{Reformer-96} \\
      \hline
      Metric      & MSE              & MAE             & MSE               & MAE               & MSE             & MAE            & MSE             & MAE             & MSE            & MAE            & MSE            & MAE            & MSE            & MAE            \\
      \hline
      ETTh1       & {\bf\color{red}0.072}            & {\bf\color{red}0.206}           & \underline{\color{blue}0.093}             & \underline{\color{blue}0.237}             & 0.111           & 0.257          & 0.105           & 0.252           & 0.199          & 0.377          & 0.345          & 0.513          & 0.624          & 0.600          \\
      ETTh2       & {\bf\color{red}0.185}            & {\bf\color{red}0.337}           & \underline{\color{blue}0.192}             & \underline{\color{blue}0.343}             & 0.206           & 0.350          & 0.218           & 0.364           & 0.243          & 0.400          & 0.252          & 0.408          & 3.472          & 1.283          \\
      ETTm1       & {\bf\color{red}0.052}            & {\bf\color{red}0.172}           & \underline{\color{blue}0.059}             & \underline{\color{blue}0.201}             & 0.069           & 0.202          & 0.081           & 0.221           & 0.281          & 0.441          & 0.231          & 0.382          & 0.523          & 0.536          \\
      ETTm2       & \underline{\color{blue}0.118}            & \underline{\color{blue}0.254}           & {\bf\color{red}0.115}             & {\bf\color{red}0.253 }            & 0.135           & 0.278          & 0.130           & 0.271           & 0.147          & 0.293          & 0.130          & 0.277          & 0.136          & 0.288          \\
      Traffic     & {\bf\color{red}0.132}            & {\bf\color{red}0.212}           & \underline{\color{blue}0.150}             & \underline{\color{blue}0.233}             & 0.177           & 0.270          & 0.261           & 0.365           & 0.309          & 0.388          & 0.341          & 0.417          & 0.375          & 0.434          \\
      Electricity & \underline{\color{blue}0.314}            & \underline{\color{blue}0.401}           & {\bf\color{red}0.298 }            & {\bf\color{red}0.389}             & 0.347           & 0.434          & 0.414           & 0.479           & 0.372          & 0.444          & 0.410          & 0.473          & 0.352          & 0.435          \\
      Weather     & {\bf\color{red}0.0015}           & {\bf\color{red}0.0293}          & \underline{\color{blue}0.0017}            & \underline{\color{blue}0.0317}            & 0.008           & 0.067          & 0.0083          & 0.0700          & 0.0033         & 0.0438         & 0.0059         & 0.0563         & 0.0115         & 0.0785         \\
      Exchange    & \underline{\color{blue}0.429}            & \underline{\color{blue}0.453}           & {\bf\color{red}0.353}             & {\bf\color{red}0.434}             & 0.499           & 0.512          & 0.578           & 0.537           & 1.511          & 1.029          & 1.350          & 0.810          & 1.028          & 0.812          \\ 
      \hline
      { $1^{\text{st}}$ Count} & {\bf\color{red} 24}           & {\bf\color{red}27}           & \underline{\color{blue}14}                   & \underline{\color{blue}12}       & 0               & 1               & 0             & 0               & 0               & 0               & 2             & 0             & 0               & 0             \\
      \bottomrule
    \end{tabular}
  \begin{tablenotes}
    \item[*] All results are averaged across all prediction lengths. The results for all prediction lengths are provided in Appendix \ref{app_uts}. ${\diamondsuit}$ denotes the maximum search range of the input length.
  \end{tablenotes}
  \end{threeparttable}
}
\vspace{-1em}
\end{table*}

\begin{table*}[!ht]
    \centering
    \caption{Mulvariate and univariate forecasting results with diverse metrics on Monash TS datasets.}
    \label{tb_metric_1}
    \resizebox{\textwidth}{!}
    {
    \Huge
    \begin{threeparttable}
    \begin{tabular}{c|ccccccc|ccccccc|ccccccc|cccccc}
      \toprule
      Mulvariate           & \multicolumn{7}{c|}{ILI}                                                        & \multicolumn{7}{c|}{Oik\_Weather}                                                                                               & \multicolumn{7}{c|}{NN5}                                                                                & \multicolumn{6}{c}{Rideshare}                                          \\ \toprule
    Metric          & MSE   & MAE   & RMSP & MAPE                          & sMAPE & MASE  & Q75   & MSE                           & MAE   & RMSP & MAPE                          & sMAPE & MASE                          & Q75   & MSE                           & MAE   & RMSP & MAPE                          & sMAPE & MASE  & Q75   & MSE   & MAE   & RMSP & sMAPE & MASE                          & Q75   \\ \toprule
    Minsuforemr  & {\bf\color{red}2.016} & {\bf\color{red}0.86}  & {\bf\color{red}0.367}  & 3.206 & {\bf\color{red}0.650}  & {\bf\color{red}0.574} & {\bf\color{red}0.955} & \underline{\color{blue}0.689} & {\bf\color{red}0.618} & {\bf\color{red}0.773}  & {\bf\color{red}4.647} & {\bf\color{red}1.082} & \underline{\color{blue}0.888} & {\bf\color{red}0.628} & {\bf\color{red}0.719} & {\bf\color{red}0.578} & {\bf\color{red}0.541}  & {\bf\color{red}2.438}  & {\bf\color{red}0.861} & {\bf\color{red}0.530}  & {\bf\color{red}0.559} & {\bf\color{red}0.364} & {\bf\color{red}0.357} & {\bf\color{red}0.233}  & {\bf\color{red}0.509} & 1.297 & {\bf\color{red}0.479} \\ 
    iTransformer & \underline{\color{blue}2.262} & \underline{\color{blue}0.958} & \underline{\color{blue}0.415}  & \underline{\color{blue}3.111}                         & \underline{\color{blue}0.703} & \underline{\color{blue}0.648} & \underline{\color{blue}1.107} & 0.720                          & 0.632 & \underline{\color{blue}0.781}  & 7.986                         & \underline{\color{blue}1.090}  & 0.935                         & \underline{\color{blue}0.636} & \underline{\color{blue}0.723}               & \underline{\color{blue}0.583} & \underline{\color{blue}0.554}  &  \underline{\color{blue}2.574}              & \underline{\color{blue}0.877} & \underline{\color{blue}0.542} & \underline{\color{blue}0.565} & \underline{\color{blue}0.496} & \underline{\color{blue}0.414} & \underline{\color{blue}0.264}  & \underline{\color{blue}0.575} & \underline{\color{blue}1.258}                         & \underline{\color{blue}0.573} \\
    DLinear      & 2.915 & 1.188 & 0.574  & 3.113                         & 0.865 & 0.791 & 1.524 & 0.701                         & 0.637 & 0.805  & 4.662               & 1.185 & 0.943                         & 0.641 & 1.448                         & 0.929 & 0.969  & 3.810                          & 1.304 & 0.835 & 0.978 & 1.028 & 0.815 & 0.840   & 1.283 & {\bf\color{red}1.049}                & 1.197 \\
    Autoformer   & 3.187 & 1.224 & 0.596  & 4.595                         & 0.909 & 0.740  & 1.349 & 0.853                         & 0.712 & 0.883  & 9.726                         & 1.201 & 0.992                         & 0.704 & 0.851                         & 0.659 & 0.636  & 2.760                          & 0.975 & 0.603 & 0.637 & 0.609 & 0.575 & 0.544  & 0.870  & 1.082                         & 0.816 \\
    Informer     & 5.208 & 1.576 & 0.792  & {\bf\color{red}2.494}                & 1.183 & 0.907 & 2.304 & {\bf\color{red}0.67}                 & \underline{\color{blue}0.625} & 0.795  & \underline{\color{blue}7.242}                         & 1.126 & {\bf\color{red}0.818}                & 0.640  & 0.967                         & 0.727 & 0.739  & 2.855                         & 1.081 & 0.662 & 0.722 & 0.612 & 0.506 & 0.365  & 0.699 & 1.072                         & 0.708 \\ \bottomrule
    
    Univariate        & \multicolumn{7}{c|}{M4 Hourly}                                                                                               & \multicolumn{7}{c|}{Us\_births}                                                                                       & \multicolumn{7}{c|}{Saugeenday}                                                                                       & \multicolumn{6}{c}{Sunspot}                                                                         \\ \toprule
      Minsuforemr  & \textbf{\color{red}0.223} & \textbf{\color{red}0.287} & \textbf{\color{red}0.230} & \textbf{\color{red}1.587} & \textbf{\color{red}0.509} & \textbf{\color{red}0.276} & \textbf{\color{red}0.311} & \textbf{\color{red}0.364} & \textbf{\color{red}0.431} & \textbf{\color{red}0.248} & \textbf{\color{red}0.753} & \underline{\color{blue}0.445}          & \textbf{\color{red}0.363} & \textbf{\color{red}0.361} & 1.194          & 0.544          & \textbf{\color{red}0.808} & 2.776          & \textbf{\color{red}1.053} & 1.232          & 0.674          & 0.377          & \textbf{\color{red}0.436} & \textbf{\color{red}0.445} & \textbf{\color{red}0.756} & 1.364          & 0.434          \\
      iTransformer &  \underline{\color{blue}0.310}          & \underline{\color{blue}0.390}          & \underline{\color{blue}0.376}          & \underline{\color{blue}2.187}          & \underline{\color{blue}0.656}          & \underline{\color{blue}0.366}         & \underline{\color{blue}0.410}          & \underline{\color{blue}0.378}          & \underline{\color{blue}0.439}          & \underline{\color{blue}0.251}          & \underline{\color{blue}0.782}          & \textbf{\color{red}0.438} & \underline{\color{blue}0.365}          & \underline{\color{blue}0.388}          & 1.206          & 0.552          & \underline{\color{blue}0.832}          & 2.858          & \underline{\color{blue}1.066}          & 1.258          & 0.687          & 0.382          & \underline{\color{blue}0.440}          & \underline{\color{blue}0.450}          & \underline{\color{blue}0.761}          & 1.381          & 0.435          \\
      N-Beats      & 0.337          & 0.423          & 0.392          & 2.558          & 0.700          & 0.390          & 0.493          & 0.670          & 0.638          & 0.409          & 1.016          & 0.625          & 0.546          & 0.834          & \underline{\color{blue}1.028}          & \underline{\color{blue}0.543}          & 0.962          & \textbf{\color{red}2.017} & 1.387          & \underline{\color{blue}0.943}          & \underline{\color{blue}0.572}          & \textbf{\color{red}0.363} & 0.445          & 0.474          & 0.804          & \underline{\color{blue}1.275}          & \textbf{\color{red}0.425} \\
      N-Hits       & 0.346          & 0.418          & 0.402          & 2.492          & 0.687          & 0.396          & 0.442        & 0.538          & 0.578          & 0.364          & 0.989          & 0.540          & 0.464          & 0.802          & \textbf{\color{red}0.982} & \textbf{\color{red}0.532} & 0.865          & \underline{\color{blue}2.597}          & 1.189          & \textbf{\color{red}0.858} & \textbf{\color{red}0.562} & \underline{\color{blue}0.365}          & 0.442          & 0.463          & 0.791          & 1.338          & \underline{\color{blue}0.430}          \\
      Autoformer   & 0.635          & 0.611          & 0.614          & 3.980          & 0.881          & 0.518          & 0.741          & 0.901          & 0.757          & 0.447          & 1.124          & 0.789          & 0.737          & 0.788          & 1.252          & 0.640          & 1.058          & 2.806          & 1.370          & 1.057          & 0.674          & 0.400          & 0.458          & 0.474          & 0.786          & \textbf{\color{red}1.255} & 0.435         \\ \bottomrule
  \end{tabular}
    \begin{tablenotes}
      \item[*] All results are averaged across all prediction lengths. The results for all prediction lengths and the experimental settings are provided in Appendix \ref{subsec_mts_monash}. The definitions of all metrics are provided in Appendix \ref{secsub_metrics}.
    \end{tablenotes}
    \end{threeparttable}
    }
    \end{table*}

\textbf{Univariate results}: 
The average results for univariate TS forecasting are shown in Table \ref{tb3}. 
It is evident that the proposed Minusformer continues to maintain a SOTA performance across various prediction length settings compared to the benchmarks.
In summary, compared with the hyperparameter-searched Periodformer, 
Minusformer yields an average {\bf 4.8\%} reduction across five datasets, and it achieves an average of {\bf 26} best terms.
For example, under the input-96-predict-96 setting, Minusformer yields a reduction of {\bf 11.2\%} (0.143$\rightarrow$0.127) in MSE for Traffic. 
Obviously, the experimental results again verify the superiority of Minusformer on univariate TS forecasting tasks.

\subsection{Evaluation on Monash TS datasets}

Further, we evaluate the proposed method on 7 Monash TS datasets \citep{godahewa2021monash} (e.g., NN5, M4 and Sunspot, etc.) and 7 diverse metrics (e.g., MAPE, sMAPE, MASE and Quantile, etc.) to systematically evaluate our model. All experiments are compared under the same input length (e.g., $I$=96) and output lengths (e.g., $O$=\{96, 192, 336 and 720\}). As shown in Table \ref{tb_metric_1}, the proposed Minusformer emerged as the frontrunner, achieving a score of \textbf{41 out of 54}. Please refer to Appendices \ref{secsub_metrics} and \ref{subsec_mts_monash} for details about the definition and experimental settings, respectively.

\subsection{Comparative analysis}
Notably, several pioneering models have also achieved competitive performance on certain datasets under particular settings.
For instance, Informer, considered a groundbreaking model in long-term TS forecasting, demonstrates advanced performance on the Solar-Energy dataset with input-96-predict-192 and -720 settings.
This is due to the substantial presence of zero values on each column attribute of the Solor-Energy dataset.
This renders the KL-divergence based ProbSparse Attention, as adopted in Informer, highly effective on this sparse dataset.
Additionally, linear-based methods (e.g., DLinear) have demonstrated promising results on the Weather dataset with input-336-predict-720 setting, while convolution-based methods (e.g., SCINet) have yielded favorable results on the PEMS dataset with input-168-predict-192 setting. 
This phenomenon can be ascribed to a twofold interplay of factors. Previously, the diversity of input settings exerts a direct influence on model generalization. Secondarily, other models exhibit a propensity to overfit non-stationary TS characterized by aperiodic fluctuations.
Remarkably, Minusformer adeptly mitigates both overfitting and underfitting challenges in multivariate TS forecasting, thereby enhancing its overall performance.
Particularly on datasets with numerous attributes, e.g., Traffic and Solor-Energy, Minusformer achieves superior performance by feeding the learned meaningful patterns to the output layer at each block.

\subsection{Effectiveness}
\label{ssec_ablation}

\begin{figure*}[t]
    \centerline{\includegraphics[width=\textwidth]{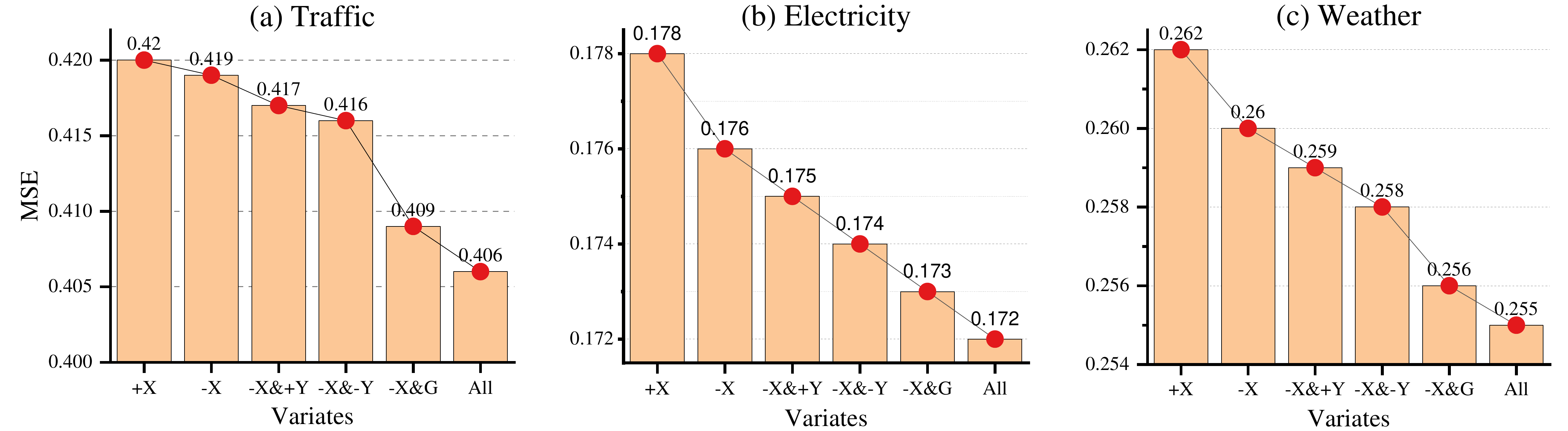}}
    \caption{Ablation studies on various components of Minusformer. All results are averaged across all prediction lengths. The variables $X$ and $Y$ represent the input and output streams, while the signs `+' and `-' denote the addition or subtraction operations used when the streams' aggregation. The letter `G' denotes  adding a gating mechanism to the output of each block.}
    \label{fig_variates}
  \end{figure*}
 
\begin{figure*}[t]
    \centerline{\includegraphics[width=\textwidth]{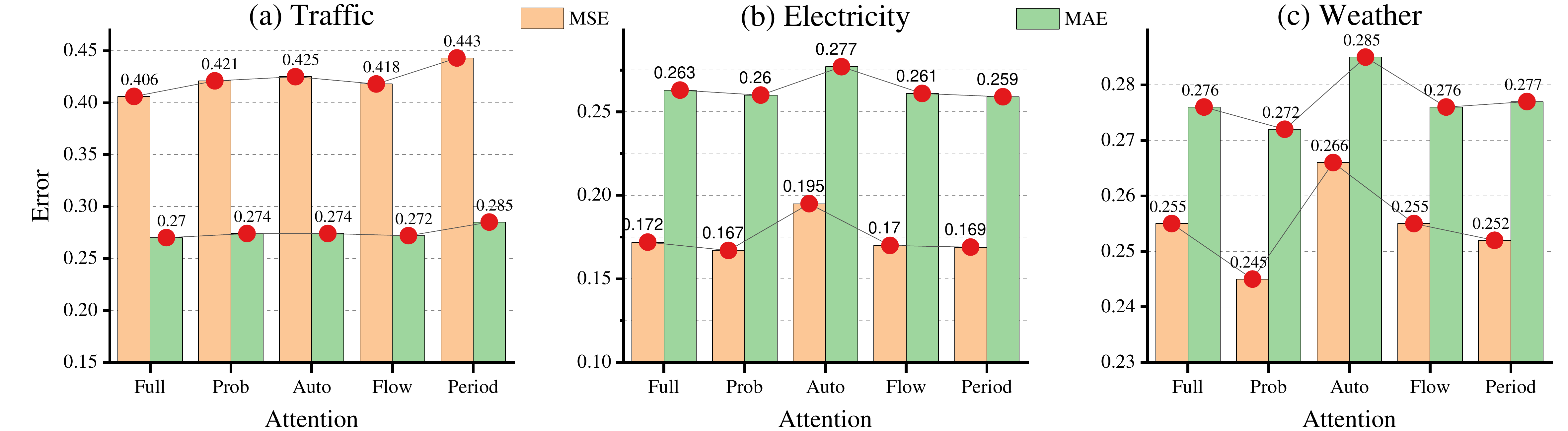}}
    \caption{Ablation studies of Minusformer using various Attention. All results are averaged across all prediction lengths. The tick labels of the X-axis are the abbreviation of Attention types. The detailed setup and results for all prediction lengths are provided in Appendix \ref{app_improve_attn}.
    }
    \label{fig_attns}
\end{figure*}

To validate the effectiveness of Minusformer components, we conduct comprehensive ablation studies encompassing both component replacement and component removal experiments, as shown in Fig. \ref{fig_variates}.
We utilize signs `+' and `-' to denote the utilization of addition or subtraction operations during the aggregation process of the input or output streams. 
In cases involving only input streams, it becomes evident that the model's average performance is superior when employing subtraction (-X) compared to when employing addition (+X).
E.g., on the Electricity dataset, forecast error is reduced by {\bf 1.1\%} (0.178$\rightarrow$0.176).
Moreover, with the introduction of a high-speed output stream to the model, shifting the aggregation method of the output stream from addition (+Y) to subtraction (-Y) is poised to further enhance the model's performance.
Afterward, incorporating gating mechanisms (G) into the model holds the potential to improve predictive performance again.
E.g., on the Traffic dataset, forecast error is reduced by {\bf 2.4\%} (0.419$\rightarrow$0.409).
In summary, integrating the advantages of the aforementioned components has the potential to significantly boost the model's performance across the board.

\begin{figure*}[t]
    \centerline{\includegraphics[width=0.99\textwidth]{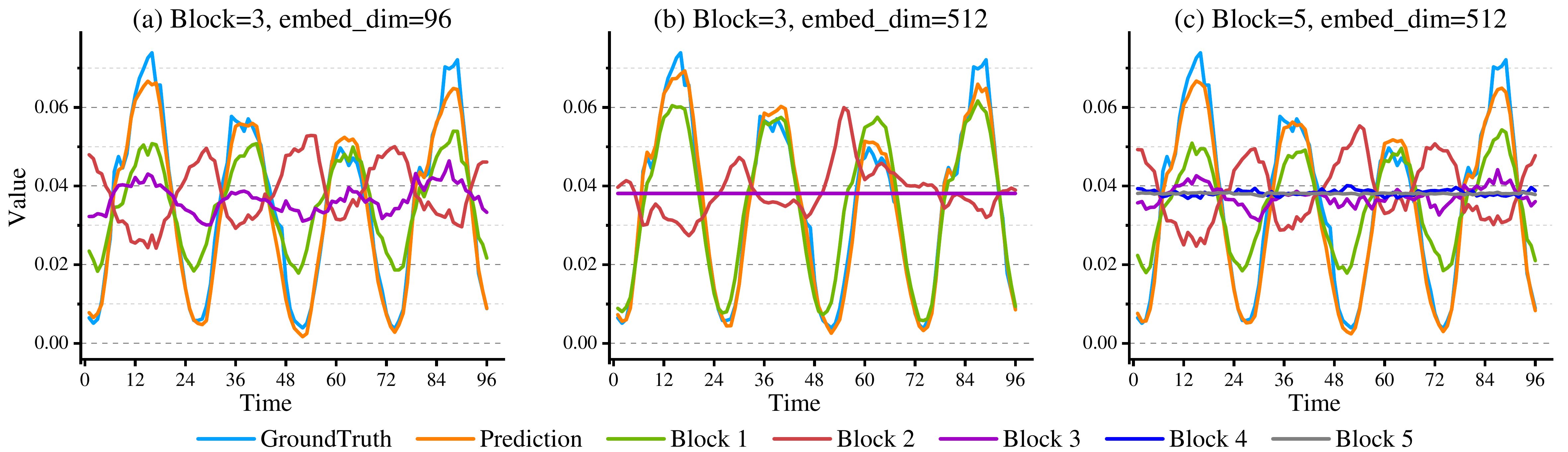}}
    \caption{Visualization depicting the output of each block in Minusformer. The experiment was implemented on the Traffic dataset using the setting of Input-96-Predict-96. The utilized models have the same hyperparameter settings and similar performance.} %
    \label{fig_visblock}
\end{figure*}

\subsection{Generality}

To investigate Minusformer's generality as a universal architecture, we substituted its original Attention with other novel Attention mechanisms to observe the resulting changes in model performance.
As shown in Fig. \ref{fig_attns}, after harnessing the newly invented Attention within Minusformer, its performance exhibited considerable variation. 
E.g., the average MSE of Prob-Attention \citep{Zhou2021Informer} on the Electricity and Weather datasets witnessed a reduction of {\bf 46\%} (0.311$\rightarrow$0.167) and {\bf 61\%} (0.634$\rightarrow$0.245), respectively, surpassing Full-Attention and achieving new SOTA performance.
Furthermore, both Period-Attention \citep{liang2023does} and Flow-Attention \citep{wu2022flowformer} demonstrate commendable performance on the aforementioned datasets.
The improvement in Auto-Correlation \citep{wu2021autoformer} falls short of expectations, primarily due to the incapacity of its autocorrelation mechanism to capture nuanced patterns of change within attributes characterized by substantial fluctuations.
The conducted experiments suggest that Minusformer can serve as a versatile architecture, amenable to the integration of novel modules, thereby facilitating the enhancement of performance in the domain of TS forecasting.

\subsection{Interpretability}
\label{ssec_intpbt}

The intrinsic characteristic of Minusformer lies in the alignment of the output from each block with the shape of the final output. This alignment, in turn, expedites the decomposition and visualization of the model's learning process.
As depicted in Fig. \ref{fig_visblock}, the output of each block in the Minusformer is visualized, with variations in both width and depth configurations.
Moreover, Attention within each block is visualized
in Fig. \ref{fig_vis_attn_all}, which corresponds to the model in Fig. \ref{fig_visblock}(a).
It becomes evident that each block discerns and assimilates meaningful patterns within the series.
Specifically, comparing Fig. \ref{fig_visblock}(a) and \ref{fig_visblock}(b), when the embedding dimension is low, each block must learn salient patterns. 
However, as the model capacity increases, the efficacy of the deep blocks diminishes. E.g., block 3 in Fig. \ref{fig_visblock}(b) solely acquires knowledge of a single constant.
Further, increasing the depth of the model, as depicted in Fig. \ref{fig_visblock}(c), it becomes evident that the amplitude of the shallow block decreases, and numerous components are transferred to the deep block. E.g., block 3 in Fig. \ref{fig_visblock}(c) exhibits the same behavior as in Fig. \ref{fig_visblock}(a).
This suggests that increasing the depth of Minusformer enhances its learning capacity without heightening the risk of overfitting.

\subsection{Go deeper}

\begin{wrapfigure}{r}{0.4\columnwidth}
      \begin{center}
      \vspace{-55pt}
        \includegraphics[width=0.4\columnwidth]{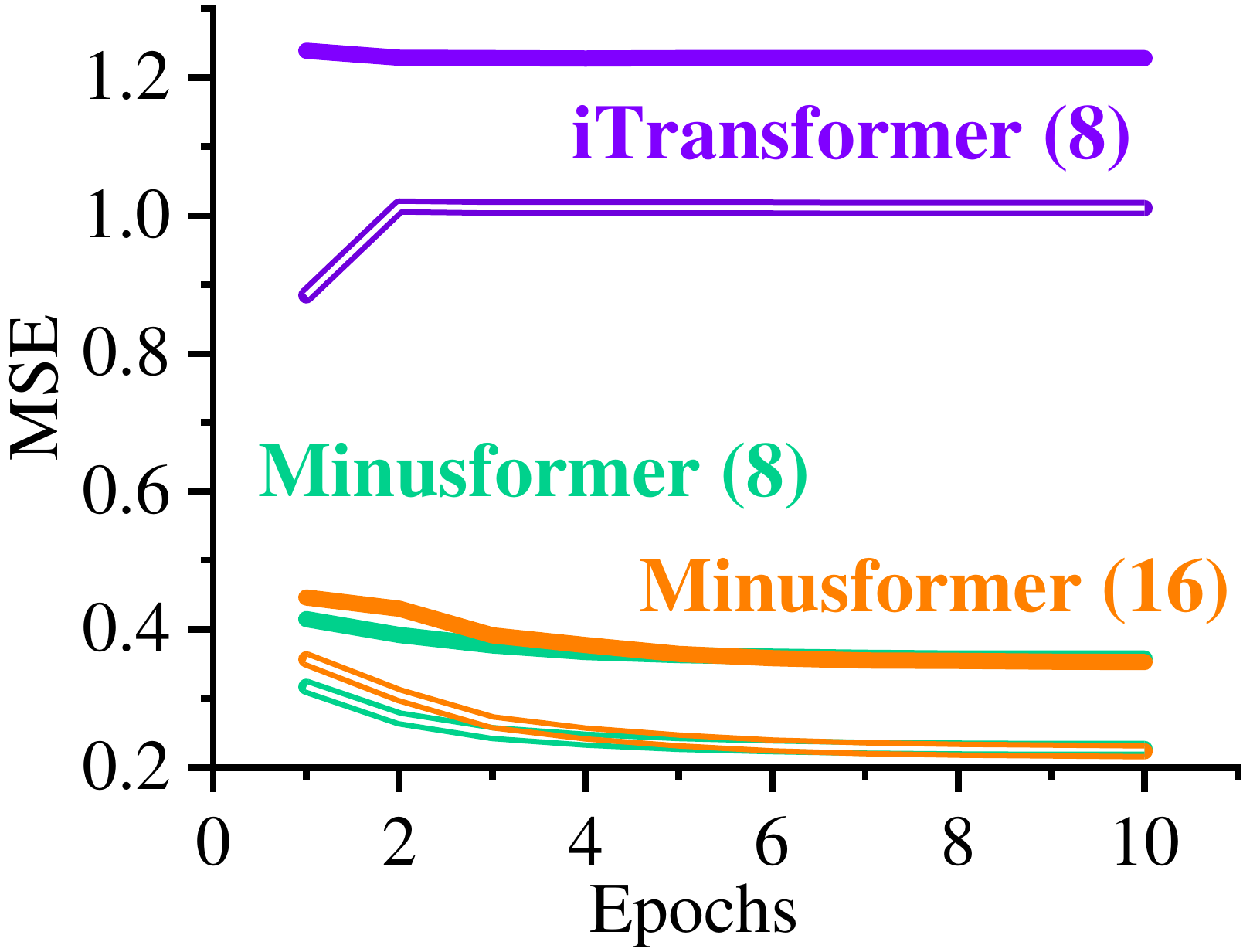}
      \end{center}
      \vspace{-10pt}
      \caption{\small{Process of training (Dash) and validation (Solid) when models go deeper.}}
      \vspace{-25pt}
      \label{fig_godeeper}
\end{wrapfigure}

Given the Minusformer's robustness against overfitting, it can be designed with considerable depth.
Fig. \ref{fig_godeeper} illustrates the scenarios when models go deeper.
Serious overfitting happens when the number of iTransformer blocks is increased from 4 to 8.
However, even with the Minusformer blocks deepened to 16, it continues to exhibit excellent performance.
In addition, Minusformer is less sensitive to hyperparameters, with details presented in Appendix \ref{app_hpyer}.

\section{Conclusion}
\label{sec_con}

In this paper, we designed a architecture utilizing exclusively minus signs for information aggregation, specifically crafted to address the ubiquitous overfitting problem prevalent in TS forecasting models, which is called Minusformer.
The designed architecture facilitates the learning-driven, implicit, progressive decomposition of both the input and output streams, and is accompanied by a theoretical rationale behind the effectiveness of the subtraction, thereby empowering the model with enhanced generality, interpretability, and resilience against overfitting.
Minusformer is less insensitive to hyperparameters and can be designed very deep without overfitting.
Extensive experiments demonstrate that Minusformer achieves SOTA performance.

\bibliographystyle{plainnat}
\bibliography{citations}


\clearpage
\appendix


\section{Related Work}
\label{sec_rework}

\subsection{Classical models for TS forecasting}

TS forecasting is a classic research field where numerous methods have been invented to utilize historical series to predict future missing values. 
Early classical methods \citep{piccolo1990distance,gardner1985exponential, li2010parsimonious} are widely applied because of their well-defined theoretical guarantee and interpretability. For example, ARIMA \citep{piccolo1990distance} initially transforms a non-stationary TS into a stationary one via difference, and subsequently approximates it using a linear model with several parameters. 
Exponential smoothing \citep{gardner1985exponential} predicts outcomes at future horizons by computing a weighted average across historical data. In addition, some regression-based methods, e.g., random forest regression (RFR) \citep{liaw2002classification} and support vector regression (SVR) \citep{castro2009online}, etc., are also applied to TS  forecasting. 
These methods are straightforward and have fewer parameters to tune, making them a reliable workhorse for TS forecasting. 
However, their shortcoming is insufficient data fitting ability, especially for high-dimensional series, resulting in limited performance.

\subsection{Deep models for TS forecasting}

The advancement of deep learning has greatly boosted the progress of TS forecasting. 
Specifically, convolutional neural networks (CNNs) \citep{lecun1998gradient} and recurrent neural networks (RNNs) \citep{connor1994recurrent} have been adopted by many works to model nonlinear dependencies of TS, 
e.g., LSTNet \citep{lai2018modeling} improve CNNs by adding recursive skip connections to capture long- and short-term temporal patterns;
DeepAR \citep{salinas2020deepar} predicts the probability distribution by combining autoregressive methods and RNNs.
Several works have improved the series aggregation forms of Attention mechanism, such as operations of exponential intervals adopted in LogTrans \citep{li2019enhancing}, ProbSparse activations in Informer \citep{Zhou2021Informer}, frequency sampling in FEDformer \citep{zhou2022fedformer} and iterative refinement in Scaleformer \citep{shabani2022scaleformer}. 
Besides, GNNs and TCNs have been utilized in some methods \citep{wu2019graph, li2023dynamic, yao2021mvstgn,liu2022scinet,wu2022timesnet} for TS forecasting on graph data.
The aforementioned methods solely concentrate on the forms of aggregating input series, overlooking the challenges posed by the overfitting problem.




\subsection{Decomposition for TS forecasting}

Time series exhibit a variety of patterns, and it is meaningful and beneficial to decompose them into several components, each representing an underlying category of patterns that evolving over time \citep{1976_timeseries}.
Several methods, e.g., STL \citep{cleveland1990stl}, Prophet \citep{taylor2018forecasting} and N-BEATS \citep{oreshkin2019n}, commonly utilize decomposition as a preprocessing phase on historical series.
There are also some methods, e.g., Autoformer \citep{wu2021autoformer}, FEDformer \citep{zhou2022fedformer} and Non-stationary Transformers \citep{liu2022non}, that harness decomposition into the Attention module.
The aforementioned methods attempt to apply decomposition to input series to enhance predictability, reduce computational complexity, or ameliorate the adverse effects of non-stationarity.
Nevertheless, these prevalent methods are susceptible to significant overfitting when applied to non-stationary TS.
In this paper, the proposed method utilizes a progressive approach to guide the learning of each time-varying pattern. This is achieved by implicitly decomposing the supervision signals, which helps address the issue of overfitting.

\section{Theoretical rationale of minusformer}
\label{app_overfit_theory}

In this subsection, we provide a theoretical analysis of how Minusformer alleviates overfitting.
In practice, observations $Y$ often contain additive noise $\varepsilon$: 
$Y  = \mathcal{Y} + \varepsilon $, 
where $\varepsilon \sim \mathcal{N}(0, \xi )$. 
Then, the estimation error (MSE) of the final model is:
\begin{align}
  \mathbb{E}[(\hat{Y} - Y )^2] & = \mathbb{E}[( \hat{Y} - \mathcal{Y} - \varepsilon )^2] \nonumber \\ 
   & = \mathbb{E}[(\hat{Y} - \mathcal{Y} )^2] + \mathbb{E}(\varepsilon^2) 
   - 2\mathbb{E}(\varepsilon(\hat{Y} - \mathcal{Y} )) \nonumber \\
   & = \mathbb{E}\left( (\hat{Y} - \mathbb{E}(\hat{Y}) + \mathbb{E}(\hat{Y}) - \mathcal{Y} )^2 \right) + \xi^2   - 2\mathbb{E}(\varepsilon(\hat{Y} - \mathcal{Y} )) \nonumber \\
   & = \mathbb{E}\left( (\hat{Y} - \mathbb{E}(\hat{Y}))^2  +(\mathbb{E}(\hat{Y}) - \mathcal{Y}  )^2 + 2(\hat{Y} - \mathbb{E}(\hat{Y}) ) (\mathbb{E}(\hat{Y}) - \mathcal{Y} )  \right)  + \xi^2   - 2\mathbb{E}(\varepsilon(\hat{Y} - \mathcal{Y} )) \nonumber \\
   & = \mathbb{E}\left( (\hat{Y} - \mathbb{E}(\hat{Y}))^2 \right)  +(\mathbb{E}(\hat{Y}) - \mathcal{Y}  )^2 
   + \xi^2   - 2\mathbb{E}(\varepsilon(\hat{Y} - \mathcal{Y} )) \nonumber \\
   & = \text{Var}(\hat{Y}) + (\text{Bias}(\hat{Y}))^2 + \xi^2   - 2\mathbb{E}(\varepsilon(\hat{Y} - \mathcal{Y} )). \label{eq_a3_0}
\end{align}
Based on this, Equation \ref{eq_a3_0} can be reformulated as
\begin{align}
  \underbrace{\text{Var}(\hat{Y}) + (\text{Bias}(\hat{Y}))^2 + \xi^2}_{\text{Test Error}}  =  \underbrace{\mathbb{E}[(\hat{Y} - Y )^2] + 2\mathbb{E}(\varepsilon(\hat{Y} - \mathcal{Y} ))}_{\text{Training Error}} .
  \label{eq_a3}
\end{align}


\subsection{Proof of Theorem \ref{th1}}
\label{seca_proof_th1}

Now, we proof how subtraction in Minusformer mitigates overfitting, i.e., how Equation \ref{eq_a2} diminishes the variance term in Equation \ref{eq_a3}.
According to the estimation error $e_l \overset{i.i.d}{\sim} \mathcal{N}(0, \nu)$, we have $\text{Var}(f_l(X))=\nu$. Then, utilizing $\text{Cov}(f_l, f_{k\neq l})=\mu$ in Theorem \ref{th1}, we have the proof as follows:
\begin{proof}
\begin{align}
  \text{Var}(\hat{Y}) & = \text{Var}(\mathcal{F}(X)) = \frac{1}{\hbar^2} \text{Var}\left( i \sum_{l=0}^{\hbar} \alpha f_{2l+1}(X) - i \sum_{l=0}^{\hbar} \alpha f_{2l}(X) \right) \nonumber \\
  & = \frac{1}{\hbar^2} \text{Var}\left( \sum_{l=0}^{\hbar} \alpha f_{2l+1}(X)\right) +   \frac{1}{\hbar^2} \text{Var}\left( \sum_{l=0}^{\hbar} \alpha f_{2l}(X) \right) 
  -  \frac{1}{\hbar^2} \text{Cov}\left( \sum_{l=0}^{\hbar} \alpha f_{2l+1}(X) \sum_{l=0}^{\hbar} \alpha f_{2l}(X)  \right) \nonumber \\
  & = \frac{1}{\hbar^2} \sum_{l=0}^{\hbar} \alpha^2 \text{Var}\left( f_{2l+1}(X)\right) 
  +  \frac{1}{\hbar^2} \sum_{l=0}^{\hbar} \sum_{k=1,k\neq j}^{\hbar} \alpha^2 \text{Cov}\left( f_{2l+1}(X)  f_{2k+1}(X)  \right) \nonumber \\ 
  & \ \ \ \ \ \ + \frac{1}{\hbar^2} \sum_{l=0}^{\hbar} \alpha^2 \text{Var}\left(  f_{2l}(X) \right)
  +  \frac{1}{\hbar^2} \sum_{l=0}^{\hbar} \sum_{k=1,k\neq j}^{\hbar} \alpha^2 \text{Cov}\left( f_{2l}(X)  f_{2k}(X)  \right)  \nonumber \\
  & \ \ \ \ \ \ -  \frac{1}{\hbar^2} \sum_{l=0}^{\hbar} \sum_{k=0}^{\hbar} \alpha^2 \text{Cov}\left(  f_{2l+1}(X)  f_{2k}(X)  \right) \nonumber \\
  & = \frac{1}{\hbar^2} \sum_{l=0}^{\hbar} \alpha^2 \nu 
  +  \frac{1}{\hbar^2} \sum_{l=0}^{\hbar} \sum_{k=1,k\neq j}^{\hbar} \alpha^2 
  + \frac{1}{\hbar^2} \sum_{l=0}^{\hbar} \alpha^2 \nu  
    +  \frac{1}{\hbar^2} \sum_{l=0}^{\hbar} \sum_{k=1,k\neq j}^{\hbar} \alpha^2 \mu 
  -  \frac{1}{\hbar^2} \sum_{l=0}^{\hbar} \sum_{k=0}^{\hbar} \alpha^2 \mu \nonumber \\
  & = \frac{1}{\hbar} \alpha^2 \nu +  \frac{\hbar-1}{\hbar}  \alpha^2 \mu 
  + \frac{1}{\hbar} \alpha^2 \nu +  \frac{\hbar-1}{\hbar} \alpha^2 \mu -  \alpha^2 \mu \nonumber \\
  & = \frac{2}{\hbar} \alpha^2 \nu +  2\frac{\hbar-1}{\hbar}  \alpha^2 \mu -  \alpha^2 \mu \nonumber \\
  & < \frac{2}{\hbar} \alpha^2 (\nu + \mu) \nonumber \\
  & \le \frac{4}{L} \alpha^2 (\nu + \mu). \label{eq_a4}
\end{align}
\end{proof}

\section{Dataset}
\label{app_dataset}

\subsection{Commonly used TS datasets}

The information of the experiment datasets used in this paper are summarized as follows: (1) Electricity Transformer Temperature (ETT) dataset \cite{Zhou2021Informer}, which contains the data collected from two electricity transformers in two separated counties in China, including the load and the oil temperature recorded every 15 minutes (ETTm) or 1 hour (ETTh) between July 2016 and July 2018. (2) Electricity (ECL) dataset \footnote[1]{https://archive.ics.uci.edu/ml/datasets/ElectricityLoadDiagrams20112014} collects the hourly electricity consumption of 321 clients (each column) from 2012 to 2014. (3) Exchange \cite{lai2018modeling} records the current exchange of 8 different countries from 1990 to 2016. (4) Traffic dataset \footnote[2]{http://pems.dot.ca.gov} records the occupation rate of freeway system across State of California measured by 861 sensors. (5) Weather dataset \footnote[3]{https://www.bgc-jena.mpg.de/wetter} records every 10 minutes for 21 meteorological indicators in Germany throughout 2020. (6) Solar-Energy \cite{lai2018modeling} documents the solar power generation of 137 photovoltaic (PV) facilities in the year 2006, with data collected at 10-minute intervals. (7) The PEMS dataset \cite{liu2022scinet} comprises publicly available traffic network data from California, collected within 5-minute intervals and encompassing 358 attributes.
(8) Illness (ILI) dataset \footnote[4]{https://gis.cdc.gov/grasp/fluview/fluportaldashboard.html} describes the influenza-like illness patients in the United States between 2002 and 2021, which records the ratio of patients seen with illness and the total number of the patients. 
The detailed statistics information of the datasets is shown in Table \ref{tb1}.

\begin{table}[h]
  \vspace{-1.5em}
  \centering
  \caption{Details of the seven TS datasets. }
  \label{tb1}
  \resizebox{0.4\textwidth}{!}
  { \tiny
    \begin{tabular}{cccc}
      \toprule
       Dataset    & length  & features & frequency \\
      \midrule
      ETTh1       & 17,420  & 7       & 1h \\
      ETTh2       & 17,420  & 7       & 1h \\
      ETTm1       & 69,680  & 7       & 15m \\
      ETTm2       & 69,680  & 7       & 15m \\
      Electricity & 26,304  & 321     & 1h  \\
      Exchange    & 7,588   & 8       & 1d  \\
      Traffic     & 17,544  & 862     & 1h  \\
      Weather     & 52,696  & 21      & 10m \\
      Solar       & 52,560  & 137     & 10m  \\
      PEMS        & 26,208  & 358     & 5m  \\
      Illness     & 966     & 7       & 7d  \\
      \bottomrule
    \end{tabular}
  }
  \vspace{-1em}
\end{table}

\subsection{Monash TS Forecasting Datasets}
(1) Saugeenday dataset \footnote[5]{https://zenodo.org/records/4656058} contains a single very long time series representing the daily mean flow of the Saugeen River at Walkerton in cubic meters per second and the length of this time series is 23741.
(2) Sunspot dataset \footnote[6]{https://www.kaggle.com/datasets/robervalt/sunspots} contains monthly numbers of sunspots, as from the World Data Center, aka SIDC, between 1749 and 2018 with a total observation of 3240 months.
(3) M4 dataset \cite{MAKRIDAKIS202054} is a collection of 100,000 time series used in the fourth Makridakis Prediction Contest. The dataset consists of a time series of annual, quarterly, monthly, and other weekly, daily, and hourly data. 
In this paper, we utilize the hourly version of the M4 dataset and standardize its length to 768.
(4) NN5 dataset \footnote[7]{http://www.neural-forecasting-competition.com/downloads/NN5/datasets/download.htm} contains daily time series originated from the observation of daily withdrawals at 111 randomly selected different cash machines at different locations within England.
(5) Oikolab Weather dataset \footnote[8]{https://docs.oikolab.com} contains hundreds of terabytes of weather data, all of them are post-processed data from national weather agencies.
(6) US Births dataset \footnote[9]{https://cran.r-project.org/web/packages/mosaicData} provides birth rates and related data across the 50 states and DC from 2016 to 2021, with a total observation of 7305.
(7) Rideshare dataset \cite{godahewa2021monash} comprises diverse hourly time series representations of attributes pertinent to Uber and Lyft rideshare services across multiple locations in New York. The data spans from November 26th, 2018, to December 18th, 2018.
All the above data sets are collected in the Monash library\footnote[10]{https://forecastingdata.org}.

\section{Model complexity and computation cost}
\label{seca_cost}

Minusformer contains gate mechanism and attention module. Among them, the gate mechanism is composed of linear layer and sigmoid activation function, i.e., $O_{l+1}=$Sigmoid(Linear($X_l$))$\cdot$Linear($X_l)$, which still maintains linear complexity. Therefore, like other Transformer models, the main complexity of our model is mainly in the attention module ($O(N^2)$).

Further, we evaluate the running time, memory usage, parameters, and FLOPs of our model in comparison to other Transformer-based models using identical settings. The computation costs are shown in the Table \ref{tb_cost}.
It reveals that the proposed model has a smaller computational cost and faster running speed compared to others. It is noteworthy that the introduction of the auxiliary output branch only marginally increases the computational cost, i.e., the differences between the indicators of our model without gate and those of the original model are negligible.

\begin{table}[h!]
  \centering
  \caption{Model complexity, training efficiency and computation cost. }
  \label{tb_cost}
  \resizebox{0.9\textwidth}{!}
  { 
    \begin{tabular}{ccccc}
      \toprule
    Models                 & Seconds/Epoch & GPU Memory Usage (GB) & Parameters (MB) & FLOPs (GB) \\ \toprule
    Minusformer            & 28.7          & 1.2              & 8.8       & 3.1   \\
    Minusformer w/o Gate & 28.0          & 1.1              & 6.5       & 2.5   \\
    PatchTST               & 30.1          & 2.3              & 8.4       & 17.4  \\
    FEDformer              & 396           & 7.1              & 14.7      & 558.1 \\
    Autoformer             & 84.5          & 4.8              & 10.5      & 66.2  \\
    Informer               & 81.8          & 2.3              & 11.3      & 63.3   \\ \bottomrule
    \end{tabular}
    }
\end{table}

\section{Hyperparameter sensitivity}
\label{app_hpyer}

\begin{figure*}[h!]
  \centering
  \centerline{\includegraphics[width=\textwidth]{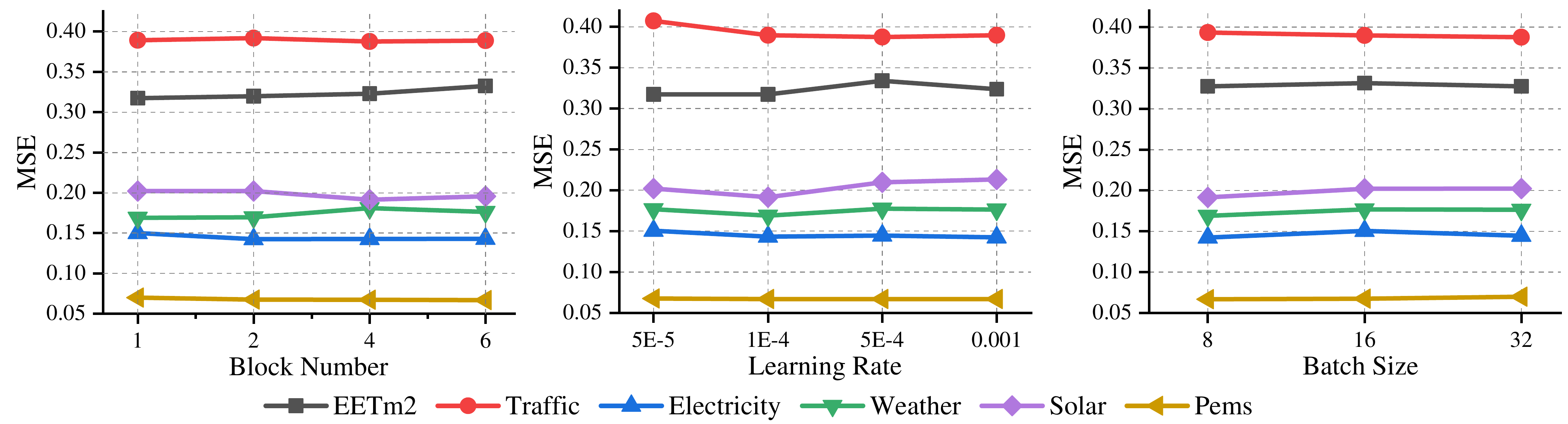}}
  \caption{Hyperparameter sensitivity with respect to the number of block, the learning rate and the number of batch size. The results are recorded with the input length $I = 96$ and the prediction length $O = 12$ for PEMS and $O = 96$ for others.}
  \label{fig_hyper} 
\end{figure*}

We evaluate the hyperparameter sensitivity of Minusformer with respect to the learning rate, the number of the block, the batch size and the embedding dimension.
As shown in Fig. \ref{fig_hyper}, the performance of Minusformer fluctuates under different hyperparameter settings.
In most cases, increasing the number of blocks tends to enhance model performance.
Once again, this confirms that Minusformer exhibits resilience against overfitting across diverse datasets.
Notably, we observe that the learning rate, being the most prevalent influencing factor, especially in scenarios involving numerous attributes.
Meanwhile, modifying the batch size induces minor fluctuations in the model's performance, albeit with a limited impact.
Furthermore, we conducted extensive analysis on hyperparameters, as shown in Tables \ref{tb_hpyer1}, \ref{tb_hpyer2}, and \ref{tb_hpyer3}. It is evident that the proposed model is less insensitive to numerous hyperparameters.
In summary, Minusformer exhibits low sensitivity to these hyperparameters, thereby enhancing its resilience against overfitting.

\begin{table}[!t]
    \centering
    \caption{Ablation studies of Minusformer's hyperparameters on the Electricity dataset.}
    \label{tb_hpyer1}
    \resizebox{0.9\textwidth}{!}
    {
    \begin{tabular}{c|ccccc| cccc|cccc}
    \toprule
      & \multicolumn{5}{c}{Block Number}            & \multicolumn{4}{c}{Embedding Dimension} & \multicolumn{4}{c}{Learning Rate} \\ \toprule
    Metrics     & 2     & 3     & 4     & 5     & 6     & 92     & 384   & 512   & 768   & 1E-3 & 5E-4 & 1E-4 & 5E-5 \\  \toprule
    MSE         & 0.137 & 0.137 & 0.137 & 0.138 & 0.137 & 0.140   & 0.137 & 0.143 & 0.136 & 0.143 & 0.143  & 0.144  & 0.145   \\
    MAE         & 0.231 & 0.231 & 0.231 & 0.230  & 0.231 & 0.233  & 0.231 & 0.235 & 0.231 & 0.231 & 0.231  & 0.230   & 0.232   \\
    RMSP        & 0.214 & 0.214 & 0.256 & 0.215 & 0.213 & 0.219  & 0.215 & 0.214 & 0.212 & 0.214 & 0.214  & 0.212  & 0.226   \\
    MAPE        & 2.295 & 2.294 & 2.34  & 2.32  & 2.32  & 2.397  & 2.318 & 2.295 & 2.305 & 2.29  & 2.303  & 2.302  & 2.35    \\
    sMAPE       & 0.466 & 0.464 & 0.467 & 0.466 & 0.464 & 0.475  & 0.467 & 0.466 & 0.468 & 0.466 & 0.466  & 0.465  & 0.469   \\
    MASE        & 0.257 & 0.256 & 0.259 & 0.257 & 0.257 & 0.266  & 0.258 & 0.257 & 0.261 & 0.257 & 0.257  & 0.258  & 0.275   \\
    Q25         & 0.232 & 0.229 & 0.232 & 0.232 & 0.231 & 0.234  & 0.232 & 0.232 & 0.233 & 0.232 & 0.229  & 0.229  & 0.238   \\
    Q75         & 0.23  & 0.232 & 0.23  & 0.231 & 0.23  & 0.24   & 0.231 & 0.23  & 0.231 & 0.23  & 0.232  & 0.233  & 0.243  \\ \bottomrule
    \end{tabular}
    }
\end{table}

\begin{table}[!t]
    \centering
    \caption{Ablation studies of Minusformer's hyperparameters on the Traffic dataset.}
    \label{tb_hpyer2}
    \resizebox{0.9\textwidth}{!}
    { 
    \begin{tabular}{c|ccccc| cccc|cccc}
        \toprule
         & \multicolumn{5}{c|}{Block Number}            & \multicolumn{4}{c|}{Embedding Dimension}       & \multicolumn{4}{c}{Learning Rate} \\ \toprule
        Metrics & 2     & 3     & 4     & 5     & 6     & 92     & 384     &  512    & 768     & 1E-3   & 5E-4   & 1E-4   & 5E-5 \\ \toprule
        MSE     & 0.400 & 0.401 & 0.400 & 0.400 & 0.401 & 0.393  & 0.387   & 0.387   & 0.389   & 0.389  & 0.391  & 0.399  & 0.404  \\
        MAE     & 0.266 & 0.267 & 0.267 & 0.267 & 0.267 & 0.265  & 0.258   & 0.259   & 0.257   & 0.257  & 0.256  & 0.261  & 0.271  \\
        RMSP    & 0.208 & 0.209 & 0.208 & 0.209 & 0.208 & 0.207  & 0.198   & 0.199   & 0.200   & 0.207  & 0.203  & 0.208  & 0.214  \\
        MAPE    & 2.730 & 2.770 & 2.765 & 2.750 & 2.750 & 2.760  & 2.670   & 2.690   & 2.700   & 2.720  & 2.660  & 2.720  & 2.820  \\
        sMAPE   & 0.485 & 0.486 & 0.486 & 0.486 & 0.486 & 0.481  & 0.470   & 0.480   & 0.480   & 0.484  & 0.482  & 0.491  & 0.492  \\
        MASE    & 0.257 & 0.258 & 0.258 & 0.258 & 0.258 & 0.255  & 0.248   & 0.250   & 0.252   & 0.249  & 0.250  & 0.274  & 0.262  \\
        Q25     & 0.262 & 0.265 & 0.264 & 0.264 & 0.264 & 0.263  & 0.256   & 0.256   & 0.254   & 0.253  & 0.255  & 0.259  & 0.268  \\
        Q75     & 0.271 & 0.270 & 0.270 & 0.270 & 0.271 & 0.267  & 0.260   & 0.262   & 0.261   & 0.261  & 0.258  & 0.264  & 0.275  \\ \bottomrule
    \end{tabular}
    }
\end{table}

\begin{table}[!t]
    \centering
    \caption{Ablation studies of Minusformer's hyperparameters on the Weather dataset.}
    \label{tb_hpyer3}
    \resizebox{0.9\textwidth}{!}
    { 
        \begin{tabular}{c|ccccc| cccc|cccc}
            \toprule
             & \multicolumn{5}{c|}{Block Number}            & \multicolumn{4}{c|}{Embedding Dimension}       & \multicolumn{4}{c}{Learning Rate} \\ \toprule
            Metrics & 2     & 3     & 4     & 5     & 6     & 92     & 384     &  512     & 768     & 1E-3   & 5E-4   & 1E-4   & 5E-5 \\ \toprule
            MSE     & 0.160 & 0.156 & 0.160 & 0.159 & 0.157 & 0.161  & 0.156   & 0.158   & 0.159   & 0.215   & 0.168  & 0.157  & 0.159 \\
            MAE     & 0.203 & 0.201 & 0.206 & 0.203 & 0.199 & 0.205  & 0.201   & 0.202   & 0.202   & 0.271   & 0.211  & 0.202  & 0.202 \\
            RMSP    & 0.213 & 0.213 & 0.220 & 0.215 & 0.206 & 0.213  & 0.211   & 0.210   & 0.212   & 0.324   & 0.218  & 0.212  & 0.210 \\
            MAPE    & 8.990 & 8.860 & 9.220 & 9.430 & 9.130 & 10.600 & 9.151   & 8.387   & 8.450   & 17.680  & 8.530  & 7.900  & 7.860 \\
            sMAPE   & 0.529 & 0.524 & 0.534 & 0.528 & 0.519 & 0.531  & 0.526   & 0.525   & 0.524   & 0.637   & 0.542  & 0.529  & 0.523 \\
            MASE    & 1.162 & 1.110 & 1.172 & 1.161 & 1.090 & 1.263  & 1.171   & 1.151   & 1.094   & 1.230   & 1.202  & 1.170  & 1.200 \\
            Q25     & 0.198 & 0.196 & 0.203 & 0.197 & 0.188 & 0.197  & 0.195   & 0.197   & 0.189   & 0.272   & 0.206  & 0.200  & 0.195 \\
            Q75     & 0.209 & 0.207 & 0.209 & 0.209 & 0.210 & 0.212  & 0.208   & 0.206   & 0.215   & 0.269   & 0.215  & 0.205  & 0.209 \\ \bottomrule
        \end{tabular}
      }
  \end{table}

\section{Different attention promotion}
\label{app_improve_attn}

\begin{figure*}[h!]
  \centering
  \centerline{\includegraphics[width=\textwidth]{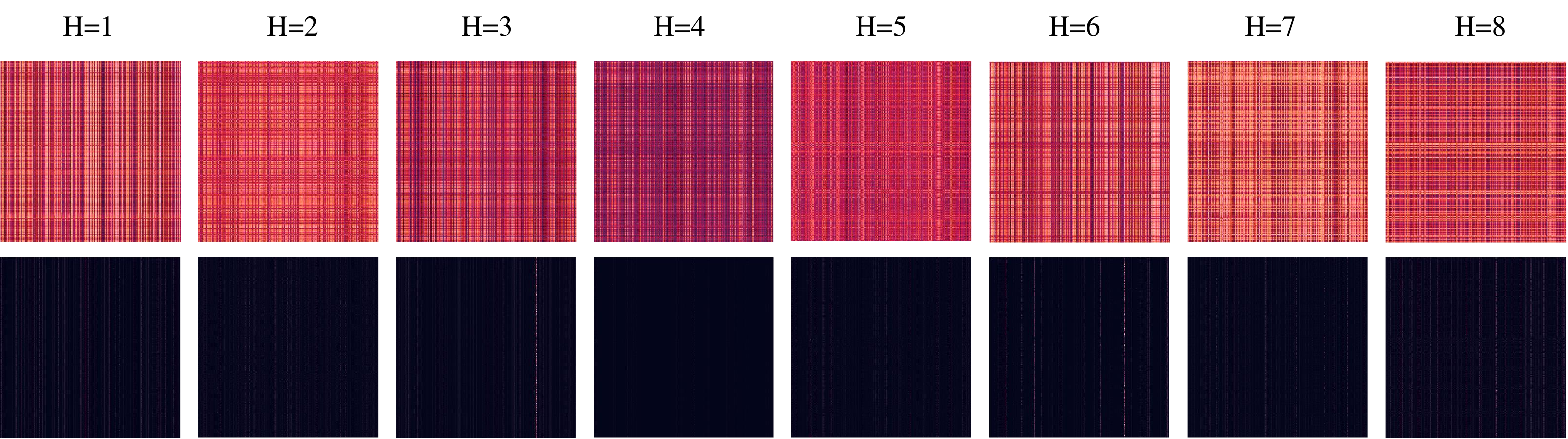}}
  \caption{Visualization of Attention utilized in Minusformer. The model is trained on Traffic dataset with 862 attributes under the setting of Input-96-Predict-96. Both the Attention score (top) and the post-softmax score (below) of the 8 heads (H) are from the first block.}
  \label{fig_vis_attn_all} 
\end{figure*}


We visualized the Attention maps for all Attention heads in the initial layer of Minusformer, and the results are presented in Fig. \ref{fig_vis_attn_all}.
It is evident that the Attention score graph exhibits numerous bar structures, which is especially prominent on the post-softmax Attention map.
This implies that there exists a row in Query that bears a striking resemblance to all the columns in Key. 
This scenario arises when there are numerous attributes, many of which are homogeneously represented.
Our speculation suggests that this capability of the Attention may explain its ability to capture subtle patterns in TS without succumbing to overfitting.
Furthermore, we substituted its original Attention with other novel Attention mechanisms to compare the resulting changes in model performance. 
The full results for all prediction lengths are provided in Table \ref{tb_attns}.

\begin{table*}[h!]
  \centering
  \caption{Ablation of different Attention layers.}
  \label{tb_attns}
  \resizebox{0.93\textwidth}{!} 
  {
    \tiny
    \begin{threeparttable}
      \begin{tabular}{ccc >{\columncolor{gray!15}} cc >{\columncolor{gray!15}} cc >{\columncolor{gray!15}} cc >{\columncolor{gray!15}} cc >{\columncolor{gray!15}} c}
        \toprule
        \multicolumn{2}{c}{Attention Layer} & \multicolumn{2}{c}{FullAttention} & \multicolumn{2}{c}{ProbAttention} & \multicolumn{2}{c}{AutoCorrelation} & \multicolumn{2}{c}{FlowAttention} & \multicolumn{2}{c}{PeriodAttention} \\
        \toprule
        Dataset                      & Length  & MSE            & MAE             & MSE             & MAE             & MSE              & MAE              & MSE             & MAE             & MSE              & MAE              \\
        \toprule
        \multirow{5}{*}{Traffic}     & 96     & \underline{\color{blue}0.386}           & \underline{\color{blue}0.258}           & 0.390           & 0.259           & 0.396            & 0.261            & {\bf\color{red}0.385}           & {\bf\color{red}0.256}           & 0.414            & 0.273            \\
                                     & 192    & {\bf\color{red}0.398}           & {\bf\color{red}0.263}           & 0.410           & 0.268           & 0.413            & 0.268            & \underline{\color{blue}0.406}           & \underline{\color{blue}0.265}           & 0.431            & 0.279            \\
                                     & 336    & {\bf\color{red}0.409}           & {\bf\color{red}0.270}           & 0.425           & 0.274           & 0.429            & 0.275            & \underline{\color{blue}0.424}           & \underline{\color{blue}0.274}           & 0.447            & 0.286            \\
                                     & 720    & {\bf\color{red}0.431}           & {\bf\color{red}0.287}           & \underline{\color{blue}0.459}           & \underline{\color{blue}0.293}           & 0.461            & 0.293            & \underline{\color{blue}0.459}           & \underline{\color{blue}0.293}           & 0.478            & 0.303            \\ \cline{2-12}
        \rowcolor{cyan!15}  \cellcolor{white} & Avg    & {\bf\color{red}0.406}           & {\bf\color{red}0.270}           & 0.421           & 0.274           & 0.425            & 0.274            & \underline{\color{blue}0.418}           & \underline{\color{blue}0.272}           & 0.443            & 0.285            \\ \toprule
        \multirow{5}{*}{Electricity} & 96     & 0.143           & 0.235           & {\bf\color{red}0.136}           & {\bf\color{red}0.229}           & 0.174            & 0.256            & \underline{\color{blue}0.141}           & \underline{\color{blue}0.233}           & 0.142            & 0.233            \\
                                     & 192    & 0.162           & 0.253           & {\bf\color{red}0.154}           & {\bf\color{red}0.246}           & 0.179            & 0.263            & 0.160           & 0.251           & \underline{\color{blue}0.157}            & \underline{\color{blue}0.247}            \\
                                     & 336    & 0.179           & 0.271           & \underline{\color{blue}0.172}           & 0.268           & 0.195            & 0.279            & 0.175           & \underline{\color{blue}0.267}           & {\bf\color{red}0.172}            & {\bf\color{red}0.264}            \\
                                     & 720    & 0.204           & 0.294           & 0.205           & 0.298           & 0.232            & 0.311            &  {\bf\color{red}0.203}           & \underline{\color{blue}0.294}           & \underline{\color{blue}0.204}            &  {\bf\color{red}0.293}            \\ \cline{2-12}
        \rowcolor{cyan!15}  \cellcolor{white} & Avg    & 0.172           & 0.263           &  {\bf\color{red}0.167}           & \underline{\color{blue}0.260}           & 0.195            & 0.277            & 0.170           & 0.261           & \underline{\color{blue}0.169}            & {\bf\color{red}0.259}            \\ \toprule
        \multirow{5}{*}{Weather}     & 96     & \underline{\color{blue}0.169}           & \underline{\color{blue}0.209}           & {\bf\color{red}0.159}           & {\bf\color{red}0.204}           & 0.179            & 0.220            & \underline{\color{blue}0.169}           & \underline{\color{blue}0.209}           & 0.170            & 0.212            \\
                                     & 192    & 0.220           & 0.254           & {\bf\color{red}0.207}           & {\bf\color{red}0.248}           & 0.228            & 0.261            & 0.220           & 0.255           & \underline{\color{blue}0.214}            & \underline{\color{blue}0.253}            \\
                                     & 336    & 0.276           & 0.296           & {\bf\color{red}0.265}           & {\bf\color{red}0.291}           & 0.288            & 0.304            & 0.276           & \underline{\color{blue}0.294}           & \underline{\color{blue}0.273}            & 0.296            \\
                                     & 720    & 0.354           & {\bf\color{red}0.346}           & {\bf\color{red}0.350}           & \underline{\color{blue}0.347}           & 0.367            & 0.356            & 0.354           & 0.348           & \underline{\color{blue}0.352}            & 0.347            \\ \cline{2-12}
        \rowcolor{cyan!15}  \cellcolor{white} & Avg    & 0.255           & \underline{\color{blue}0.276}           & {\bf\color{red}0.245}           & {\bf\color{red}0.273}           & 0.266            & 0.285            & 0.255           & \underline{\color{blue}0.276}           & \underline{\color{blue}0.252}            & 0.277            \\ 
        \bottomrule
        \end{tabular}
    \begin{tablenotes}
      \item[*] The input length $I$ is set as 96, while the prediction lengths $O \in $ \{96, 192, 336, 720\}. 
    \end{tablenotes}
    \end{threeparttable}
}
\end{table*}

\newpage
\section{Full results on ETT dataset}
\label{app_ett}
The ETT dataset records electricity data of four different granularities and types. 
We offer an in-depth comparison of Minusformer utilizing the complete ETT dataset to facilitate future research endeavors.
Detailed results are provided in Table \ref{tb_ettm}.
It is evident that Minusformer demonstrates excellent performance on the complete ETT dataset.

\begin{table*}[!ht]
    \centering
    \caption{Full Multivariate Forecasting Results on ETT dataset.}
    \label{tb_ettm}
    \resizebox{\textwidth}{!} 
    {
      \tiny
      \begin{threeparttable}
        \begin{tabular}{ccc >{\columncolor{gray!15}} cc >{\columncolor{gray!15}} cc >{\columncolor{gray!15}} cc >{\columncolor{gray!15}} cc >{\columncolor{gray!15}} cc >{\columncolor{gray!15}} cc >{\columncolor{gray!15}} c}
        \toprule
        \multicolumn{2}{c}{Models}      & \multicolumn{2}{c}{Minusformer-96} & \multicolumn{2}{c}{Periodformer-144$^{\diamondsuit}$} & \multicolumn{2}{c}{FEDformer-96} & \multicolumn{2}{c}{Autoformer-96} & \multicolumn{2}{c}{Informer-96} & \multicolumn{2}{c}{LogTrans-96} & \multicolumn{2}{c}{Reformer-96} \\
        \toprule
                            & Length & MSE              & MAE             & MSE             & MAE            & MSE           & MAE           & MSE            & MAE           & MSE           & MAE          & MSE           & MAE          & MSE           & MAE          \\
        \toprule
        \multirow{5}{*}{\rotatebox{90}{ETTh1}} & 96     & {\bf\color{red}0.370}             & {\bf\color{red}0.394}           & \underline{\color{blue} 0.375}           & \underline{\color{blue} 0.395}          & 0.395         & 0.424         & 0.449          & 0.459         & 0.865         & 0.713        & 0.878         & 0.74         & 0.837         & 0.728        \\
                            & 192    & \underline{\color{blue} 0.423}            & \underline{\color{blue} 0.427}           & {\bf\color{red}0.413}           & {\bf\color{red}0.421}          & 0.469         & 0.47          & 0.5            & 0.482         & 1.008         & 0.792        & 1.037         & 0.824        & 0.923         & 0.766        \\
                            & 336    & \underline{\color{blue} 0.465}            & \underline{\color{blue} 0.446}           & {\bf\color{red}0.443}           & {\bf\color{red}0.441}          & 0.530          & 0.499         & 0.521          & 0.496         & 1.107         & 0.809        & 1.238         & 0.932        & 1.097         & 0.835        \\
                            & 720    & {\bf\color{red}0.465}            & {\bf\color{red}0.464}           & \underline{\color{blue} 0.467}           & \underline{\color{blue} 0.469}          & 0.598         & 0.544         & 0.514          & 0.512         & 1.181         & 0.865        & 1.135         & 0.852        & 1.257         & 0.889        \\ \cline{2-16}
        \rowcolor{cyan!15}  \cellcolor{white} & Avg    & \underline{\color{blue} 0.431}            & \underline{\color{blue} 0.433}           & {\bf\color{red}0.425}           & {\bf\color{red}0.432}          & 0.498         & 0.484         & 0.496          & 0.487         & 1.040         & 0.795        & 1.072         & 0.837        & 1.029         & 0.805        \\ 
        \toprule
        \multirow{5}{*}{\rotatebox{90}{ETTh2}} & 96     & {\bf\color{red}0.291}            & {\bf\color{red}0.342}           & \underline{\color{blue} 0.313}           & \underline{\color{blue} 0.356}          & 0.394         & 0.414         & 0.358          & 0.397         & 3.755         & 1.525        & 2.116         & 1.197        & 2.626         & 1.317        \\
                            & 192    & {\bf\color{red}0.371}            & {\bf\color{red}0.391}           & \underline{\color{blue} 0.389}           & \underline{\color{blue} 0.405}          & 0.439         & 0.445         & 0.456          & 0.452         & 5.602         & 1.931        & 4.315         & 1.635        & 11.12         & 2.979        \\
                            & 336    & {\bf\color{red}0.412}            & {\bf\color{red}0.427}           & \underline{\color{blue} 0.418}           & \underline{\color{blue} 0.432}          & 0.482         & 0.48          & 0.482          & 0.486         & 4.721         & 1.835        & 1.124         & 1.604        & 9.323         & 2.769        \\
                            & 720    & {\bf\color{red}0.418}            & {\bf\color{red}0.438}           & \underline{\color{blue} 0.427}           & \underline{\color{blue} 0.444}          & 0.5           & 0.509         & 0.515          & 0.511         & 3.647         & 1.625        & 3.188         & 1.54         & 3.874         & 1.697        \\ \cline{2-16}
        \rowcolor{cyan!15}  \cellcolor{white} & Avg    & {\bf\color{red}0.373}            & {\bf\color{red}0.400}           & \underline{\color{blue} 0.387}           & \underline{\color{blue} 0.409}          & 0.454         & 0.462         & 0.453          & 0.462         & 4.431         & 1.729        & 2.686         & 1.494        & 6.736         & 2.191        \\
        \toprule
        \multirow{5}{*}{\rotatebox{90}{ETTm1}} & 96     & {\bf\color{red}0.317}            & {\bf\color{red}0.356}           & \underline{\color{blue} 0.337}           & \underline{\color{blue} 0.378}          & 0.378         & 0.418         & 0.505         & 0.475         & 0.672         & 0.571        & 0.600           & 0.546        & 0.538         & 0.528        \\
                            & 192    & {\bf\color{red}0.363}            & {\bf\color{red}0.379}           & \underline{\color{blue} 0.413}           & \underline{\color{blue} 0.431}          & 0.464         & 0.463         & 0.553          & 0.496         & 0.795         & 0.669        & 0.837         & 0.7          & 0.658         & 0.592        \\
                            & 336    & {\bf\color{red}0.397}            & {\bf\color{red}0.407}           & \underline{\color{blue} 0.428}           & \underline{\color{blue} 0.441}          & 0.508         & 0.487         & 0.621          & 0.537         & 1.212         & 0.871        & 1.124         & 0.832        & 0.898         & 0.721        \\
                            & 720    & {\bf\color{red}0.454}            & {\bf\color{red}0.442}           & \underline{\color{blue} 0.483}           & \underline{\color{blue} 0.483}          & 0.561         & 0.515         & 0.671          & 0.561         & 1.166         & 0.823        & 1.153         & 0.82         & 1.102         & 0.841        \\ \cline{2-16}
        \rowcolor{cyan!15}  \cellcolor{white} & Avg    & {\bf\color{red}0.383}            & {\bf\color{red}0.396}           & \underline{\color{blue} 0.415}           & \underline{\color{blue} 0.433}          & 0.478         & 0.471         & 0.588          & 0.517         & 0.961         & 0.734        & 0.929         & 0.725        & 0.799         & 0.671        \\
        \toprule
        \multirow{5}{*}{\rotatebox{90}{ETTm2}} & 96     & {\bf\color{red}0.177}            & {\bf\color{red}0.259}           & \underline{\color{blue} 0.186}           & \underline{\color{blue} 0.274}          & 0.204         & 0.288         & 0.255          & 0.339         & 0.365         & 0.453        & 0.768         & 0.642        & 0.658         & 0.619        \\
                            & 192    & {\bf\color{red}0.239}            & {\bf\color{red}0.299}           & \underline{\color{blue} 0.252}           & \underline{\color{blue} 0.317}          & 0.316         & 0.363         & 0.281          & 0.34          & 0.533         & 0.563        & 0.989         & 0.757        & 1.078         & 0.827        \\
                            & 336    & {\bf\color{red}0.298}            & {\bf\color{red}0.340}            & \underline{\color{blue} 0.311}           & \underline{\color{blue} 0.355}          & 0.359         & 0.387         & 0.339          & 0.372         & 1.363         & 0.887        & 1.334         & 0.872        & 1.549         & 0.972        \\
                            & 720    & {\bf\color{red}0.394}            & {\bf\color{red}0.394}           & \underline{\color{blue} 0.402}           & \underline{\color{blue} 0.405}          & 0.433         & 0.432         & 0.422          & 0.419         & 3.379         & 1.338        & 3.048         & 1.328        & 2.631         & 1.242        \\ \cline{2-16}
        \rowcolor{cyan!15}  \cellcolor{white} & Avg    & {\bf\color{red}0.277}            & {\bf\color{red}0.323}           & \underline{\color{blue} 0.288}           & \underline{\color{blue} 0.338}          & 0.328         & 0.368         & 0.324          & 0.368         & 1.410         & 0.810        & 1.535         & 0.900        & 1.479         & 0.915        \\
        \toprule
        \multicolumn{2}{c}{ $1^{\text{st}}$ Count} & {\bf\color{red} 17}           & {\bf\color{red}17}           & \underline{\color{blue}3}                   & \underline{\color{blue}3}       & 0               & 0               & 0             & 0               & 0               & 0               & 2             & 0             & 0               & 0             \\
        \bottomrule
        \end{tabular}
        \begin{tablenotes}
        \item[*] ${\diamondsuit}$ denotes the maximum search range of the input length.
      \end{tablenotes}
      \end{threeparttable}
  }
\end{table*}

\newpage
\section{Pseudocode of Minusformer}
\label{sec_algo}

To facilitate a comprehensive understanding of Minusformer's working principle, we offer detailed pseudocode outlining its implementation, as shown in Algorithm \ref{algo}.
The implementation presented here outlines the core ideas of Minusformer. 
For specific code implementation, please refer to this \href{https://drive.google.com/drive/folders/1JZ_ByOU7M5j6I5UM1ebkyLODlTOKin-J}{repository}.
It is evident that the deployment procedure of Minusformer exhibits a relative simplicity, characterized by the inclusion of several iteratively applied blocks.
This property renders it highly versatile for integrating newly devised Attention mechanisms or modules.
As demonstrated in Appendix \ref{app_improve_attn}, the substitution of Attention mechanisms in Minusformer with novel alternatives yields superior generalization.

\begin{algorithm}[htbp]
  \caption{Minusformer Architecture.}\label{algo}
  \begin{algorithmic}[1]
  \Require   
  Batch size $B$, input lookback time series $\mathbf{X}\in\mathbb{R}^{B \times I\times D}$; input length $I$; predicted length $O$; embedding dimension $E$; the number of the block $L$; the output length $H$ in each block.

    \State $\mathbf{X}_0=\texttt{StandardScaler}(\mathbf{X}^T)$ \Comment{$\mathbf{X}_0\in\mathbb{R}^{B \times D \times I}$}

    \State $\triangleright \ $ Apply a linear transformation to the temporal dimension of X to align it with the embedding dimension.

    \State $\mathbf{X}_{1}=\texttt{Linear}(\mathbf{X}_0)$ \  \Comment{$\mathbf{X}^{1}\in\mathbb{R}^{B \times D \times E}$} \  \Comment{The embedded X enters the backbone as an input stream.}

    \State $O_0 = 0$ \ \Comment{Set the initial value of the output stream to 0.}

    \State $\textbf{for}\ l\ \textbf{in}\ \{1,\cdots,L\}\textbf{:}$ \ \Comment{Run through the backbone.}

    \State $\textbf{\textcolor{white}{for}}$  $\triangleright \ $ Apply attention to the input stream.
    \State $\textbf{\textcolor{white}{for}} \  \mathbf{\hat{X}}_{l,1} = \texttt{Attention}(\mathbf{X}_{l,1})$ \ \Comment{$\mathbf{\hat{X}}_{l,1}\in\mathbb{R}^{B \times D \times E}$} 
    
    \State $\textbf{\textcolor{white}{for}}$  $\triangleright \ $ Subtracting the attention output from the input.
    \State $\textbf{\textcolor{white}{for}}\ \mathbf{R}_{l, 1} = \mathbf{X}_{l,1} - \delta \ \texttt{Dropout}(\mathbf{\hat{X}}_{l,1})$ \ 
    \Comment{$\mathbf{R}_{l, 1}\in\mathbb{R}^{B \times D \times E}$}
    
    \State $\textbf{\textcolor{white}{for}}$ $\triangleright \ $ LayerNorm is adopted to reduce attributes discrepancies.
    \State $\textbf{\textcolor{white}{for}} \mathbf{X}_{l,2} = \mathbf{R}_{l, 1} = \texttt{LayerNorm}(\mathbf{R}_{l,1}) $ 

    \State $\textbf{\textcolor{white}{for}}$ $\triangleright \ $The feedforward exclusively performs nonlinear transformations on the temporal aspect.
    \State $\textbf{\textcolor{white}{for}}\ \mathbf{\hat{X}}_{l,2} = \texttt{FeedForward}(\mathbf{R}_{l, 1})$ \ \Comment{$\mathbf{H}^{l}\in\mathbb{R}^{B \times D \times E}$}
    
    \State $\textbf{\textcolor{white}{for}}$  $\triangleright \ $ Subtracting the feedforward output from the input.
    \State $\textbf{\textcolor{white}{for}}\ \mathbf{R}_{l, 2} = \mathbf{X}_{l,2} - \mathbf{\hat{X}}_{l,2}$ \ \Comment{$\mathbf{R}_{l, 2}\in\mathbb{R}^{B \times D \times E}$}
    
    \State $\textbf{\textcolor{white}{for}}$  $\triangleright \ $ Add gate mechanism to input stream.
    \State $\textbf{\textcolor{white}{for}}\ \mathbf{X}_{l+1} = \texttt{Sigmoid}\big(\texttt{Linear} (\mathbf{R}_{l,2})\big) \cdot \texttt{Linear} (\mathbf{R}_{l,2})$ \ \Comment{$\mathbf{X}_{l+1}\in\mathbb{R}^{B \times D \times E}$}

    \State $\textbf{\textcolor{white}{for}}$  $\triangleright \ $ Add gate mechanism to output stream.
    \State $\textbf{\textcolor{white}{for}}\ \mathbf{\hat{O}}_{l+1} = \texttt{Sigmoid}\big(\texttt{Linear} ([\mathbf{\hat{X}}_{l,1}, \mathbf{\hat{X}}_{l,2}] )\big) \cdot \texttt{Linear} ([\mathbf{\hat{X}}_{l,1}, \mathbf{\hat{X}}_{l,2}])$ \ \Comment{$\mathbf{\hat{O}}_{l+1}\in\mathbb{R}^{B \times D \times H}$} 

    \State $\textbf{\textcolor{white}{for}}$  $\triangleright \ $ Subtract the previously learned output.
    \State $\textbf{\textcolor{white}{for}}\ \mathbf{O}_{l+1} = \mathbf{\hat{O}}_{l+1} - \mathbf{O}_{l}$ \ \Comment{$\mathbf{O}_{l+1}\in\mathbb{R}^{B \times D \times H}$}
    
    \State $\textbf{End for}$

    \State  $\triangleright \ $ Align the final output with the predicted length.
    \State $\textbf{if}\  H \neq O \ \ \textbf{then}\ $:
    \State $\textbf{\textcolor{white}{for}}$ $ \mathbf{O}_L = \texttt{Linear} (\mathbf{O}_{L-1})$ \ \Comment{$\mathbf{O}_{L}\in\mathbb{R}^{B \times D \times O}$}
    \State $ \textbf{Output}\ \texttt{InvertedScaler}(\mathbf{O}_L^T)$ \ \Comment{Output the final prediction results $\mathbf{O}_{L}^T\in\mathbb{R}^{B \times O \times D}$}
  \end{algorithmic} 
\end{algorithm}

\newpage
\section{Full univariate TS forecasting results}
\label{app_uts}

The full results for univariate TS forecasting are presented in Table \ref{tb4}. 
As other models, e.g., iTransformer \citep{liu2023itransformer}, PatchTST \citep{nie2022time} and Crossformer \citep{zhang2022crossformer} do not offer performance information for all prediction lengths, we compare our method with those that provide comprehensive performance analysis, including Periodformer \citep{liang2023does}, FEDformer \citep{zhou2022fedformer}, Autoformer \citep{wu2021autoformer}, Informer \citep{Zhou2021Informer}, LogTrans \citep{li2019enhancing} and Reformer \citep{Kitaev2020Reformer}.
Despite Periodformer being a model that determines optimal hyperparameters through search, the proposed method outperforms benchmark approaches by achieving the highest count of leading terms across various prediction lengths.
This reaffirms the effectiveness of Minusformer.

\begin{table*}[!ht]
  \centering
  \caption{Univariate TS forecasting results on benchmark datasets.}
  \label{tb4}
  \resizebox{0.95\textwidth}{!}
  {
  
  \begin{threeparttable}
  \begin{tabular}{ccc >{\columncolor{gray!15}} cc >{\columncolor{gray!15}} cc >{\columncolor{gray!15}} cc >{\columncolor{gray!15}} cc >{\columncolor{gray!15}} cc >{\columncolor{gray!15}} cc >{\columncolor{gray!15}} c}
    \toprule
    \multicolumn{2}{c}{Model}          & \multicolumn{2}{c}{Minusformer-96} & \multicolumn{2}{c}{Periodformer-144$^{\diamondsuit}$} & \multicolumn{2}{c}{FEDformer-96} & \multicolumn{2}{c}{Autoformer-96} & \multicolumn{2}{c}{Informer-96} & \multicolumn{2}{c}{LogTrans-96} & \multicolumn{2}{c}{Reformer-96} \\
    \toprule
                              & Length & MSE              & MAE             & MSE               & MAE               & MSE             & MAE            & MSE             & MAE             & MSE            & MAE            & MSE            & MAE            & MSE            & MAE            \\ \toprule
    \multirow{5}{*}{\rotatebox{90}{ETTh1}}       & 96  & {\bf\color{red}0.055}            & {\bf\color{red}0.177}           & \underline{\color{blue} 0.068}             & \underline{\color{blue} 0.203}             & 0.079           & 0.215          & 0.071           & 0.206           & 0.193          & 0.377          & 0.283          & 0.468          & 0.532          & 0.569          \\
                                & 192 & {\bf\color{red}0.072}            & {\bf\color{red}0.204}           & \underline{\color{blue} 0.088}             & \underline{\color{blue} 0.228}             & 0.104           & 0.245          & 0.114           & 0.262           & 0.217          & 0.395          & 0.234          & 0.409          & 0.568          & 0.575          \\
                                & 336 & {\bf\color{red}0.08}             & {\bf\color{red}0.219}           & \underline{\color{blue} 0.105}             & \underline{\color{blue} 0.256}             & 0.119           & 0.270           & 0.107           & 0.258           & 0.202          & 0.381          & 0.386          & 0.546          & 0.635          & 0.589          \\
                                & 720 & {\bf\color{red}0.079}            & {\bf\color{red}0.224}           & \underline{\color{blue} 0.109}             & \underline{\color{blue} 0.262}             & 0.142           & 0.299          & 0.126           & 0.283           & 0.183          & 0.355          & 0.475          & 0.628          & 0.762          & 0.666          \\ \cline{2-16}
    \rowcolor{cyan!15}  \cellcolor{white} & Avg & {\bf\color{red}0.072}            & {\bf\color{red}0.206}           & \underline{\color{blue} 0.093}             & \underline{\color{blue} 0.237}             & 0.111           & 0.257          & 0.105           & 0.252           & 0.199          & 0.377          & 0.345          & 0.513          & 0.624          & 0.600          \\ \toprule
    \multirow{5}{*}{\rotatebox{90}{ETTh2}}       & 96  & 0.129            & 0.275           & {\bf\color{red} 0.125}             & \underline{\color{blue}0.272}             & \underline{\color{blue} 0.128}           & {\bf\color{red} 0.271}          & 0.153           & 0.306           & 0.213          & 0.373          & 0.217          & 0.379          & 1.411          & 0.838          \\ 
                                & 192 & \underline{\color{blue} 0.178}            & \underline{\color{blue} 0.329}           & {\bf\color{red} 0.175}             & {\bf\color{red}0.329}             & 0.185           & 0.33           & 0.204           & 0.351           & 0.227          & 0.387          & 0.281          & 0.429          & 5.658          & 1.671          \\
                                & 336 & {\bf\color{red}0.211}            & {\bf\color{red}0.365}           & \underline{\color{blue} 0.219}             & \underline{\color{blue} 0.372}             & 0.231           & 0.378          & 0.246           & 0.389           & 0.242          & 0.401          & 0.293          & 0.437          & 4.777          & 1.582          \\
                                & 720 & \underline{\color{blue}0.220}            & {\bf\color{red}0.377}           &  0.249             &  0.400              & 0.278           & 0.42           & 0.268           & 0.409           & 0.291          & 0.439          & {\bf\color{red}0.218}          & \underline{\color{blue} 0.387}          & 2.042          & 1.039          \\ \cline{2-16}
   \rowcolor{cyan!15}  \cellcolor{white} & Avg & {\bf\color{red}0.185}            & {\bf\color{red}0.337}           & \underline{\color{blue} 0.192}             & \underline{\color{blue} 0.343}             & 0.206           & 0.350          & 0.218           & 0.364           & 0.243          & 0.400          & 0.252          & 0.408          & 3.472          & 1.283          \\ \toprule
    \multirow{5}{*}{\rotatebox{90}{ETTm1}}       & 96  & {\bf\color{red}0.029}            & {\bf\color{red}0.126}           & \underline{\color{blue} 0.033}             & \underline{\color{blue} 0.139}             & 0.033           & 0.140           & 0.056           & 0.183           & 0.109          & 0.277          & 0.049          & 0.171          & 0.296          & 0.355          \\ 
                                & 192 & {\bf\color{red}0.044}            & {\bf\color{red}0.158}           & \underline{\color{blue}0.052}             & \underline{\color{blue} 0.177}             &  0.058           & 0.186          & 0.081           & 0.216           & 0.151          & 0.310           & 0.157          & 0.317          & 0.429          & 0.474          \\
                                & 336 & {\bf\color{red}0.057}            & {\bf\color{red}0.185}           & \underline{\color{blue} 0.070}              & 0.267             & 0.084           &  0.231          & 0.076           & \underline{\color{blue}0.218}           & 0.427          & 0.591          & 0.289          & 0.459          & 0.585          & 0.583          \\
                                & 720 & {\bf\color{red}0.080}            & {\bf\color{red}0.218}           & \underline{\color{blue} 0.081}             & \underline{\color{blue} 0.221}             & 0.102           & 0.250           & 0.110            & 0.267           & 0.438          & 0.586          & 0.430           & 0.579          & 0.782          & 0.73           \\ \cline{2-16}
     \rowcolor{cyan!15}  \cellcolor{white} & Avg & {\bf\color{red}0.052}            & {\bf\color{red}0.172}           & \underline{\color{blue} 0.059}             & \underline{\color{blue} 0.201}             & 0.069          & 0.202          & 0.081           & 0.221           & 0.281          & 0.441          & 0.231          & 0.382          & 0.523          & 0.536          \\ \toprule
    \multirow{5}{*}{\rotatebox{90}{ETTm2}}       & 96  & 0.064            & {\bf\color{red} 0.180}          & {\bf\color{red} 0.060}             & \underline{\color{blue}0.182}             & \underline{\color{blue}0.063}           & 0.189          & 0.065           & 0.189           & 0.08           & 0.217          & 0.075          & 0.208          & 0.077          & 0.214          \\ 
                                & 192 & {\bf\color{red}0.099}            & {\bf\color{red}0.233}           & \underline{\color{blue}0.099}             & \underline{\color{blue}0.236}             & 0.110           & 0.252          & 0.118           & 0.256           & 0.112          & 0.259          & 0.129          & 0.275          & 0.138          & 0.290           \\
                                & 336 & {\bf\color{red}0.129}            & {\bf\color{red}0.273}           & \underline{\color{blue}0.129}             & \underline{\color{blue}0.275}             & 0.147           & 0.301          & 0.154           & 0.305           & 0.166          & 0.314          & 0.154          & 0.302          & 0.160          & 0.313          \\
                                & 720 & 0.180    & 0.329           & \underline{\color{blue}0.170}             & {\bf\color{red}0.317}             & 0.219           & 0.368          & 0.182           & 0.335           & 0.228      & 0.380           & {\bf\color{red}0.160}          & \underline{\color{blue}0.322}         & 0.168          & 0.334          \\ \cline{2-16}
    \rowcolor{cyan!15}  \cellcolor{white}& Avg & \underline{\color{blue}0.118}            & \underline{\color{blue}0.254}           & {\bf\color{red}0.115}             & {\bf\color{red}0.253}             & 0.135           & 0.278          & 0.130           & 0.271           & 0.147          & 0.293          & 0.130          & 0.277          & 0.136          & 0.288          \\ \toprule
    \multirow{5}{*}{\rotatebox{90}{Traffic}}     & 96  & {\bf\color{red}0.127}            & {\bf\color{red}0.202}           & \underline{\color{blue}0.143}             & \underline{\color{blue}0.222}             & 0.170           & 0.263          & 0.246           & 0.346           & 0.257          & 0.353          & 0.226          & 0.317          & 0.313          & 0.383          \\ 
                                & 192 & {\bf\color{red}0.135}            & {\bf\color{red}0.211}           & \underline{\color{blue}0.146}             & \underline{\color{blue}0.227}             & 0.173           & 0.265          & 0.266           & 0.37            & 0.299          & 0.376          & 0.314          & 0.408          & 0.386          & 0.453          \\
                                & 336 & {\bf\color{red}0.130}             & {\bf\color{red}0.215}           & \underline{\color{blue}0.147}             & \underline{\color{blue}0.231}             & 0.178           & 0.266          & 0.263           & 0.371           & 0.312          & 0.387          & 0.387          & 0.453          & 0.423          & 0.468          \\
                                & 720 & {\bf\color{red}0.135}            & {\bf\color{red}0.218}           & \underline{\color{blue}0.164}             & \underline{\color{blue}0.252}             & 0.187           & 0.286          & 0.269           & 0.372           & 0.366          & 0.436          & 0.437          & 0.491          & 0.378          & 0.433          \\ \cline{2-16}
     \rowcolor{cyan!15}  \cellcolor{white} & Avg & {\bf\color{red}0.132}            & {\bf\color{red}0.212}           & \underline{\color{blue}0.150}             & \underline{\color{blue}0.233}             & 0.177           & 0.270          & 0.261           & 0.365           & 0.309          & 0.388          & 0.341          & 0.417          & 0.375          & 0.434          \\ \toprule
    \multirow{5}{*}{\rotatebox{90}{Electricity}} & 96  & \underline{\color{blue}0.249}            & \underline{\color{blue}0.358}           & {\bf\color{red}0.236}             & {\bf\color{red}0.349}             & 0.262           & 0.378          & 0.341           & 0.438           & 0.258          & 0.367          & 0.288          & 0.393          & 0.275          & 0.379          \\
                                & 192 & 0.286            & \underline{\color{blue}0.379}           & {\bf\color{red}0.277}             & {\bf\color{red}0.369}             & 0.316           & 0.410          & 0.345           & 0.428           & \underline{\color{blue}0.285}          & 0.388          & 0.432          & 0.483          & 0.304          & 0.402          \\
                                & 336 & 0.337           & \underline{\color{blue}0.413}           & {\bf\color{red}0.324}             & {\bf\color{red}0.400}               & 0.361           & 0.445          & 0.406           & 0.470          & \underline{\color{blue}0.336}          & 0.423          & 0.430          & 0.483          & 0.37           & 0.448          \\
                                & 720 & \underline{\color{blue}0.385}            & \underline{\color{blue}0.454}           & {\bf\color{red}0.353}             & {\bf\color{red}0.437}             & 0.448           & 0.501          & 0.565           & 0.581           & 0.607          & 0.599          & 0.491          & 0.531          & 0.46           & 0.511          \\ \cline{2-16}
    \rowcolor{cyan!15}  \cellcolor{white}& Avg & \underline{\color{blue}0.314}            & \underline{\color{blue}0.401}           & {\bf\color{red}0.298}             & {\bf\color{red}0.389}             & 0.347           & 0.434          & 0.414           & 0.479           & 0.372          & 0.444          & 0.410          & 0.473          & 0.352          & 0.435          \\ \toprule
    \multirow{5}{*}{\rotatebox{90}{Weather}}     & 96  & {\bf\color{red}0.0012}           & {\bf\color{red}0.0263}          & \underline{\color{blue}0.0015}            & \underline{\color{blue}0.0300}              & 0.0035          & 0.046          & 0.0110          & 0.081           & 0.004          & 0.044          & 0.0046         & 0.052          & 0.012          & 0.087          \\
                                & 192 & {\bf\color{red}0.0014}           & {\bf\color{red}0.0283}          & \underline{\color{blue}0.0015}            & \underline{\color{blue}0.0307}            & 0.0054          & 0.059          & 0.0075          & 0.067           & 0.002          & 0.040          & 0.006          & 0.060          & 0.0098         & 0.044          \\
                                & 336 & {\bf\color{red}0.0015}           & {\bf\color{red}0.0294}          & \underline{\color{blue}0.0017}            & \underline{\color{blue}0.0313}            & 0.008           & 0.072          & 0.0063          & 0.062           & 0.004          & 0.049          & 0.006          & 0.054          & 0.013          & 0.100          \\
                                & 720 & {\bf\color{red}0.002}            & {\bf\color{red}0.0333}         & \underline{\color{blue}0.0020}             & \underline{\color{blue}0.0348}            & 0.015           & 0.091          & 0.0085          & 0.070           & 0.003          & 0.042          & 0.007          & 0.059          & 0.011          & 0.083          \\ \cline{2-16}
     \rowcolor{cyan!15}  \cellcolor{white}& Avg & {\bf\color{red}0.0015}           & {\bf\color{red}0.0293}          & \underline{\color{blue}0.0017}            & \underline{\color{blue}0.0317}            & 0.008           & 0.067          & 0.0083          & 0.0700          & 0.0033         & 0.0438         & 0.0059         & 0.0563         & 0.0115         & 0.0785         \\ \toprule
    \multirow{5}{*}{\rotatebox{90}{Exchange}}    & 96  & \underline{\color{blue}0.096}            & \underline{\color{blue}0.226}           & {\bf\color{red}0.092}             & {\bf\color{red}0.226}             & 0.131           & 0.284          & 0.241           & 0.387           & 1.327          & 0.944          & 0.237          & 0.377          & 0.298          & 0.444          \\
                                & 192 & \underline{\color{blue}0.200}              & {\bf\color{red}0.332}           & {\bf\color{red}0.198}             &  \underline{\color{blue}0.341}             &  0.277           & 0.420          & 0.300           & 0.369           & 1.258          & 0.924          & 0.738          & 0.619          & 0.777          & 0.719          \\
                                & 336 & \underline{\color{blue}0.400}              & \underline{\color{blue}0.473}           & {\bf\color{red}0.370}             & {\bf\color{red}0.471}             & 0.426           & 0.511          & 0.509           & 0.524           & 2.179          & 1.296          & 2.018          & 1.0700         & 1.833          & 1.128          \\
                                & 720 & \underline{\color{blue}1.020}             & \underline{\color{blue}0.779}           & {\bf\color{red}0.753}             & {\bf\color{red}0.696}             & 1.162           & 0.832          & 1.260           & 0.867           & 1.28           & 0.953          & 2.405          & 1.175          & 1.203          & 0.956          \\ \cline{2-16}
    \rowcolor{cyan!15}  \cellcolor{white} & Avg & \underline{\color{blue}0.429}            & \underline{\color{blue}0.453}           & {\bf\color{red}0.353}             & {\bf\color{red}0.434}             & 0.499           & 0.512          & 0.578           & 0.537           & 1.511          & 1.029          & 1.350          & 0.810          & 1.028          & 0.812          \\ \toprule
    \multicolumn{2}{c}{ $1^{\text{st}}$ Count} & {\bf\color{red} 24}           & {\bf\color{red}27}           & \underline{\color{blue}14}                   & \underline{\color{blue}12}       & 0               & 1               & 0             & 0               & 0               & 0               & 2             & 0             & 0               & 0             \\
    \bottomrule
    \end{tabular}
  \begin{tablenotes}
    \item[*] ${\diamondsuit}$ denotes the maximum search range of the input length.
  \end{tablenotes}
  \end{threeparttable}
}
\end{table*}

\newpage
\section{Evaluation metrics}
\label{secsub_metrics}

This paper uses a variety of metrics, including MSE (Mean Square Error), MAE (Mean Absolute Error), RMSP (Root Median Square Percent), MAPE (Mean Absolute Percentage Error), sMAPE (symmetric MAPE), MASE (Mean Absolute Scaled Error) and Quantile loss. 
All evaluation metrics adopted in this paper are as follows:

\begin{align}
  \text{MSE}(Y, \hat{Y}) &= \frac{1}{N} \sum_{i=1}^{N} (y_i - \hat{y}_i)^2 , \\
  \text{MAE}(Y, \hat{Y}) &= \frac{1}{N} \sum_{i=1}^{N} |y_i - \hat{y}_i| , \\
  \text{RMSP}(Y, \hat{Y}) &= \sqrt{ \text{Median} \left((\frac{y_i - \hat{y}_i}{y_i})^2 \right) } ,\\
  \text{MAPE}(Y, \hat{Y}) &= \frac{1}{N} \sum_{i=1}^{N}  \frac{|y_i - \hat{y}_i|}{|y_i|} ,\\
  \text{sMAPE}(Y, \hat{Y}) &= \frac{2}{N} \sum_{i=1}^{N} \frac{|y_i - \hat{y}_i|}{|y_i| + |\hat{y}_i|} ,\\
  \text{MASE}(Y, \hat{Y}) &= \frac{1}{N} \sum_{i=1}^{N} \frac{|y_i - \hat{y}_i|}{\frac{1}{N-m} \sum_{i=m+1}^{N} |y_i - y_{i-m}|} ,\\
  \text{QuantileLoss}(Y, \hat{Y}, q) &= \frac{1}{N} \sum_{i=1}^{N} \mathbb{I}_{\hat{y}_i \ge y_i} (1- q) |y_i - \hat{y}_i| + \mathbb{I}_{\hat{y}_i < y_i}q |y_i - \hat{y}_i|.
\end{align}

\newpage
\section{Full TS forecasting on Monash's repository}
\label{subsec_mts_monash}

The full results for mulvariate and univariate TS forecasting on 10 Monash TS datasets are presented in Table \ref{tb4}. 
Due to different characteristics of the dataset, the output lengths Y1, Y2, Y3 and Y4,  are slightly different. Specifically, for Traffic, Electricity, Solar, OiK\_Weather, Sunspot, Saugeenday and M4, the input length is 96, and the output lengths are 96, 192, 336, and 720, respectively. 
For ILI dataset, the input length is 36, and the output lengths are 24, 36, 48, and 60, respectively. 
For NN5 and Rideshare datasets, the input length is 48, and the output lengths are 12, 24, 36, and 48, respectively. 
It is evident that the proposed method outperforms benchmark approaches by achieving the highest count of leading terms across various prediction lengths.
This reaffirms the effectiveness of Minusformer.

\begin{table}[!ht]
  \centering
  \caption{Mulvariate and univariate forecasting results with diverse metrics on Monash TS datasets.}
  \label{tb_mts_monash_full}
  \resizebox{\textwidth}{!}
  {
    \Huge
  \begin{tabular}{c|c|cccc >{\columncolor{cyan!15}} c| >{\columncolor{green!10}} c| cccc >{\columncolor{cyan!15}} c| cccc >{\columncolor{cyan!15}} c| cccc >{\columncolor{cyan!15}} c| cccc >{\columncolor{cyan!15}} c}
    \toprule
  \multicolumn{2}{c|}{Mulvariate}              & \multicolumn{6}{c|}{Minsuforemr-96}    & \multicolumn{5}{c|}{iTransformer-96}             & \multicolumn{5}{c|}{DLinear-96}       & \multicolumn{5}{c|}{Autoformer-96}      & \multicolumn{5}{c}{Informer-96}       \\ \toprule
  Dataset                       & Metric & Y1    & Y2    & Y3    & Y4    & Avg   & IMP     & Y1    & Y2    & Y3    & Y4    & Avg   & Y1    & Y2    & Y3    & Y4    & Avg   & Y1     & Y2    & Y3    & Y4    & Avg   & Y1    & Y2    & Y3    & Y4    & Avg   \\ \toprule
  \multirow{8}{*}{\rotatebox{90}{Traffic}}      
                                & MSE    & 0.386 & 0.398 & 0.409 & 0.431 & {\bf\color{red}0.406} & 5.14\%  & 0.395 & 0.417 & 0.433 & 0.467 & \underline{\color{blue}0.428} & 0.650 & 0.598 & 0.605 & 0.645 & 0.625 & 0.613  & 0.616 & 0.622 & 0.660 & 0.628 & 0.719 & 0.696 & 0.777 & 0.864 & 0.764 \\
                                & MAE    & 0.258 & 0.263 & 0.270 & 0.287 & {\bf\color{red}0.270} & 4.26\%  & 0.268 & 0.276 & 0.283 & 0.302 & \underline{\color{blue}0.282} & 0.396 & 0.370 & 0.373 & 0.394 & 0.383 & 0.388  & 0.382 & 0.337 & 0.408 & 0.379 & 0.391 & 0.379 & 0.420 & 0.472 & 0.416 \\
                                & RMSP & 0.205 & 0.200 & 0.207 & 0.231 & {\bf\color{red}0.211} & 6.64\%  & 0.214 & 0.219 & 0.225 & 0.247 & \underline{\color{blue}0.226} & 0.350 & 0.317 & 0.322 & 0.356 & 0.336 & 0.437  & 0.409 & 0.384 & 0.432 & 0.416 & 0.423 & 0.416 & 0.522 & 0.715 & 0.519 \\
                                & MAPE   & 2.657 & 2.694 & 2.715 & 2.797 & {\bf\color{red}2.716} & 8.24\%  & 2.885 & 2.955 & 2.942 & 3.059 & \underline{\color{blue}2.960} & 5.018 & 4.477 & 4.290 & 4.196 & 4.495 & 4.377  & 4.198 & 4.254 & 4.288 & 4.279 & 4.760 & 4.677 & 6.115 & 7.016 & 5.642 \\
                                & sMAPE  & 0.479 & 0.476 & 0.485 & 0.514 & {\bf\color{red}0.489} & 2.59\%  & 0.485 & 0.494 & 0.501 & 0.529 & \underline{\color{blue}0.502} & 0.662 & 0.625 & 0.629 & 0.663 & 0.645 & 0.721  & 0.687 & 0.657 & 0.718 & 0.696 & 0.718 & 0.712 & 0.810 & 0.978 & 0.805 \\
                                & MASE   & 0.256 & 0.257 & 0.264 & 0.285 & {\bf\color{red}0.266} & 1.12\%  & 0.255 & 0.263 & 0.268 & 0.288 & \underline{\color{blue}0.269} & 0.397 & 0.365 & 0.366 & 0.390 & 0.380 & 0.405  & 0.390 & 0.364 & 0.401 & 0.390 & 0.415 & 0.406 & 0.468 & 0.595 & 0.471 \\
                                & Q25    & 0.255 & 0.264 & 0.268 & 0.281 & {\bf\color{red}0.267} & 4.64\%  & 0.271 & 0.279 & 0.279 & 0.290 & \underline{\color{blue}0.280} & 0.380 & 0.357 & 0.357 & 0.373 & 0.367 & 0.440  & 0.414 & 0.395 & 0.419 & 0.417 & 0.397 & 0.372 & 0.461 & 0.614 & 0.461 \\
                                & Q75    & 0.270 & 0.270 & 0.280 & 0.304 & {\bf\color{red}0.281} & 1.40\%  & 0.267 & 0.276 & 0.286 & 0.311 & \underline{\color{blue}0.285} & 0.412 & 0.383 & 0.390 & 0.417 & 0.401 & 0.391  & 0.387 & 0.374 & 0.416 & 0.392 & 0.456 & 0.465 & 0.536 & 0.621 & 0.520 \\ \toprule
  \multirow{8}{*}{\rotatebox{90}{Electricity}}
                                & MSE    & 0.143 & 0.162 & 0.179 & 0.204 & {\bf\color{red}0.172} & 3.37\%  & 0.148 & 0.162 & 0.178 & 0.225 & \underline{\color{blue}0.178} & 0.197 & 0.196 & 0.209 & 0.245 & 0.212 & 0.201  & 0.222 & 0.231 & 0.254 & 0.227 & 0.274 & 0.296 & 0.300 & 0.373 & 0.311 \\
                                & MAE    & 0.235 & 0.253 & 0.271 & 0.294 & {\bf\color{red}0.263} & 2.59\%  & 0.240 & 0.253 & 0.269 & 0.317 & \underline{\color{blue}0.270} & 0.282 & 0.285 & 0.301 & 0.333 & 0.300 & 0.317  & 0.334 & 0.338 & 0.361 & 0.338 & 0.368 & 0.386 & 0.394 & 0.439 & 0.397 \\
                                & RMSP & 0.215 & 0.230 & 0.253 & 0.287   & {\bf\color{red}0.246} & 2.38\%  & 0.223 & 0.239 & 0.256 & 0.290 & \underline{\color{blue}0.252} & 0.310 & 0.316 & 0.334 & 0.370 & 0.333 & 0.324  & 0.346 & 0.367 & 0.374 & 0.353 & 0.429 & 0.796 & 0.487 & 0.680 & 0.598 \\
                                & MAPE   & 2.296 & 2.431 & 2.699 & 3.194 & {\bf\color{red}2.655} & 2.10\%  & 2.493 & 2.732 & 2.713 & 2.910 & \underline{\color{blue}2.712} & 2.695 & 2.715 & 2.667 & 2.760 & 2.709 & 3.515  & 3.309 & 3.626 & 3.401 & 3.463 & 3.789 & 3.354 & 2.354 & 3.826 & 3.331 \\
                                & sMAPE  & 0.466 & 0.487 & 0.514 & 0.558 & {\bf\color{red}0.506} & 1.75\%  & 0.479 & 0.498 & 0.519 & 0.564 & \underline{\color{blue}0.515} & 0.598 & 0.601 & 0.621 & 0.660 & 0.620 & 0.584  & 0.619 & 0.639 & 0.642 & 0.621 & 0.711 & 1.148 & 0.786 & 0.988 & 0.908 \\
                                & MASE   & 0.257 & 0.276 & 0.293 & 0.321 & {\bf\color{red}0.287} & 3.37\%  & 0.267 & 0.284 & 0.298 & 0.340 & \underline{\color{blue}0.297} & 0.357 & 0.359 & 0.374 & 0.405 & 0.374 & 0.329  & 0.359 & 0.380 & 0.379 & 0.362 & 0.414 & 0.832 & 0.445 & 0.614 & 0.576 \\
                                & Q25    & 0.232 & 0.239 & 0.270 & 0.321 & 0.266 & -1.53\% & 0.232 & 0.255 & 0.269 & 0.290 & {\bf\color{red}0.262} & 0.314 & 0.317 & 0.327 & 0.356 & 0.329 & 0.349  & 0.342 & 0.378 & 0.384 & 0.363 & 0.406 & 0.665 & 0.421 & 0.590 & 0.521 \\
                                & Q75    & 0.230 & 0.254 & 0.261 & 0.276 & {\bf\color{red}0.255} & 7.27\%  & 0.250 & 0.258 & 0.272 & 0.318 & \underline{\color{blue}0.275} & 0.319 & 0.323 & 0.339 & 0.370 & 0.338 & 0.285  & 0.334 & 0.341 & 0.338 & 0.325 & 0.411 & 0.663 & 0.420 & 0.566 & 0.515 \\ \toprule
  \multirow{8}{*}{\rotatebox{90}{Solar}}
                                & MSE    & 0.192 & 0.230 & 0.243 & 0.243 & {\bf\color{red}0.227} & 2.58\%  & 0.203 & 0.233 & 0.248 & 0.249 & \underline{\color{blue}0.233} & 0.290 & 0.320 & 0.353 & 0.356 & 0.330 & 0.884  & 0.834 & 0.941 & 0.882 & 0.885 & 0.236 & 0.217 & 0.249 & 0.241 & 0.235 \\ 
                                & MAE    & 0.222 & 0.251 & 0.263 & 0.265 & {\bf\color{red}0.250} & 4.58\%  & 0.237 & 0.261 & 0.273 & 0.275 & \underline{\color{blue}0.262} & 0.378 & 0.398 & 0.415 & 0.413 & 0.401 & 0.711  & 0.692 & 0.723 & 0.717 & 0.711 & 0.259 & 0.269 & 0.283 & 0.317 & 0.280 \\ 
                                & RMSP & 0.061 & 0.060 & 0.072 & 0.090   & {\bf\color{red}0.071} & 19.32\% & 0.080 & 0.089 & 0.090 & 0.093 & \underline{\color{blue}0.088} & 0.363 & 0.374 & 0.389 & 0.385 & 0.378 & 0.449  & 0.563 & 0.694 & 0.868 & 0.644 & 0.059 & 0.075 & 0.105 & 0.150 & 0.097 \\
                                & MAPE   & 1.813 & 1.948 & 1.984 & 2.081 & {\bf\color{red}1.957} & 2.25\%  & 1.842 & 2.052 & 1.986 & 2.129 & \underline{\color{blue}2.002} & 2.225 & 2.378 & 2.558 & 2.633 & 2.449 & 2.349  & 2.403 & 2.509 & 2.042 & 2.326 & 2.200 & 2.471 & 2.775 & 2.531 & 2.494 \\
                                & sMAPE  & 0.381 & 0.426 & 0.437 & 0.443 & {\bf\color{red}0.422} & 32.15\% & 0.394 & 0.416 & 0.422 & 1.256 & \underline{\color{blue}0.622} & 0.688 & 0.714 & 0.727 & 0.714 & 0.711 & 0.767  & 0.861 & 1.062 & 1.260 & 0.988 & 0.380 & 0.425 & 0.465 & 0.517 & 0.447 \\
                                & MASE   & 0.286 & 0.379 & 0.377 & 0.374 & {\bf\color{red}0.354} & 28.19\% & 0.301 & 0.369 & 0.362 & 0.939 & \underline{\color{blue}0.493} & 0.436 & 0.511 & 0.519 & 0.507 & \underline{\color{blue}0.493} & 0.562  & 0.670 & 0.817 & 0.905 & 0.739 & 0.273 & 0.344 & 0.363 & 0.415 & 0.349 \\
                                & Q25    & 0.225 & 0.242 & 0.254 & 0.258 & {\bf\color{red}0.245} & 32.13\% & 0.223 & 0.255 & 0.250 & 0.716 & \underline{\color{blue}0.361} & 0.449 & 0.476 & 0.496 & 0.491 & 0.478 & 0.453  & 0.519 & 0.647 & 0.712 & 0.583 & 0.229 & 0.272 & 0.327 & 0.329 & 0.289 \\
                                & Q75    & 0.230 & 0.283 & 0.294 & 0.301 & {\bf\color{red}0.277} & 25.94\% & 0.250 & 0.253 & 0.277 & 0.715 & \underline{\color{blue}0.374} & 0.308 & 0.321 & 0.336 & 0.336 & 0.325 & 0.504  & 0.544 & 0.636 & 0.682 & 0.592 & 0.229 & 0.262 & 0.280 & 0.333 & 0.276 \\ \toprule
  \multirow{8}{*}{\rotatebox{90}{ILI}}
                                & MSE    & 2.065 & 1.917 & 1.966 & 2.114 & {\bf\color{red}2.016} & 10.88\% & 2.329 & 2.189 & 2.217 & 2.314 & \underline{\color{blue}2.262} & 3.052 & 2.804 & 2.829 & 2.973 & 2.915 & 3.410  & 3.365 & 3.125 & 2.847 & 3.187 & 5.421 & 5.001 & 5.098 & 5.312 & 5.208 \\ 
                                & MAE    & 0.835 & 0.852 & 0.854 & 0.900 & {\bf\color{red}0.860} & 10.23\% & 0.939 & 0.946 & 0.956 & 0.989 & \underline{\color{blue}0.958} & 1.245 & 1.153 & 1.161 & 1.193 & 1.188 & 1.296  & 1.252 & 1.200 & 1.146 & 1.224 & 1.606 & 1.549 & 1.564 & 1.584 & 1.576 \\ 
                                & RMSP & 0.341 & 0.371 & 0.371 & 0.386   & {\bf\color{red}0.367} & 11.57\% & 0.393 & 0.413 & 0.421 & 0.434 & \underline{\color{blue}0.415} & 0.608 & 0.550 & 0.561 & 0.576 & 0.574 & 0.643  & 0.590 & 0.581 & 0.570 & 0.596 & 0.827 & 0.791 & 0.778 & 0.773 & 0.792 \\
                                & MAPE   & 3.701 & 3.193 & 3.054 & 2.874 & 3.206 & -3.05\% & 3.638 & 3.169 & 2.937 & 2.701 & {\bf\color{red}3.111} & 4.334 & 2.953 & 2.662 & 2.504 & \underline{\color{blue}3.113} & 5.282  & 4.880 & 4.463 & 3.753 & 4.595 & 2.545 & 2.625 & 2.545 & 2.260 & 2.494 \\
                                & sMAPE  & 0.633 & 0.657 & 0.649 & 0.660 & {\bf\color{red}0.650} & 7.54\%  & 0.683 & 0.702 & 0.707 & 0.718 & \underline{\color{blue}0.703} & 0.903 & 0.840 & 0.849 & 0.866 & 0.865 & 0.957  & 0.897 & 0.899 & 0.882 & 0.909 & 1.203 & 1.175 & 1.172 & 1.181 & 1.183 \\
                                & MASE   & 0.557 & 0.520 & 0.549 & 0.668 & {\bf\color{red}0.574} & 11.42\% & 0.622 & 0.587 & 0.629 & 0.753 & \underline{\color{blue}0.648} & 0.860 & 0.715 & 0.726 & 0.861 & 0.791 & 0.777  & 0.718 & 0.716 & 0.747 & 0.740 & 0.932 & 0.856 & 0.861 & 0.980 & 0.907 \\
                                & Q25    & 0.821 & 0.761 & 0.742 & 0.740 & {\bf\color{red}0.766} & 5.20\%  & 0.873 & 0.798 & 0.789 & 0.771 & \underline{\color{blue}0.808} & 0.986 & 0.825 & 0.804 & 0.792 & 0.852 & 1.248  & 1.150 & 1.054 & 0.941 & 1.098 & 0.824 & 0.833 & 0.857 & 0.876 & 0.848 \\
                                & Q75    & 0.849 & 0.943 & 0.966 & 1.061 & {\bf\color{red}0.955} & 13.73\% & 1.005 & 1.094 & 1.123 & 1.207 & \underline{\color{blue}1.107} & 1.503 & 1.480 & 1.518 & 1.593 & 1.524 & 1.345  & 1.354 & 1.346 & 1.350 & 1.349 & 2.388 & 2.265 & 2.271 & 2.293 & 2.304 \\ \toprule
  \multirow{8}{*}{\rotatebox{90}{OiK\_Weather}} 
                                & MSE    & 0.631 & 0.686 & 0.706 & 0.733 & \underline{\color{blue}0.689} & 4.31\%  & 0.672 & 0.718 & 0.734 & 0.755 & 0.720 & 0.662 & 0.697 & 0.713 & 0.731 & 0.701 & 0.733  & 0.820 & 0.899 & 0.960 & 0.853 & 0.569 & 0.647 & 0.713 & 0.752 & {\bf\color{red}0.670} \\
                                & MAE    & 0.582 & 0.616 & 0.630 & 0.645 & {\bf\color{red}0.618} & 2.22\%  & 0.601 & 0.630 & 0.642 & 0.654 & 0.632 & 0.613 & 0.635 & 0.645 & 0.656 & 0.637 & 0.642  & 0.696 & 0.739 & 0.769 & 0.712 & 0.560 & 0.617 & 0.660 & 0.662 & \underline{\color{blue}0.625} \\
                                & RMSP & 0.712 & 0.770 & 0.790 & 0.820   & {\bf\color{red}0.773} & 1.02\%  & 0.727 & 0.776 & 0.800 & 0.820 & \underline{\color{blue}0.781} & 0.764 & 0.800 & 0.817 & 0.837 & 0.805 & 0.777  & 0.865 & 0.925 & 0.966 & 0.883 & 0.712 & 0.787 & 0.855 & 0.825 & 0.795 \\
                                & MAPE   & 4.005 & 4.934 & 5.270 & 4.378 & {\bf\color{red}4.647} & 41.81\% & 9.158 & 7.506 & 7.768 & 7.510 & 7.986 & 5.019 & 4.616 & 4.613 & 4.400 & 4.662 & 10.976 & 8.417 & 9.650 & 9.859 & 9.726 & 6.597 & 7.261 & 7.739 & 7.372 & 7.242 \\
                                & sMAPE  & 1.024 & 1.079 & 1.101 & 1.125 & {\bf\color{red}1.082} & 0.73\%  & 1.040 & 1.085 & 1.110 & 1.126 & \underline{\color{blue}1.090} & 1.133 & 1.178 & 1.200 & 1.229 & 1.185 & 1.087  & 1.175 & 1.243 & 1.299 & 1.201 & 1.035 & 1.123 & 1.201 & 1.143 & 1.126 \\
                                & MASE   & 0.897 & 0.888 & 0.889 & 0.876 & \underline{\color{blue}0.888} & 5.03\%  & 0.947 & 0.921 & 0.939 & 0.931 & 0.935 & 0.966 & 0.944 & 0.935 & 0.925 & 0.943 & 0.878  & 0.943 & 1.042 & 1.103 & 0.992 & 0.781 & 0.814 & 0.868 & 0.807 & {\bf\color{red}0.818} \\
                                & Q25    & 0.563 & 0.605 & 0.629 & 0.638 & {\bf\color{red}0.609} & 3.03\%  & 0.602 & 0.624 & 0.633 & 0.652 & 0.628 & 0.609 & 0.631 & 0.641 & 0.652 & 0.633 & 0.668  & 0.702 & 0.739 & 0.768 & 0.719 & 0.547 & 0.608 & 0.642 & 0.645 & \underline{\color{blue}0.611} \\
                                & Q75    & 0.600 & 0.628 & 0.631 & 0.653 & {\bf\color{red}0.628} & 1.26\%  & 0.600 & 0.636 & 0.651 & 0.655 & \underline{\color{blue}0.636} & 0.616 & 0.639 & 0.649 & 0.661 & 0.641 & 0.617  & 0.690 & 0.739 & 0.771 & 0.704 & 0.574 & 0.626 & 0.679 & 0.680 & 0.640 \\ \toprule
  \multirow{8}{*}{\rotatebox{90}{NN5}}
                                & MSE    & 0.798 & 0.737 & 0.680 & 0.659 & {\bf\color{red}0.719} & 0.55\%  & 0.793 & 0.745 & 0.685 & 0.668 &  \underline{\color{blue}0.723} & 1.474 & 1.484 & 1.406 & 1.429 & 1.448 & 0.891  & 0.915 & 0.739 & 0.859 & 0.851 & 0.969 & 0.983 & 0.939 & 0.977 & 0.967 \\
                                & MAE    & 0.590 & 0.582 & 0.570 & 0.568 & {\bf\color{red}0.578} & 0.86\%  & 0.593 & 0.589 & 0.574 & 0.574 &  \underline{\color{blue}0.583} & 0.929 & 0.943 & 0.916 & 0.927 & 0.929 & 0.659  & 0.683 & 0.617 & 0.678 & 0.659 & 0.714 & 0.734 & 0.725 & 0.733 & 0.727 \\
                                & RMSP & 0.526 & 0.539 & 0.545 & 0.554   & {\bf\color{red}0.541} & 2.35\%  & 0.536 & 0.554 & 0.557 & 0.570 &  \underline{\color{blue}0.554} & 0.942 & 0.988 & 0.964 & 0.980 & 0.969 & 0.614  & 0.653 & 0.606 & 0.669 & 0.636 & 0.702 & 0.750 & 0.749 & 0.753 & 0.739 \\
                                & MAPE   & 2.332 & 2.563 & 2.426 & 2.431 & {\bf\color{red}2.438} & 5.28\%  & 2.491 & 2.640 & 2.560 & 2.605 &  \underline{\color{blue}2.574} & 3.845 & 3.785 & 3.848 & 3.763 & 3.810 & 2.700  & 2.936 & 2.642 & 2.760 & 2.760 & 2.698 & 2.893 & 2.823 & 3.006 & 2.855 \\
                                & sMAPE  & 0.848 & 0.862 & 0.863 & 0.871 & {\bf\color{red}0.861} & 1.82\%  & 0.860 & 0.877 & 0.878 & 0.892 &  \underline{\color{blue}0.877} & 1.276 & 1.326 & 1.296 & 1.316 & 1.304 & 0.954  & 0.992 & 0.941 & 1.014 & 0.975 & 1.046 & 1.092 & 1.096 & 1.088 & 1.081 \\
                                & MASE   & 0.528 & 0.541 & 0.530 & 0.519 & {\bf\color{red}0.530} & 2.21\%  & 0.538 & 0.553 & 0.543 & 0.534 &  \underline{\color{blue}0.542} & 0.820 & 0.856 & 0.830 & 0.833 & 0.835 & 0.589  & 0.623 & 0.572 & 0.628 & 0.603 & 0.626 & 0.685 & 0.670 & 0.667 & 0.662 \\
                                & Q25    & 0.598 & 0.607 & 0.594 & 0.587 & {\bf\color{red}0.597} & 0.67\%  & 0.601 & 0.611 & 0.597 & 0.593 &  \underline{\color{blue}0.601} & 0.876 & 0.897 & 0.867 & 0.880 & 0.880 & 0.679  & 0.708 & 0.644 & 0.698 & 0.682 & 0.730 & 0.730 & 0.748 & 0.717 & 0.731 \\
                                & Q75    & 0.582 & 0.558 & 0.546 & 0.550 & {\bf\color{red}0.559} & 1.06\%  & 0.585 & 0.568 & 0.550 & 0.555 &  \underline{\color{blue}0.565} & 0.982 & 0.989 & 0.966 & 0.974 & 0.978 & 0.640  & 0.659 & 0.590 & 0.659 & 0.637 & 0.698 & 0.739 & 0.702 & 0.748 & 0.722 \\ \toprule
  \multirow{7}{*}{\rotatebox{90}{Rideshare}}
                                & MSE    & 0.225 & 0.261 & 0.467 & 0.504 & {\bf\color{red}0.364} & 26.61\% & 0.247 & 0.273 & 0.580 & 0.885 &  \underline{\color{blue}0.496} & 0.933 & 1.148 & 1.063 & 0.967 & 1.028 & 0.339  & 0.514 & 0.791 & 0.793 & 0.609 & 0.846 & 0.435 & 0.537 & 0.628 & 0.612 \\
                                & MAE    & 0.273 & 0.296 & 0.412 & 0.445 & {\bf\color{red}0.357} & 13.77\% & 0.291 & 0.300 & 0.473 & 0.591 &  \underline{\color{blue}0.414} & 0.771 & 0.864 & 0.824 & 0.801 & 0.815 & 0.416  & 0.528 & 0.673 & 0.681 & 0.575 & 0.628 & 0.399 & 0.472 & 0.524 & 0.506 \\
                                & RMSP & 0.177 & 0.189 & 0.256 & 0.310   & {\bf\color{red}0.233} & 11.74\% & 0.187 & 0.186 & 0.312 & 0.369 &  \underline{\color{blue}0.264} & 0.777 & 0.874 & 0.866 & 0.842 & 0.840 & 0.401  & 0.503 & 0.632 & 0.638 & 0.544 & 0.494 & 0.248 & 0.332 & 0.384 & 0.365 \\
                                & sMAPE  & 0.388 & 0.406 & 0.589 & 0.652 & {\bf\color{red}0.509} & 11.48\% & 0.420 & 0.423 & 0.675 & 0.781 &  \underline{\color{blue}0.575} & 1.220 & 1.314 & 1.288 & 1.309 & 1.283 & 0.603  & 0.817 & 1.022 & 1.038 & 0.870 & 0.865 & 0.554 & 0.657 & 0.721 & 0.699 \\
                                & MASE   & 1.401 & 1.294 & 1.218 & 1.275 & 1.297 & -3.10\% & 1.342 & 1.299 & 1.183 & 1.208 & 1.258 & 1.025 & 1.030 & 1.047 & 1.094 & \underline{\color{blue}1.049} & 1.074  & 1.060 & 1.066 & 1.128 & 1.082 & 1.027 & 1.052 & 1.073 & 1.137 & \underline{\color{blue}1.072} \\
                                & Q25    & 0.202 & 0.230 & 0.247 & 0.258 & {\bf\color{red}0.234} & 8.24\%  & 0.217 & 0.211 & 0.272 & 0.320 & \underline{\color{blue}0.255} & 0.417 & 0.459 & 0.437 & 0.421 & 0.434 & 0.330  & 0.300 & 0.355 & 0.351 & 0.334 & 0.355 & 0.257 & 0.285 & 0.318 & 0.304 \\
                                & Q75    & 0.344 & 0.362 & 0.578 & 0.633 & {\bf\color{red}0.479} & 16.40\% & 0.365 & 0.390 & 0.675 & 0.863 & \underline{\color{blue}0.573} & 1.126 & 1.270 & 1.211 & 1.182 & 1.197 & 0.503  & 0.757 & 0.992 & 1.011 & 0.816 & 0.902 & 0.541 & 0.658 & 0.730 & 0.708 \\ \bottomrule
  \multicolumn{2}{c|}{Univariate}              & \multicolumn{6}{c|}{Minsuforemr-96}    & \multicolumn{5}{c|}{iTransformer-96}             & \multicolumn{5}{c|}{N-Beats-96}       & \multicolumn{5}{c|}{N-Hits-96}      & \multicolumn{5}{c}{Autoformer-96}       \\ \toprule
  Dataset                       & Metric & Y1    & Y2    & Y3    & Y4    & Avg   & IMP     & Y1    & Y2    & Y3    & Y4    & Avg   & Y1    & Y2    & Y3    & Y4    & Avg   & Y1     & Y2    & Y3    & Y4    & Avg   & Y1    & Y2    & Y3    & Y4    & Avg   \\ \toprule  
  \multirow{8}{*}{\rotatebox{90}{M4 Hourly}}         
                                & MSE    & 0.177 & 0.203 & 0.242 & 0.270  & {\bf\color{red}0.223} & 28.06\% & 0.277 & 0.290 & 0.326 & 0.345  & \underline{\color{blue}0.310} & {0.326}  & 0.335  & 0.342  & 0.345 & 0.337 & {0.300}  & 0.342  & 0.372  & 0.37   & 0.346 & {0.598}  & 0.653  & 0.587  & 0.703   & 0.635 \\ \toprule
                                & MAE    & 0.245 & 0.270 & 0.304 & 0.328  & {\bf\color{red}0.287} & 26.41\% & 0.369 & 0.373 & 0.401 & 0.415  & \underline{\color{blue}0.390} & {0.412}  & 0.423  & 0.427  & 0.431 & 0.423 & {0.39}   & 0.421  & 0.433  & 0.428  & 0.418 & {0.583}  & 0.621  & 0.589  & 0.649   & 0.611 \\
                                & RMDSPE & 0.182 & 0.215 & 0.248 & 0.275  & {\bf\color{red}0.230} & 38.83\% & 0.354 & 0.354 & 0.389 & 0.405  & \underline{\color{blue}0.376} & {0.375}  & 0.394  & 0.396  & 0.403 & 0.392 & {0.367}  & 0.41   & 0.418  & 0.414  & 0.402 & {0.576}  & 0.635  & 0.592  & 0.651   & 0.614 \\
                                & MAPE   & 1.364 & 1.478 & 1.688 & 1.819  & {\bf\color{red}1.587} & 27.43\% & 2.108 & 2.109 & 2.224 & 2.305  & \underline{\color{blue}2.187} & {2.532}  & 2.576  & 2.510  & 2.614 & 2.558 & {2.344}  & 2.523  & 2.569  & 2.531  & 2.492 & {3.830}  & 4.019  & 3.701  & 4.369   & 3.980 \\
                                & sMAPE  & 0.445 & 0.488 & 0.534 & 0.569  & {\bf\color{red}0.509} & 22.41\% & 0.631 & 0.636 & 0.669 & 0.687  & \underline{\color{blue}0.656} & {0.680}  & 0.701  & 0.707  & 0.711 & 0.700 & {0.644}  & 0.689  & 0.708  & 0.705  & 0.687 & {0.853}  & 0.896  & 0.865  & 0.909   & 0.881 \\
                                & MASE   & 0.226 & 0.265 & 0.292 & 0.322  & {\bf\color{red}0.276} & 24.59\% & 0.339 & 0.354 & 0.376 & 0.395  & \underline{\color{blue}0.366} & {0.365}  & 0.396  & 0.399  & 0.398 & 0.390 & {0.359}  & 0.395  & 0.415  & 0.415  & 0.396 & {0.481}  & 0.522  & 0.521  & 0.546   & 0.518 \\
                                & Q25    & 0.230 & 0.249 & 0.276 & 0.294  & {\bf\color{red}0.262} & 29.00\% & 0.353 & 0.354 & 0.379 & 0.389  & 0.369 & {0.344}  & 0.351  & 0.358  & 0.360 & 0.353 & {0.38}   & 0.4    & 0.407  & 0.394  & 0.395 & {0.460}  & 0.451  & 0.449  & 0.563   & 0.481 \\
                                & Q75    & 0.259 & 0.291 & 0.332 & 0.362  & {\bf\color{red}0.311} & 24.15\% & 0.384 & 0.392 & 0.424 & 0.441  & \underline{\color{blue}0.410} & {0.479}  & 0.496  & 0.496  & 0.502 & 0.493 & {0.401}  & 0.442  & 0.46   & 0.463  & 0.442 & {0.707}  & 0.792  & 0.729  & 0.735   & 0.741 \\ \toprule
  \multirow{8}{*}{\rotatebox{90}{Us\_births}} 
                                & MSE    & 0.316 & 0.401 & 0.334 & 0.405  & {\bf\color{red}0.364} & 3.70\%  & 0.353 & 0.437 & 0.354 & 0.367  & \underline{\color{blue}0.378} & {0.562}  & 0.705  & 0.608  & 0.804 & 0.670 & {0.371}  & 0.528  & 0.448  & 0.806  & 0.538 & {0.545}  & 0.812  & 0.984  & 1.264   & 0.901 \\
                                & MAE    & 0.391 & 0.459 & 0.412 & 0.462  & {\bf\color{red}0.431} & 1.82\%  & 0.413 & 0.473 & 0.434 & 0.436  & \underline{\color{blue}0.439} & {0.586}  & 0.653  & 0.599  & 0.715 & 0.638 & {0.467}  & 0.579  & 0.529  & 0.737  & 0.578 & {0.583}  & 0.709  & 0.793  & 0.943   & 0.757 \\
                                & RMDSPE & 0.217 & 0.267 & 0.235 & 0.271  & {\bf\color{red}0.248} & 1.20\%  & 0.228 & 0.268 & 0.252 & 0.256  & \underline{\color{blue}0.251} & {0.376}  & 0.414  & 0.377  & 0.469 & 0.409 & {0.283}  & 0.363  & 0.331  & 0.477  & 0.364 & {0.356}  & 0.417  & 0.458  & 0.558   & 0.447 \\
                                & MAPE   & 0.754 & 0.765 & 0.709 & 0.784  & {\bf\color{red}0.753} & 3.71\%  & 0.776 & 0.817 & 0.813 & 0.720  & \underline{\color{blue}0.782} & {0.974}  & 1.121  & 0.932  & 1.037 & 1.016 & {0.851}  & 1.024  & 0.879  & 1.201  & 0.989 & {0.844}  & 0.968  & 1.235  & 1.447   & 1.124 \\
                                & sMAPE  & 0.403 & 0.462 & 0.416 & 0.497  & \underline{\color{blue}0.445} & -1.60\% & 0.409 & 0.466 & 0.420 & 0.458  & {\bf\color{red}0.438} & {0.585}  & 0.628  & 0.590  & 0.698 & 0.625 & {0.447}  & 0.536  & 0.492  & 0.684  & 0.540 & {0.634}  & 0.762  & 0.819  & 0.940   & 0.789 \\
                                & MASE   & 0.328 & 0.393 & 0.351 & 0.381  & {\bf\color{red}0.363} & 0.55\%  & 0.336 & 0.380 & 0.376 & 0.367  & \underline{\color{blue}0.365} & {0.504}  & 0.560  & 0.513  & 0.605 & 0.546 & {0.390}  & 0.486  & 0.438  & 0.542  & 0.464 & {0.546}  & 0.676  & 0.768  & 0.956   & 0.737 \\
                                & Q25    & 0.462 & 0.517 & 0.438 & 0.590  & 0.502 & -2.45\% & 0.497 & 0.559 & 0.395 & 0.510  & 0.490 & {0.440}  & 0.459  & 0.421  & 0.450 & \underline{\color{blue}0.443} & {0.309}  & 0.356  & 0.332  & 0.419  & {\bf\color{red}0.354} & {0.587}  & 0.676  & 0.753  & 0.891   & 0.727 \\
                                & Q75    & 0.321 & 0.401 & 0.387 & 0.335  & {\bf\color{red}0.361} & 6.96\%  & 0.329 & 0.386 & 0.472 & 0.363  & \underline{\color{blue}0.388} & {0.732}  & 0.846  & 0.777  & 0.979 & 0.834 & {0.624}  & 0.801  & 0.727  & 1.056  & 0.802 & {0.580}  & 0.743  & 0.833  & 0.995   & 0.788 \\ \toprule
  \multirow{8}{*}{\rotatebox{90}{Sunspot}}    
                                & MSE    & 1.185 & 1.265 & 1.170 & 1.155  & 1.194 & 1.00\%  & 1.207 & 1.286 & 1.175 & 1.156  & 1.206 & {1.044}  & 1.049  & 1.009  & 1.010 & \underline{\color{blue}1.028} & {0.990}  & 1.004  & 0.971  & 0.961  & {\bf\color{red}0.982} & {1.220}  & 1.290  & 1.224  & 1.275   & 1.252 \\
                                & MAE    & 0.535 & 0.564 & 0.542 & 0.536  & 0.544 & 1.45\%  & 0.542 & 0.572 & 0.548 & 0.545  & 0.552 & {0.539}  & 0.550  & 0.541  & 0.540 & \underline{\color{blue}0.543} & {0.544}  & 0.532  & 0.526  & 0.525  & {\bf\color{red}0.532} & {0.607}  & 0.660  & 0.640  & 0.652   & 0.640 \\
                                & RMDSPE & 0.805 & 0.838 & 0.798 & 0.789  & {\bf\color{red}0.808} & 2.88\%  & 0.810 & 0.868 & 0.822 & 0.828  & \underline{\color{blue}0.832} & {0.960}  & 0.984  & 0.954  & 0.949 & 0.962 & {0.880}  & 0.887  & 0.823  & 0.870  & 0.865 & {1.024}  & 1.087  & 1.046  & 1.073   & 1.058 \\
                                & MAPE   & 2.725 & 2.716 & 2.818 & 2.844  & 2.776 & 2.87\%  & 2.817 & 2.824 & 2.886 & 2.906  & 2.858 & {1.896}  & 1.941  & 2.090  & 2.142 & {\bf\color{red}2.017} & {2.770}  & 2.315  & 2.729  & 2.575  & \underline{\color{blue}2.597} & {2.487}  & 2.928  & 2.775  & 3.033   & 2.806 \\
                                & sMAPE  & 1.044 & 1.059 & 1.056 & 1.051  & {\bf\color{red}1.053} & 1.22\%  & 1.028 & 1.083 & 1.070 & 1.083  & \underline{\color{blue}1.066} & {1.406}  & 1.419  & 1.369  & 1.352 & 1.387 & {1.186}  & 1.214  & 1.139  & 1.216  & 1.189 & {1.349}  & 1.416  & 1.396  & 1.317   & 1.370 \\
                                & MASE   & 1.413 & 1.315 & 1.134 & 1.065  & 1.232 & 2.07\%  & 1.518 & 1.275 & 1.133 & 1.107  & 1.258 & {1.069}  & 0.925  & 0.891  & 0.885 & \underline{\color{blue}0.943} & {0.926}  & 0.863  & 0.811  & 0.830  & {\bf\color{red}0.858} & {1.022}  & 1.014  & 1.060  & 1.130   & 1.057 \\
                                & Q25    & 0.391 & 0.397 & 0.416 & 0.416  & {\bf\color{red}0.405} & 2.64\%  & 0.391 & 0.406 & 0.426 & 0.442  & \underline{\color{blue}0.416} & {0.509}  & 0.514  & 0.514  & 0.518 & 0.514 & {0.523}  & 0.486  & 0.488  & 0.512  & 0.502 & {0.545}  & 0.616  & 0.619  & 0.641   & 0.605 \\
                                & Q75    & 0.669 & 0.701 & 0.670 & 0.657  & 0.674 & 1.89\%  & 0.692 & 0.738 & 0.670 & 0.648  & 0.687 & {0.569}  & 0.586  & 0.568  & 0.563 & \underline{\color{blue}0.572} & {0.566}  & 0.578  & 0.565  & 0.538  & {\bf\color{red}0.562} & {0.669}  & 0.704  & 0.660  & 0.664   & 0.674 \\ \toprule
  \multirow{8}{*}{\rotatebox{90}{Saugeenday}} 
                                & MSE    & 0.276 & 0.31  & 0.362 & 0.56   & 0.377 & 1.31\%  & 0.279 & 0.312 & 0.366 & 0.569  & 0.382 & {0.276}  & 0.308  & 0.357  & 0.512 & {\bf\color{red}0.363} & {0.282}  & 0.312  & 0.362  & 0.505  & \underline{\color{blue}0.365} & {0.296}  & 0.324  & 0.373  & 0.607   & 0.400 \\
                                & MAE    & 0.371 & 0.397 & 0.431 & 0.543  & {\bf\color{red}0.436} & 0.91\%  & 0.374 & 0.399 & 0.436 & 0.551  & \underline{\color{blue}0.440}  & {0.378}  & 0.405  & 0.444  & 0.552 & 0.445 & {0.384}  & 0.401  & 0.449  & 0.534  & 0.442 & {0.393}  & 0.412  & 0.442  & 0.584   & 0.458 \\
                                & RMDSPE & 0.367 & 0.398 & 0.439 & 0.575  & {\bf\color{red}0.445} & 1.11\%  & 0.369 & 0.4   & 0.444 & 0.587  & \underline{\color{blue}0.450}  & {0.373}  & 0.406  & 0.458  & 0.657 & 0.474 & {0.382}  & 0.404  & 0.456  & 0.608  & 0.463 & {0.392}  & 0.413  & 0.453  & 0.636   & 0.474 \\
                                & MAPE   & 71.15 & 78.47 & 86.28 & 100.96 & 84.21 & 1.36\%  & 71.77 & 78.79 & 87.96 & 102.97 & 85.37 & {67.265} & 71.937 & 77.889 & 74.39 & {\bf\color{red}72.87} & {63.643} & 70.235 & 81.053 & 77.865 & \underline{\color{blue}73.20}  & {72.197} & 78.669 & 87.303 & 107.281 & 86.36 \\
                                & sMAPE  & 0.671 & 0.709 & 0.752 & 0.892  & {\bf\color{red}0.756} & 0.66\%  & 0.674 & 0.711 & 0.758 & 0.901  & \underline{\color{blue}0.761} & {0.687}  & 0.732  & 0.793  & 1.003 & 0.804 & {0.701}  & 0.724  & 0.791  & 0.949  & 0.791 & {0.703}  & 0.73   & 0.768  & 0.944   & 0.786 \\
                                & MASE   & 1.211 & 1.234 & 1.323 & 1.688  & 1.364 & 1.23\%  & 1.229 & 1.293 & 1.34  & 1.66   & 1.381 & {1.199}  & 1.235  & 1.307  & 1.357 & \underline{\color{blue}1.275} & {1.282}  & 1.268  & 1.367  & 1.434  & 1.338 & {1.092}  & 1.143  & 1.229  & 1.556   & {\bf\color{red}1.255} \\
                                & Q25    & 0.369 & 0.397 & 0.425 & 0.536  & {\bf\color{red}0.432} & 2.92\%  & 0.375 & 0.401 & 0.443 & 0.561  & \underline{\color{blue}0.445} & {0.387}  & 0.42   & 0.465  & 0.584 & 0.464 & {0.394}  & 0.393  & 0.49   & 0.537  & 0.454 & {0.406}  & 0.431  & 0.447  & 0.639   & 0.481 \\
                                & Q75    & 0.374 & 0.393 & 0.435 & 0.535  & 0.434 & 0.23\%  & 0.373 & 0.397 & 0.429 & 0.542  & 0.435 & {0.368}  & 0.391  & 0.422  & 0.519 & {\bf\color{red}0.425} & {0.373}  & 0.408  & 0.409  & 0.53   & \underline{\color{blue}0.430} & {0.38}   & 0.394  & 0.437  & 0.53    & 0.435 \\ \bottomrule
  \end{tabular}
  }
\end{table}

\newpage
\section{Visualization of TS forecasting}
\label{app_vis_tsf}

\begin{figure*}[!ht]
  \centering
  \vspace{0.2em}
  \centerline{\includegraphics[width=\textwidth]{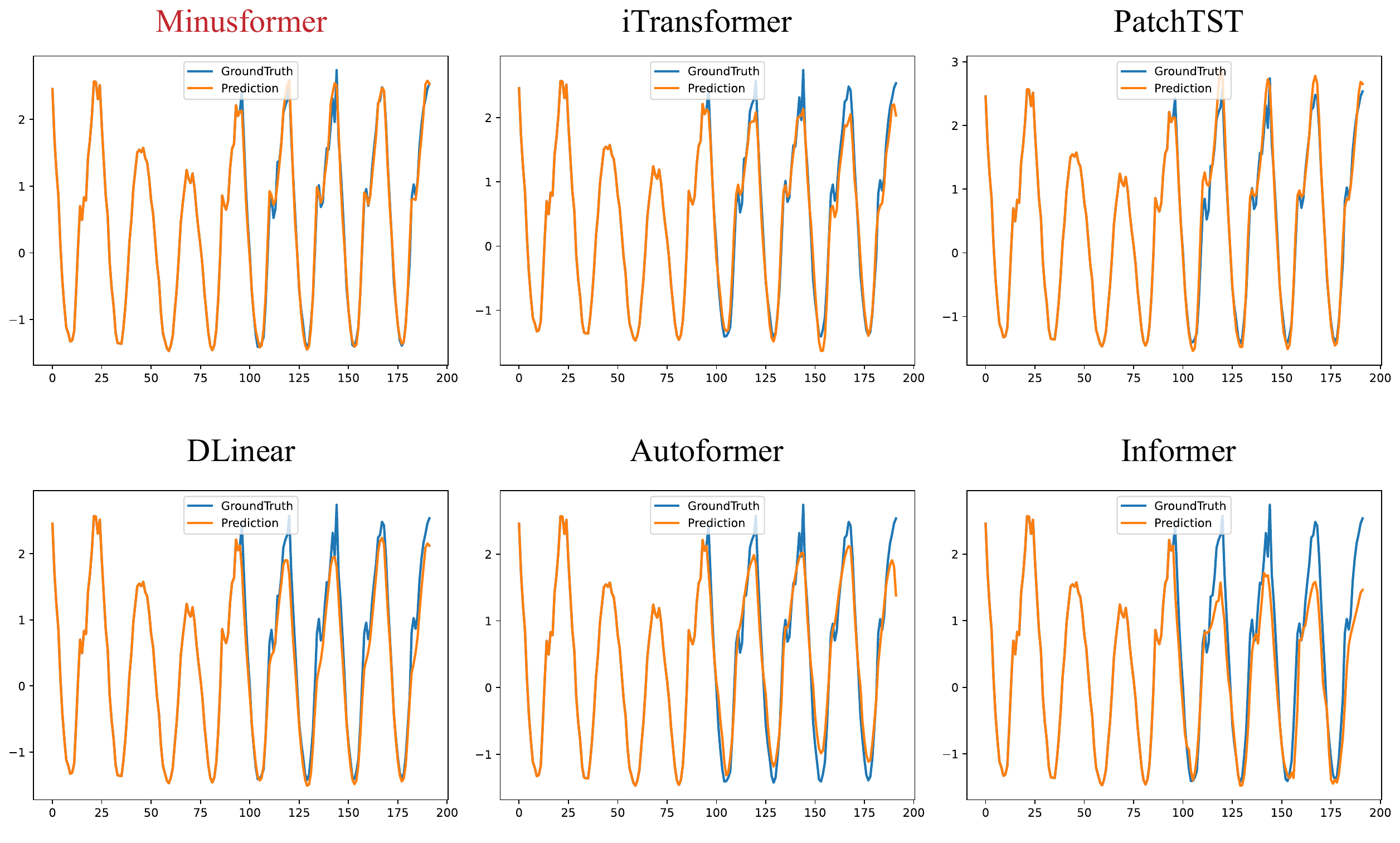}}
  \caption{Prediction cases from the Traffic dataset under the input-96-predict-96 setting.}
  \label{fig_traffic} 
\end{figure*}

\begin{figure*}[!ht]
  \centering
  \centerline{\includegraphics[width=\textwidth]{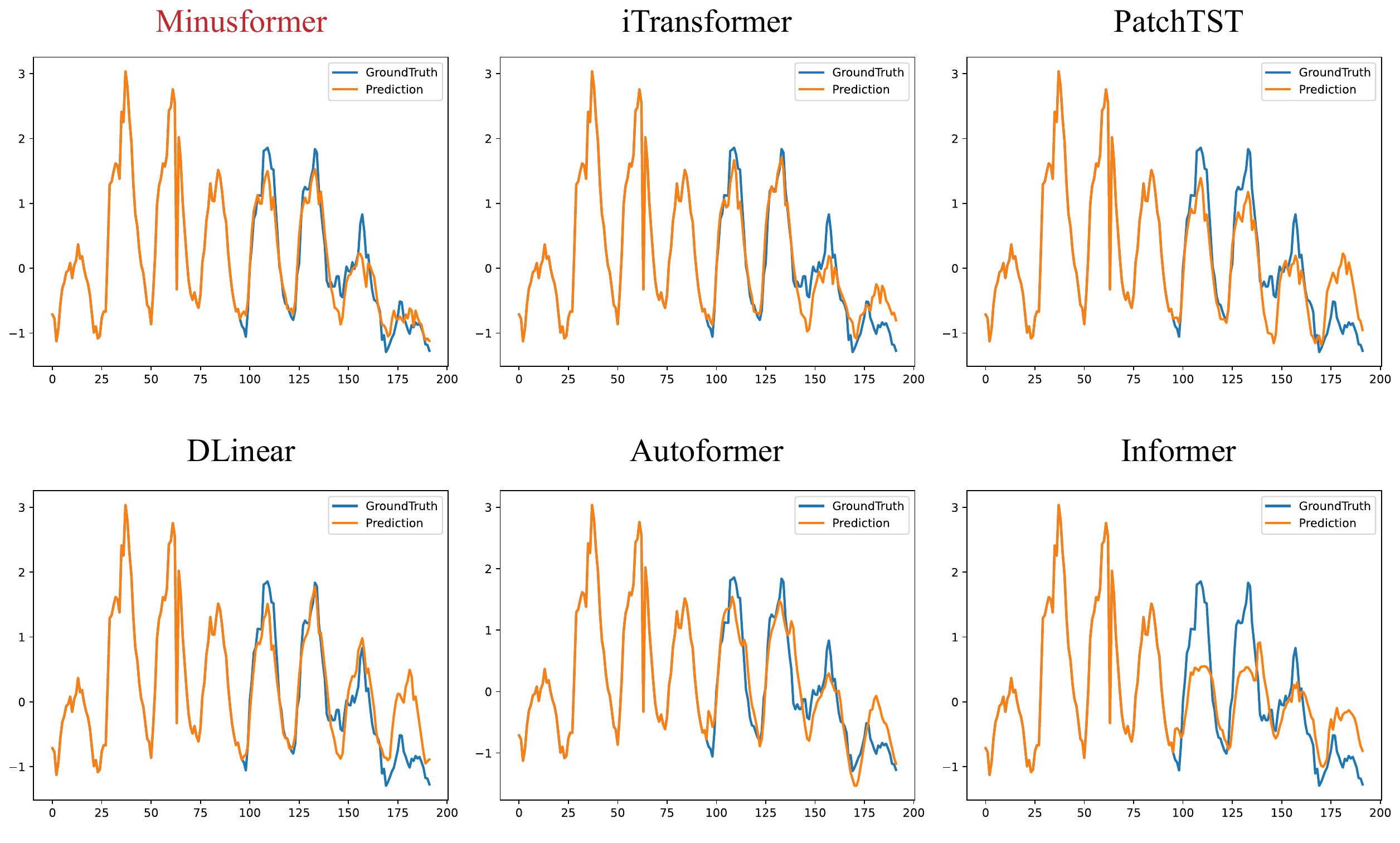}}
  \caption{Prediction cases from the Electricity dataset under the input-96-predict-96 setting.}
  \label{fig_elc} 
\end{figure*}

\begin{figure*}[!ht]
  \centering
  \centerline{\includegraphics[width=\textwidth]{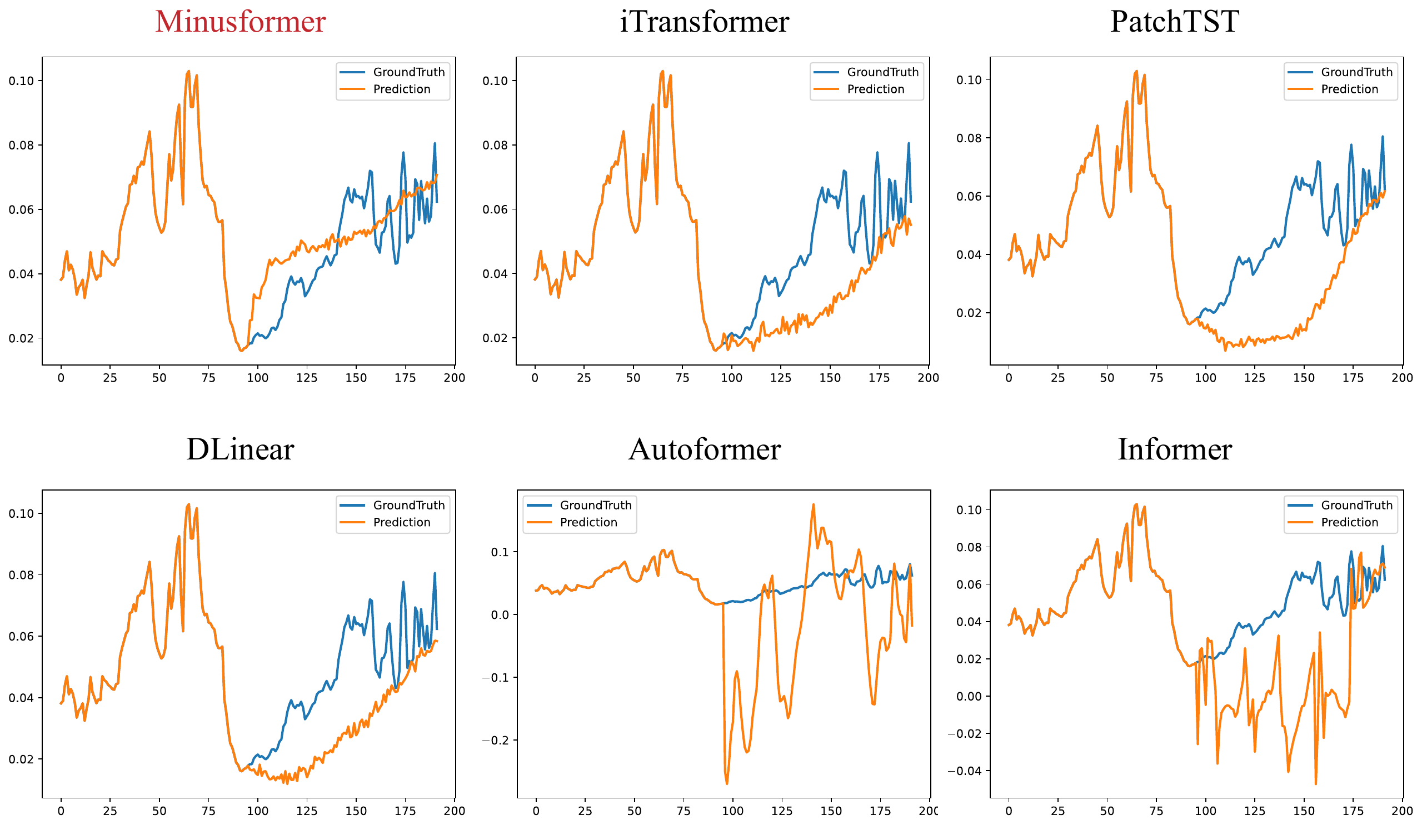}}
  \caption{Prediction cases from the Weather dataset under the input-96-predict-96 setting.}
  \label{fig_weather} 
\end{figure*}

For clarity and comparison among different models, we present supplementary prediction showcases for three representative datasets in Fig. \ref{fig_traffic}, \ref{fig_elc}, and \ref{fig_weather}. 
Visualization of different models for qualitative comparisons. Prediction cases from the Traffic, Electricity and Weather datasets.
These showcases correspond to predictions made by the following models: iTransformer \citep{liu2023itransformer}  and PatchTST \citep{nie2022time}, DLinear \citep{zeng2023transformers}, Autoformer \citep{wu2021autoformer} and Informer \citep{Zhou2021Informer}.
Among the various models considered, the proposed Minusformer stands out for its ability to predict future series variations with exceptional precision, demonstrating superior performance.

\section{Limitations of the proposed Minusformer} 
\label{app_limitations}

The proposed Minusformer has several limitations that are worth mentioning. Minsformer is a generalization of the Transformer architecture, and as such, it inherits some of its limitations, e.g., the computational complexity is $\mathcal{O}(N^2)$. This complexity is a drawback for long sequences, and it is the main reason why the Transformer is not suitable for long sequences. The proposed Minusformer is not an exception to this limitation. 
Minsformer is only given the input sequence and the target sequence, and it does not have any other information about the data. This limitation is the drawback for some applications, e.g., when the data has a specific structure, such as images, videos, or text. In these cases, the Transformer can exploit the structure of the data to improve its performance.

\end{document}